\documentclass[acmsmall, review=false, screen]{acmart}
\setcitestyle{super,sort&compress}
\usepackage{booktabs} 
\usepackage{xcolor}
\usepackage{colortbl}
\usepackage{moresize}

\usepackage{amsmath}
\usepackage{mathtools}
\DeclareMathOperator*{\argmin}{arg\,min}
\DeclareMathOperator*{\e}{\mathbb{E}}
\DeclareMathOperator*{\p}{\mathbb{P}}
\newcommand{\Var}{\operatorname{Var}} 

\usepackage{graphicx}
\usepackage{subcaption}
\usepackage{algorithm,algorithmicx,algpseudocode}
\usepackage{multirow}

\usepackage[bottom]{footmisc}

\definecolor{lightgray}{gray}{0.9}
\definecolor{darkgreen}{RGB}{0,127,0}
\acmArticle{1}
\setcopyright{acmcopyright}

\begin{document}
\title{Demystifying Parallel and Distributed Deep Learning: An In-Depth Concurrency Analysis}
\author{Tal Ben-Nun}
\orcid{0000-0002-3657-6568}
\email{talbn@inf.ethz.ch}
\author{Torsten Hoefler}
\email{htor@inf.ethz.ch}
\affiliation{%
  \institution{ETH Zurich}
  \department{Department of Computer Science}
  \city{Z\"{u}rich}
  \postcode{8006}
  \country{Switzerland}
}

\begin{abstract}
Deep Neural Networks (DNNs) are becoming an important tool in modern computing applications. Accelerating their training is a major challenge and techniques range from distributed algorithms to low-level circuit design. In this survey, we describe the problem from a theoretical perspective, followed by approaches for its parallelization. We present trends in DNN architectures and the resulting implications on parallelization strategies. We then review and model the different types of concurrency in DNNs: from the single operator, through parallelism in network inference and training, to distributed deep learning. We discuss asynchronous stochastic optimization, distributed system architectures, communication schemes, and neural architecture search. Based on those approaches, we extrapolate potential directions for parallelism in deep learning.
\end{abstract}

%
%


%
%


\maketitle

\section{Introduction}
Machine Learning, and in particular Deep Learning \cite{nature15}, is rapidly taking over a variety of aspects in our daily lives. At the core of deep learning lies the Deep Neural Network (DNN), a construct inspired by the interconnected nature of the human brain. Trained properly, the expressiveness of DNNs provides accurate solutions for problems previously thought to be unsolvable, merely by observing large amounts of data.
Deep learning has been successfully implemented for a plethora of fields, ranging from image classification \cite{densenet}, through speech recognition \cite{deepspeech} and medical diagnosis \cite{breastcancer}, to autonomous driving \cite{selfdriving} and defeating human players in complex games \cite{silver2017mastering}.

Since the 1980s, neural networks have attracted the attention of the machine learning community \cite{mnistlecun}. However, DNNs' rise into prominence was tightly coupled to the available computational power, which allowed to exploit their inherent parallelism. Consequently, deep learning managed to outperform all existing approaches in speech recognition \cite{lee09audio} and image classification \cite{alexnet}, where the latter (AlexNet) increased the accuracy by a factor of two, sparking interest outside of the community and even academia.

As datasets increase in size and DNNs in complexity, the computational intensity and memory demands of deep learning increase proportionally. Training a DNN to competitive accuracy today essentially requires a high-performance computing cluster. To harness such systems, different aspects of training and inference (evaluation) of DNNs are modified to increase concurrency.

In this survey, we discuss the variety of topics in the context of parallelism and distribution in deep learning, spanning from vectorization to efficient use of supercomputers. In particular, we present parallelism strategies for DNN evaluation and implementations thereof, as well as extensions to training algorithms and systems targeted at supporting distributed environments. To provide comparative measures on the approaches, we analyze their concurrency and average parallelism using the Work-Depth model \cite{wdmodel}.

\subsection{Related Surveys}

Other surveys in the field focus on applications of deep learning \cite{Najafabadi2015}, neural networks and their history \cite{nature15,schmidhuber15,wang17survey,li17surveydrl}, scaling up deep learning \cite{Bengio2013}, and hardware architectures for DNNs \cite{ienne93,lacey16,sze17survey}.

In particular, three surveys \cite{nature15,schmidhuber15,wang17survey} describe DNNs and the origins of deep learning methodologies from a historical perspective, as well as discuss the potential capabilities of DNNs w.r.t. learnable functions and representational power. Two of the three surveys \cite{schmidhuber15,wang17survey} also describe optimization methods and applied regularization techniques in detail.

Bengio \cite{Bengio2013} discusses scaling deep learning from various perspectives, focusing on models, optimization algorithms, and datasets. The paper also overviews some aspects of distributed computing, including asynchronous and sparse communication.

Surveys of hardware architectures mostly focus on the computational side of training rather than the optimization. This includes a recent survey \cite{sze17survey} that reviews computation techniques for DNN operators (layer types) and mapping computations to hardware, exploiting inherent parallelism. The survey also includes discussion on data representation reduction (e.g., via quantization) to reduce overall memory bandwidth within the hardware. Other surveys discuss accelerators for traditional neural networks \cite{ienne93} and the use of FPGAs in deep learning \cite{lacey16}.

\subsection{Scope}

\begin{figure}[t]
	\centering
	\includegraphics[height=1.18in]{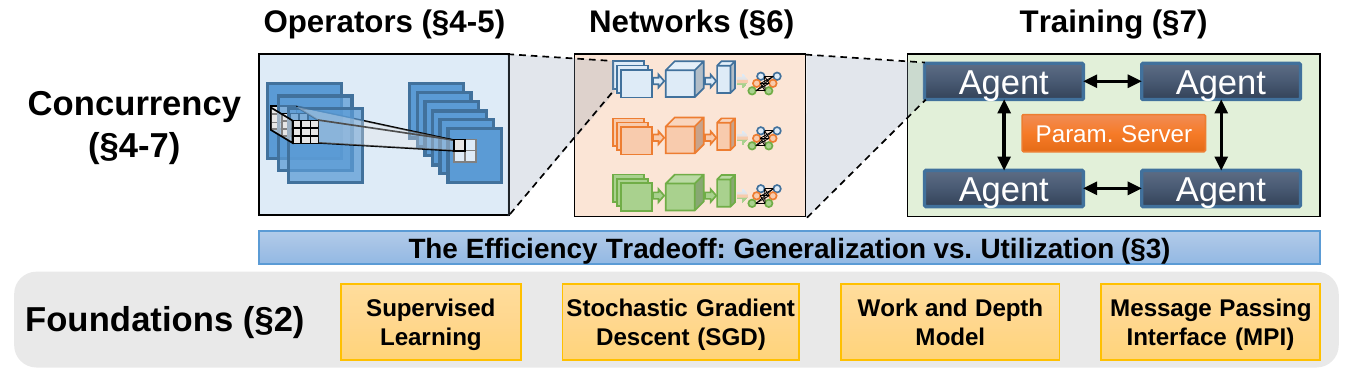}
	\vspace{-1em}
	\caption{Summary of Concurrency in Deep Learning}
	\label{fig:infra}
	\vspace{-1.2em}
\end{figure}

In this paper, we provide a comprehensive review and analysis of parallel and distributed deep learning, summarized in Fig. \ref{fig:infra} and organized as follows:

\begin{itemize}
	\item Section \ref{sec:term} defines our terminology and algorithms.
	\item Section \ref{sec:tradeoff} discusses the tradeoffs between concurrency and accuracy in deep learning.
	\item Section \ref{sec:dnn} describes DNN operators, how they are computed, and Section \ref{sec:layercomp} shows how they can be modified to expose concurrency.
	\item Section \ref{sec:par} explores and analyzes the main approaches for parallelism in computations of full networks for training and inference.
	\item Section \ref{sec:dist} provides an overview of distributed training, describing algorithm modifications, techniques to reduce communication, and system implementations.
	\item Section \ref{sec:conclusions} gives concluding remarks and extrapolates potential directions in the field.
\end{itemize}

The paper surveys 240 other works, obtained by recursively tracking relevant bibliography from 
seminal papers in the field, dating back to the year 1984. We include additional papers resulting from 
keyword searches on Google Scholar\footnote{\url{https://scholar.google.com/}} and arXiv\footnote{\url{https://www.arxiv.org/}}. Due to the quadratic increase in deep learning papers on the latter source (Table \ref{fig:arxiv}), some works may not have been included. The full list of categorized papers in this survey can be found online\footnote{\url{https://spcl.inf.ethz.ch/Research/Parallel\_Programming/DistDL/}}.
\begin{table}[t]
	\caption{Yearly arXiv Papers in Computer Science (AI and Computer Vision)}
	\vspace{-1em}
	\label{fig:arxiv}
	\scriptsize
	\begin{tabular}{ l c c c c c c }
		\toprule
		\bf Year   & \bf 2012 & \bf 2013 & \bf 2014 & \bf 2015 & \bf 2016 & \bf 2017\\
		\midrule
		\bf cs.AI & 1,081 & 1,765 & 1,022 & 1,105 & 1,929 & 2,790\\
		\bf cs.CV & 577   & 852   & 1,349 & 2,261 & 3,627 & 5,693\\
		\bottomrule		
	\end{tabular}
\end{table}

\section{Terminology and Algorithms}
\label{sec:term}

This section establishes theory and naming conventions for the material presented in the survey. We first discuss the class of supervised learning problems, followed by relevant foundations of parallel programming.

\subsection{Supervised Learning}
\label{sec:term:sl}

In machine learning, Supervised Learning \cite{shalev14} is the process of optimizing a function from a set of labeled samples (\textit{dataset}) such that, given a sample, the function would return a value that approximates the label. It is assumed that both the dataset and other, unobserved samples, are sampled from the same probability distribution.

\begin{figure}[t]
	\centering
	\begin{minipage}[]{.475\textwidth}
		\centering
		\tiny
		\renewcommand{\arraystretch}{1.2}
		\begin{tabular}{ l p{0.66\linewidth} }
			\toprule
			\bf Name & \bf Definition \\\midrule
			$\mathcal{D}$ & Data probability distribution \\
			$S$ & Training dataset \\
			$w\in \mathcal{H}$ & Model parameters. $w^{(t)}_i$ denotes parameter $i$ at SGD iteration $t$\\
			$f_w(z)$ & Model function (learned predictor) \\
			$h(z)$ & Ground-truth label (in Supervised Learning)\\
			$\ell(w, z)$ & Per-sample loss function \\
			$\nabla\ell(w, z)$ & Gradient of $\ell$ \\
			$u(g,w,t)$ & Parameter update rule. Function of loss gradient $g$, parameters $w$, and iteration $t$\\
			\bottomrule
		\end{tabular}\vspace{0.1in}
		\renewcommand{\arraystretch}{1}
		\captionof{table}{Summary of Notations}
		\label{tbl:term}
	\end{minipage}
	\begin{minipage}[]{.5\textwidth}
		\centering
		\includegraphics[height=1.2in,trim={1cm 0.9cm 5cm 4cm},clip]{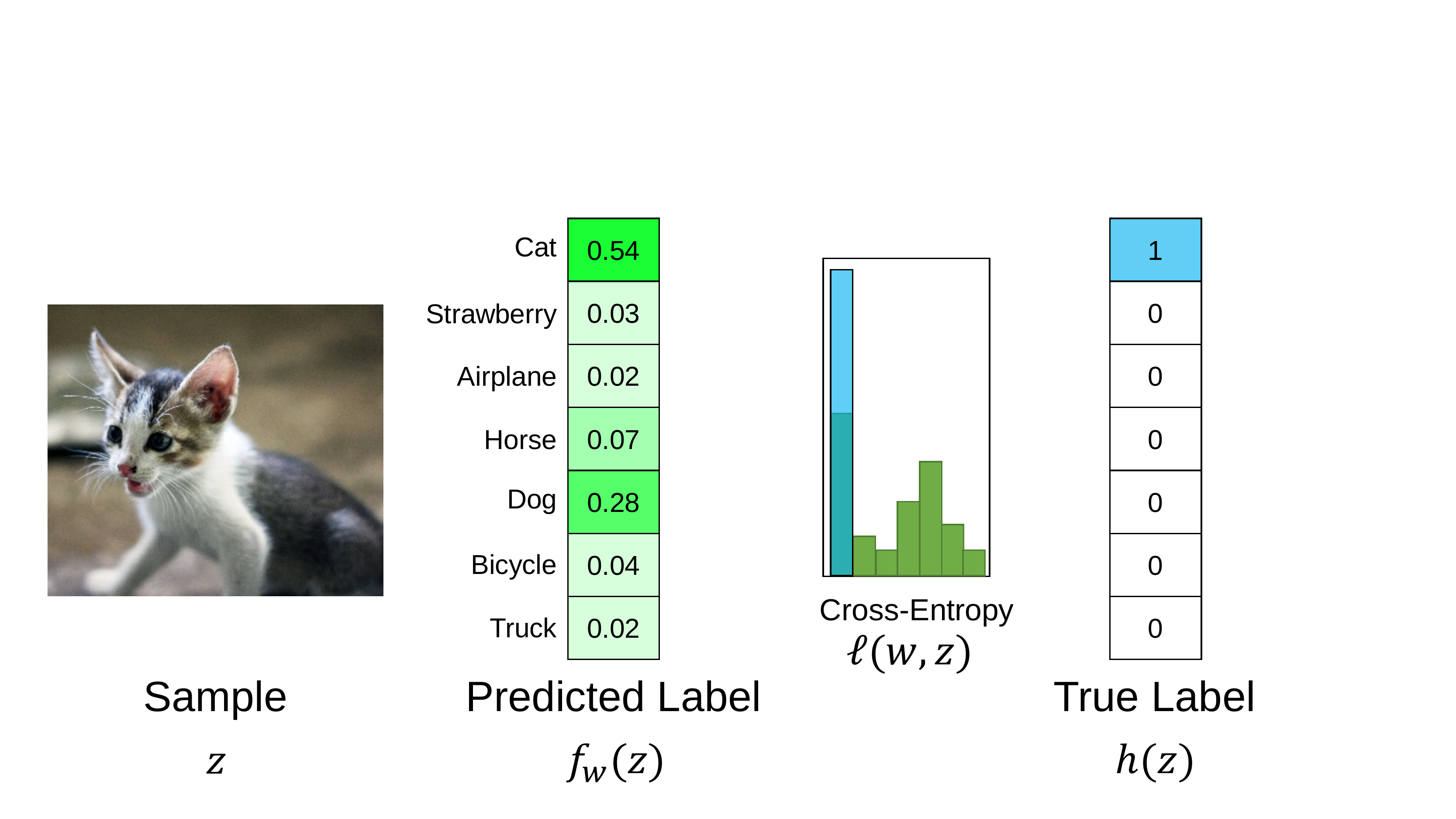}
		\captionof{figure}{Multi-Class Classification Loss}
		\label{fig:loss}
	\end{minipage}%
	\vspace{-2em}
\end{figure}

Throughout the survey, we refer to the operators $\p$ and $\e$ as the probability and expectation of random variables; $z\sim \mathcal{D}$ denotes that a random variable $z$ is sampled from a probability distribution $\mathcal{D}$; and $\e_{z\sim \mathcal{D}}\left[f(z)\right]$ denotes the expected value of $f(z)$ for a random variable $z$. The notations are summarized in Table \ref{tbl:term}.

Formally, given a probability distribution of data $\mathcal{D}$, random variable $z\sim \mathcal{D}$, a domain $X$ where we construct samples from, a label domain $Y$, and a hypothesis class $\mathcal{H}$ containing functions $f:X\rightarrow Y$, we wish to minimize the \textit{generalization error}, defined by the loss function $L_\mathcal{D}(f)\equiv \p\left[f(z)\ne h(z)\right]$, 
where $h(z)$ represents the true label of $z$.
In practice, it is common to use a class $\mathcal{H}$ of functions $f_w$ that are defined by a vector of parameters $w$ (sometimes denoted as $\theta$), in order to define a continuous hypothesis space. For example, $\mathcal{H}$ may represent an N-dimensional hyperplane that separates between samples of two classes, where $w_i$ are its coefficients. In deep neural networks, we define $w$ in multiple layers, namely, $w_{l,i}$ is the parameter at layer $l$ and index $i$.

We wish to find $w^*$ that minimizes the above loss function, as follows:
\begin{equation} \label{eq:loss}
w^*=\argmin_{w \in \mathcal{H}} L_\mathcal{D}\left(f_w\right)=\argmin_{w \in \mathcal{H}} \e_{z\sim\mathcal{D}}\left[\ell\left(w, z\right)\right],
\end{equation}
where $\ell:\mathcal{H}\times X\rightarrow \mathbb{R}_+$ is the loss of an individual sample.

In this work, we consider two types of supervised learning problems, from which the sample loss functions are derived: \textit{(multi-class) classification} and \textit{regression}. In the former, the goal is to identify which class a sample most likely belongs to, e.g., inferring what type of animal appears in an image. In regression, the goal is to find a relation between the domains $X$ and $Y$, predicting values in $Y$ for new samples in $X$. For instance, such a problem might predict the future temperature of a region, given past observations.

For minimization purposes, a sample loss function $\ell$ should be continuous and differentiable. In regression problems, it is possible to use straightforward loss functions such as the squared difference $\ell(w,z)=\left(f_w(z)-h(z)\right)^2$. On the other hand, in classification problems, a simple definition of loss such as $\ell(w,z)=0$ if $f_w(z) = h(z)$ or $1$ otherwise (also known as binary or 0--1 loss), does not match the continuity and differentiability criteria.

To resolve this issue, prominent multi-class classification problems define $Y$ as a probability distribution of the inferred class types (see Fig. \ref{fig:loss}), instead of a single label. The model output is typically normalized to a distribution using the \textit{softmax} function $\sigma(z)_i=\frac{\exp(z_i)}{\sum_k{\exp(z_k)}}$. The loss function then computes the difference of the prediction from the true label ``distribution'', e.g., using cross-entropy: $\ell(w,z)=-\sum_i{h(z)_i \log \sigma\left(f_w(z)\right)_i}$.
The cross-entropy loss can be seen as a generalization of logistic regression, inducing a continuous loss function for multi-class classification.

Minimizing the loss function can be performed by using different approaches, such as iterative methods (e.g., BFGS \cite{nocedal06numopt}) or meta-heuristics (e.g., evolutionary algorithms \cite{regevo}). Optimization in machine learning is prominently performed via Gradient Descent. Since the full $\mathcal{D}$ is, however, never observed, it is necessary to obtain an unbiased estimator of the gradient. Observe that $\nabla L_\mathcal{D} \left(w\right) = \e_{z\sim\mathcal{D}}\left[\nabla \ell\left(w, z\right)\right]$  (Eq. \ref{eq:loss}, linearity of the derivative). Thus, in expectation, we can descend using randomly sampled data in each iteration, applying \textit{Stochastic Gradient Descent} (SGD) \cite{sgd}. \vspace{-0.5em}

\begin{algorithm}[h]
	\begin{algorithmic}[1]
		\For{$t=0$ \textbf{to} $T$} \Comment{SGD iterations}\label{alg:ssgd:stop}
		\State $z\leftarrow$ Random element from $S$ \Comment{Sample dataset $S$}\label{alg:ssgd:sample}
		\State $g\leftarrow \nabla\ell(w^{(t)},z)$ \Comment{Compute gradient of $\ell$}\label{alg:ssgd:grad}
		\State $w^{(t+1)}\leftarrow w^{(t)} + u(g, w^{(0,\dots,t)}, t)$ \Comment{Update weights with function $u$}\label{alg:ssgd:update}
		\EndFor
	\end{algorithmic}
	\caption{Stochastic Gradient Descent (SGD)}
	\label{alg:ssgd}
\end{algorithm}

\vspace{-0.5em}
SGD (Algorithm \ref{alg:ssgd}) iteratively optimizes parameters defined by the sequence $\lbrace w^{(t)}\rbrace_{t=0}^T$, using samples from a dataset $S$ sampled from $\mathcal{D}$ with replacement. SGD is proven to converge at a rate of $\mathcal{O}(1/\sqrt{T})$ for convex functions with Lipschitz-continuous and bounded gradient \cite{nemirovski09}.

Prior to running SGD, one must choose an initial estimate for the weights $w^{(0)}$. Due to the ill-posed nature of some problems, the selection of $w^{(0)}$ is important and may reflect on the final quality of the result. The choice of initial weights can originate from random values, informed decisions (e.g., Xavier initialization \cite{glorot10a}), or from pre-trained weights in a methodology called \textit{Transfer Learning} \cite{transferlearning}. In deep learning, recent works state that the optimization space is riddled with saddle points \cite{nature15}, and assume that the value of $w^{(0)}$ does not affect the final loss. In practice, however, improper initialization may have an adverse effect on generalization as networks become deeper \cite{prelu}.

In line \ref{alg:ssgd:stop}, $T$ denotes the number of steps to run SGD for (known as the \textit{stopping condition} or \textit{computational budget}). Typically, real-world instances of SGD run for a constant number of steps, for a fixed period of time, or until a desired accuracy is achieved. Line \ref{alg:ssgd:sample} then samples random elements from the dataset, so as to provide the unbiased loss estimator. The gradient of the loss function with respect to the weights $w^{(t)}$ is subsequently computed (line \ref{alg:ssgd:grad}). In deep neural networks, the gradient is obtained with respect to each layer ($w^{(t)}_l$) using backpropagation (Section \ref{sec:dnn:bprop}). This gradient is then used for updating the weights, using a \textit{weight update rule} (line \ref{alg:ssgd:update}). 

\begin{table}[t]
	\caption{Popular Weight Update Rules}
	\vspace{-1em}
	\label{tbl:wupdate}
	\scriptsize
	\renewcommand{\arraystretch}{1.3}
	\begin{tabular}{ l l l }
		\toprule
		\bf Method & \bf Formula & \bf Definitions \\\midrule 
		
		Learning Rate & $w^{(t+1)}=w^{(t)} -\eta \cdot \nabla w^{(t)}$ & $\nabla w^{(t)} \equiv \nabla \ell (w^{(t)},z)$\\
		
		Adaptive Learning Rate & $w^{(t+1)}= w^{(t)} -\eta_t \cdot \nabla w^{(t)}$ &\\
		Momentum \cite{momentum} & $w^{(t+1)}=w^{(t)} + \mu\cdot(w^{(t)}-w^{(t-1)}) - \eta \cdot \nabla w^{(t)}$&\\
		Nesterov Momentum \cite{nesterov1983} & $w^{(t+1)}=w^{(t)}+ v_t$ & $v_{t+1}=\mu\cdot v_{t} - \eta \cdot\nabla\ell(w^{(t)} - \mu\cdot v_{t},z)  $\\\addlinespace
		AdaGrad \cite{adagrad} & $w_i^{(t+1)}=w_i^{(t)}-\frac{\eta \cdot \nabla w_i^{(t)}}{\sqrt{A_{i,t}}+\varepsilon}$& $A_{i,t}=\sum_{\tau=0}^{t}{\left(\nabla w_i^{(t)}\right)^2}$ \\\addlinespace
		RMSProp \cite{rmsprop} & $w_i^{(t+1)}=w_i^{(t)}-\frac{\eta \cdot \nabla w_i^{(t)}}{\sqrt{A'_{i,t}}+\varepsilon}$ & $A'_{i,t}=\beta\cdot A'_{t-1}+(1-\beta)\left(\nabla w_i^{(t)}\right)^2$ \\\addlinespace
		Adam \cite{adamopt} & $w_i^{(t+1)}=w_i^{(t)}-\frac{\eta \cdot M^{(1)}_{i,t} }{\sqrt{M^{(2)}_{i,t}}+\varepsilon}$ & $M^{(m)}_{i,t} = \frac{\beta_m\cdot M^{(m)}_{i,t-1} + (1-\beta_m)\left(\nabla w_i^{(t)}\right)^m}{1-\beta_m^t}$ \\\bottomrule
	\end{tabular}
	\renewcommand{\arraystretch}{1}
	\vspace{-1em}
\end{table}

\subsubsection{Weight Update Rules}
The weight update rule, denoted as $u$ in Algorithm \ref{alg:ssgd}, can be defined as a function of the gradient $g$, the previous weight values $w^{(0)},\cdots,w^{(t)}$, and the current iteration $t$. Table \ref{tbl:wupdate} summarizes the popular $u$ functions used in training. In the table, the basic SGD update rule is $u_{sgd}(g)=-\eta\cdot g$, where $\eta$ represents the \textit{learning rate}. $\eta$ controls how much the gradient values will overall affect the next estimate $w^{(t+1)}$, and in iterative nonlinear optimization methods finding the correct $\eta$ is a considerable part of the computation \cite{nocedal06numopt}. In machine learning problems, it is customary to fix $\eta$, or set an iteration-based weight update rule $u_{alr}(g,t)=-\eta_t\cdot g$, where $\eta_t$ decreases (decays) over time to bound the modification size and avoid local divergence.

Other popular weight update rules include \textit{Momentum}, which uses the difference between current and past weights $w^{(t)}-w^{(t-1)}$ to avoid local minima and redundant steps with natural motion \cite{momentum,nesterov1983}. More recent update rules, such as RMSProp \cite{rmsprop} and Adam \cite{adamopt}, use the first and second moments of the gradient in order to adapt the learning rate per-weight, enhancing sparser updates over others. 

Factors such as the learning rate and other symbols found in Table \ref{tbl:wupdate} are called \textit{hyper-parameters}, and are set before the optimization process begins. In the table, $\mu,\beta,\beta_1$, and $\beta_2$ represent the momentum, RMS decay rate, and first and second moment decay rate hyper-parameters, respectively. To obtain the best results, hyper-parameters must be tuned, which can be performed by value sweeps or by meta-optimization (Section \ref{sec:training:hyperparams}). The multitude of hyper-parameters and the reliance upon them is considered problematic by a part of the community \cite{alchemy}.

\vspace{-0.5em}
\begin{algorithm}[h]
	\begin{algorithmic}[1]
		\For{$t = 0$ \textbf{to} $\frac{|S|}{B}\cdot epochs$}
		\State $\vec{z}\leftarrow$ Sample $B$ elements from $S$ \label{alg:sgd:readsamples}\Comment Obtain samples from dataset
		\State $w_{mb}\leftarrow w^{(t)}$       \label{alg:sgd:readweights}\Comment Load parameters
		\State $f \leftarrow \ell(w_{mb}, \vec{z}, h(\vec{z}))$ \label{alg:sgd:fwd} \Comment Compute forward evaluation
		\State $g_{mb}\leftarrow\nabla\ell(w_{mb}, f)$ \label{alg:sgd:glocal} \label{alg:sgd:bwd} \Comment Compute gradient using backpropagation
		\State $\Delta w\leftarrow u(g_{mb},w^{(0,\dots,t)}, t)$ \label{alg:sgd:accgrad} \Comment Weight update rule
		\State $w^{(t+1)}\leftarrow w_{mb}+\Delta w$ \label{alg:sgd:writeweights} \Comment Store parameters
		\EndFor
	\end{algorithmic}
	\caption{Minibatch Stochastic Gradient Descent with Backpropagation}
	\label{alg:sgd}
\end{algorithm}
\vspace{-0.5em}

\subsubsection{Minibatch SGD}
\label{sec:term:sgd}

When performing SGD, it is common to decrease the number of weight updates by computing the sample loss in \textit{minibatches} (Algorithm \ref{alg:sgd}), averaging the gradient with respect to subsets of the data \cite{le11}. Minibatches represent a tradeoff between traditional SGD, which is proven to converge when drawing one sample at a time, and batch methods \cite{nocedal06numopt}, which make use of the entire dataset at each iteration.

In practice, minibatch sampling is implemented by shuffling the dataset $S$, and processing that permutation by obtaining contiguous segments of size $B$ from it. An entire pass over the dataset is called an \textit{epoch}, and a full training procedure usually consists of tens to hundreds of such epochs \cite{goyal17,24min}. As opposed to the original SGD, shuffle-based processing entails without-replacement sampling. Nevertheless, minibatch SGD was proven \cite{shamir16} to provide similar convergence guarantees.

\subsection{Unsupervised and Reinforcement Learning}

Two other classes in machine learning are \textit{unsupervised} and \textit{reinforcement learning}. In the former class, the dataset $S$ is not labeled (i.e., $h(z)$ does not exist) and training typically results in different objective functions, intended to infer structure from the unlabeled data. The latter class refers to tasks where an environment is observed at given points in time, and training optimizes an action policy function to maximize the reward of the observer.

In the context of deep learning, unsupervised learning has two useful implementations: auto-encoders, and Generative Adversarial Networks (GANs) \cite{gan}. Auto-encoders can be constructed as neural networks that receive a sample $x$ as input, and output a value as close to $x$ as possible. When training such networks, it is possible to, for instance, feed samples with artificially-added noise and optimize the network to return the original sample (e.g., using a squared loss function), in order to learn de-noising filters. Alternatively, similar techniques can be used to learn image compression \cite{Balle17a}.

GANs \cite{gan} are a recent development in machine learning. They employ deep neural networks to generate realistic data (typically images) by simultaneously training two networks. The first (discriminator) network is trained to distinguish ``real'' dataset samples from ``fake'' generated samples, while the second (generator) network is trained to generate samples that are as similar to the real dataset as possible. 

The class of Reinforcement Learning (RL) utilizes DNNs \cite{li17surveydrl} for different purposes, such as defining policy functions and reward functions. Training algorithms for RL differ from supervised and unsupervised learning, using methods such as 
Deep Q Learning \cite{dqn} and A3C \cite{a3c}. These algorithms are out of the scope of this survey, but their parallelization techniques are similar.

\subsection{Parallel Computer Architecture}
\label{sec:term:pararch}

We continue with a brief overview of parallel hardware architectures
that are used to execute learning problems in practice. They can be
roughly classified into single-machine (often shared memory) and
multi-machine (often distributed memory) systems.

\subsubsection{Single-machine Parallelism}

Parallelism is ubiquitous in today's computer architecture, internally
on the chip in the form of pipelining and out-of-order execution as well
as exposed to the programmer in the form of multi-core or multi-socket
systems. Multi-core systems have a long tradition and can be programmed
with either multiple processes (different memory domains), multiple
threads (shared memory domains), or a mix of both. The main difference is
that multi-process parallel programming forces the programmer to
consider the distribution of the data as a first-class concern while
multi-threaded programming allows the programmer to only reason about
the parallelism, leaving the data shuffling to the hardware
system (often through hardware cache-coherence protocols). 

General-purpose CPUs have been optimized for general workloads ranging
from event-driven desktop applications to datacenter server tasks (e.g.,
serving web-pages and executing complex business workflows). Machine learning tasks are often compute intensive, making them similar to traditional high-performance computing (HPC) applications.
Thus, large learning workloads perform very well on accelerated systems
such as general purpose graphics processing units (GPU) or
field-programmable gate arrays (FPGA) that have been used in the
HPC field for more than a decade now.  Those devices focus on compute
throughput by specializing their architecture to utilize the high data
parallelism in HPC workloads. As we will see later, most
learning researchers utilize accelerators such as GPUs or FPGAs for
their computations. We emphasize that the main technique for
acceleration is to exploit the inherent parallelism in learning
workloads.

\begin{figure}[t]
	\centering
	\begin{subfigure}{.465\textwidth}
		\centering
		\includegraphics[height=1.5in]{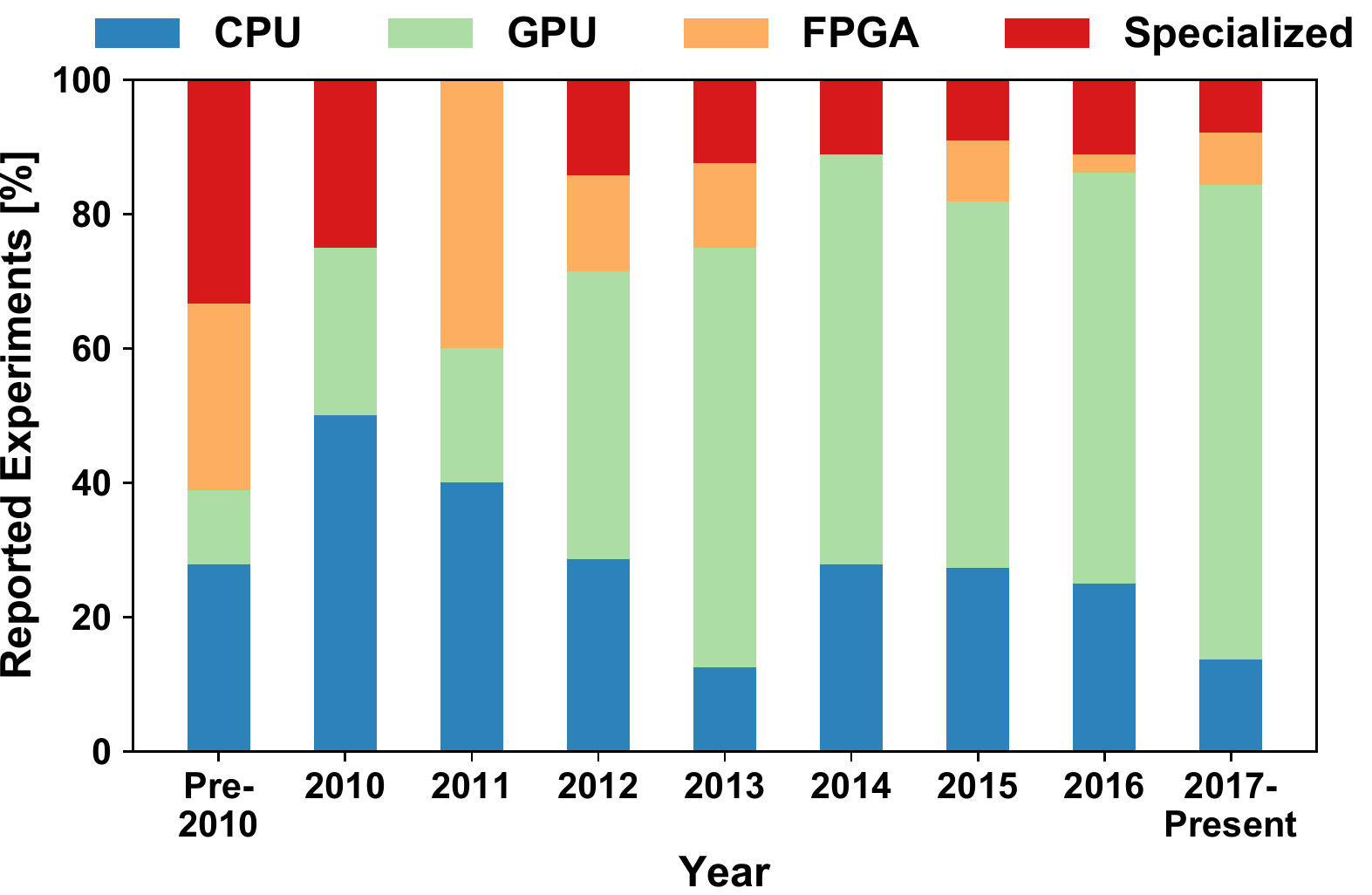}
		\caption{Hardware Architectures}
		\label{fig:devs}
		\centering
	\end{subfigure}
	\qquad
	\begin{subfigure}{.45\textwidth}
		\centering
		\includegraphics[height=1.5in]{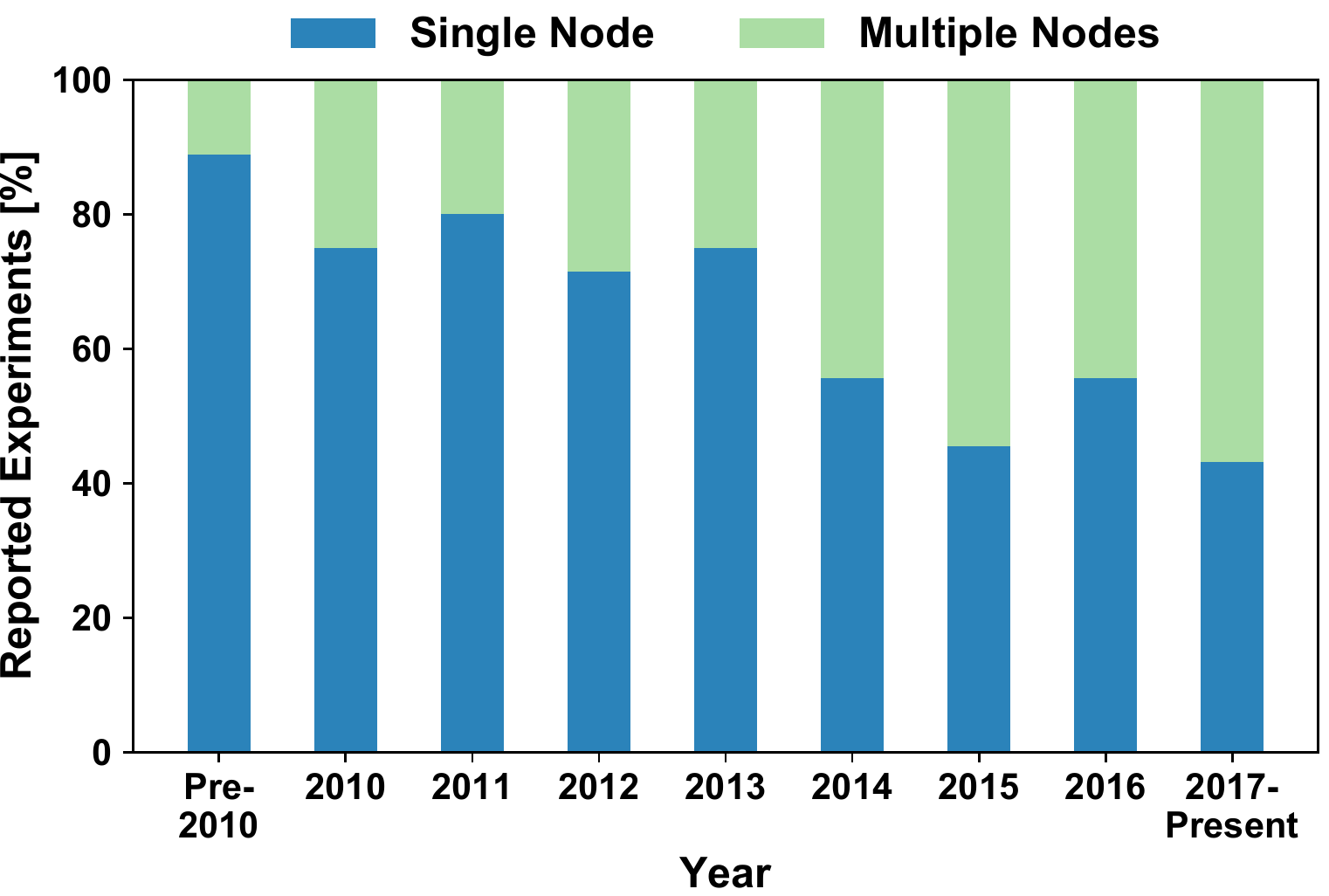}
		\caption{Training with Single vs. Multiple Nodes}
		\label{fig:nodes}
	\end{subfigure}
	\vspace{-1em}
	\caption{Parallel Architectures in Deep Learning}
	\vspace{-1.5em}
\end{figure}

Out of the 240 reviewed papers, 147 papers present
empirical results and provide details about their hardware setup. 
Fig. \ref{fig:devs} shows a summary of the machine architectures used in
research papers over the years. We see a clear trend towards GPUs, which
dominate the publications beginning from 2013. However, even accelerated 
nodes are not sufficient for the large computational workload. 
Fig. \ref{fig:nodes} illustrates the quickly growing multi-node parallelism
in those works. 
This shows that, beginning from 2015, distributed-memory architectures with
accelerators such as GPUs have become the default option for machine learning
at all scales today.

\subsubsection{Multi-machine Parallelism}

Training large-scale models is a very compute-intensive task.  Thus,
single machines are often not capable to finish this task in a desired
time-frame. To accelerate the computation further, it can
be distributed across multiple machines connected by a network. The most
important metrics for the interconnection network (short: interconnect)
are latency, bandwidth, and message-rate. Different network technologies
provide different performance. For example, both modern Ethernet and
InfiniBand provide high bandwidth but InfiniBand has significantly lower
latencies and higher message rates. Special-purpose HPC interconnection
networks can achieve higher performance in all three metrics. Yet,
network communication remains generally slower than intra-machine
communication. 

\begin{figure}[t]
	\centering
	\begin{subfigure}{.485\textwidth}
		\centering
		\includegraphics[height=1.5in]{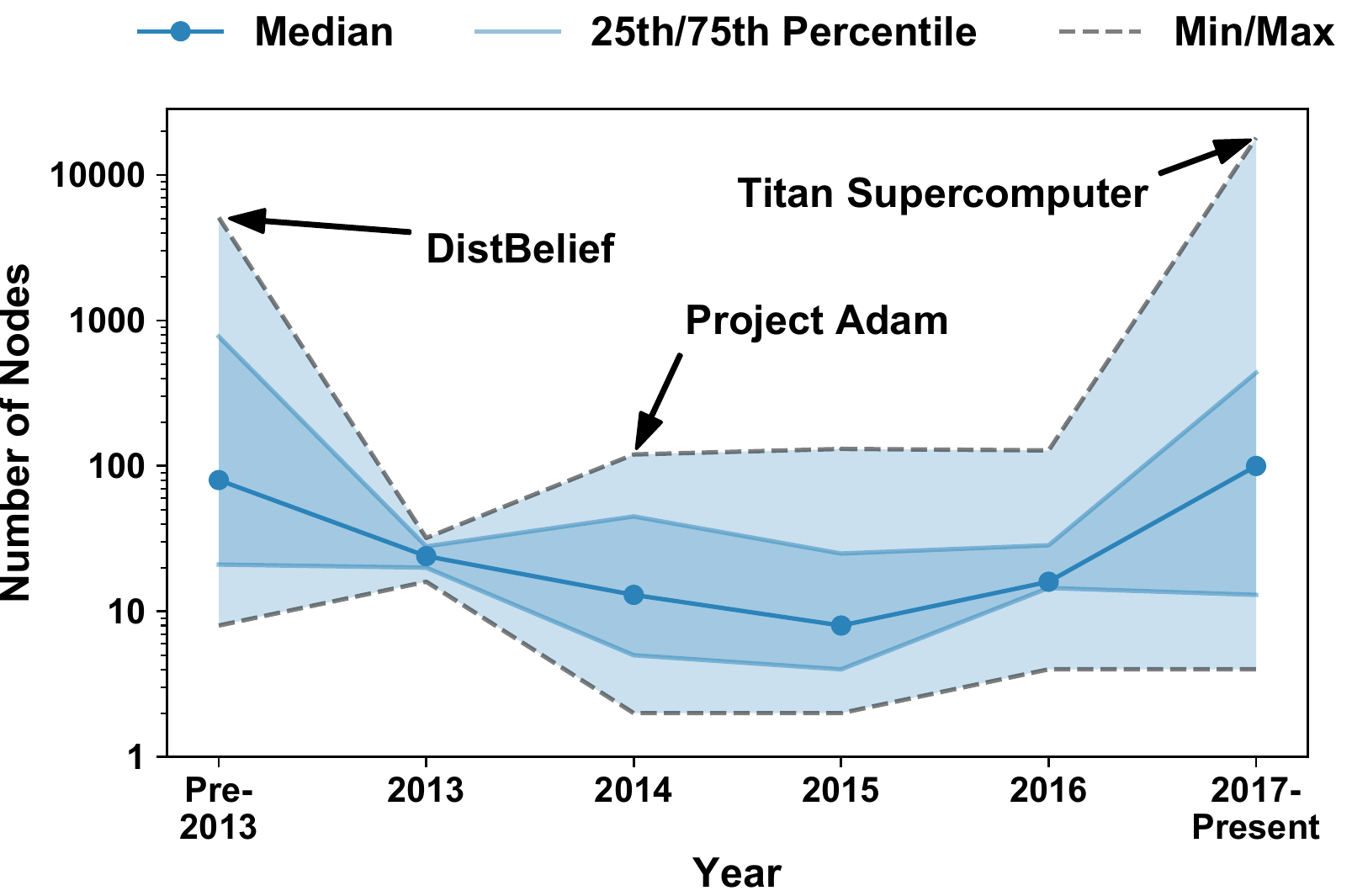}
		\caption{Node Count}
		\label{fig:char:number}
	\end{subfigure}
	\quad
	\begin{subfigure}{.46\textwidth}
		\centering
		\includegraphics[height=1.5in]{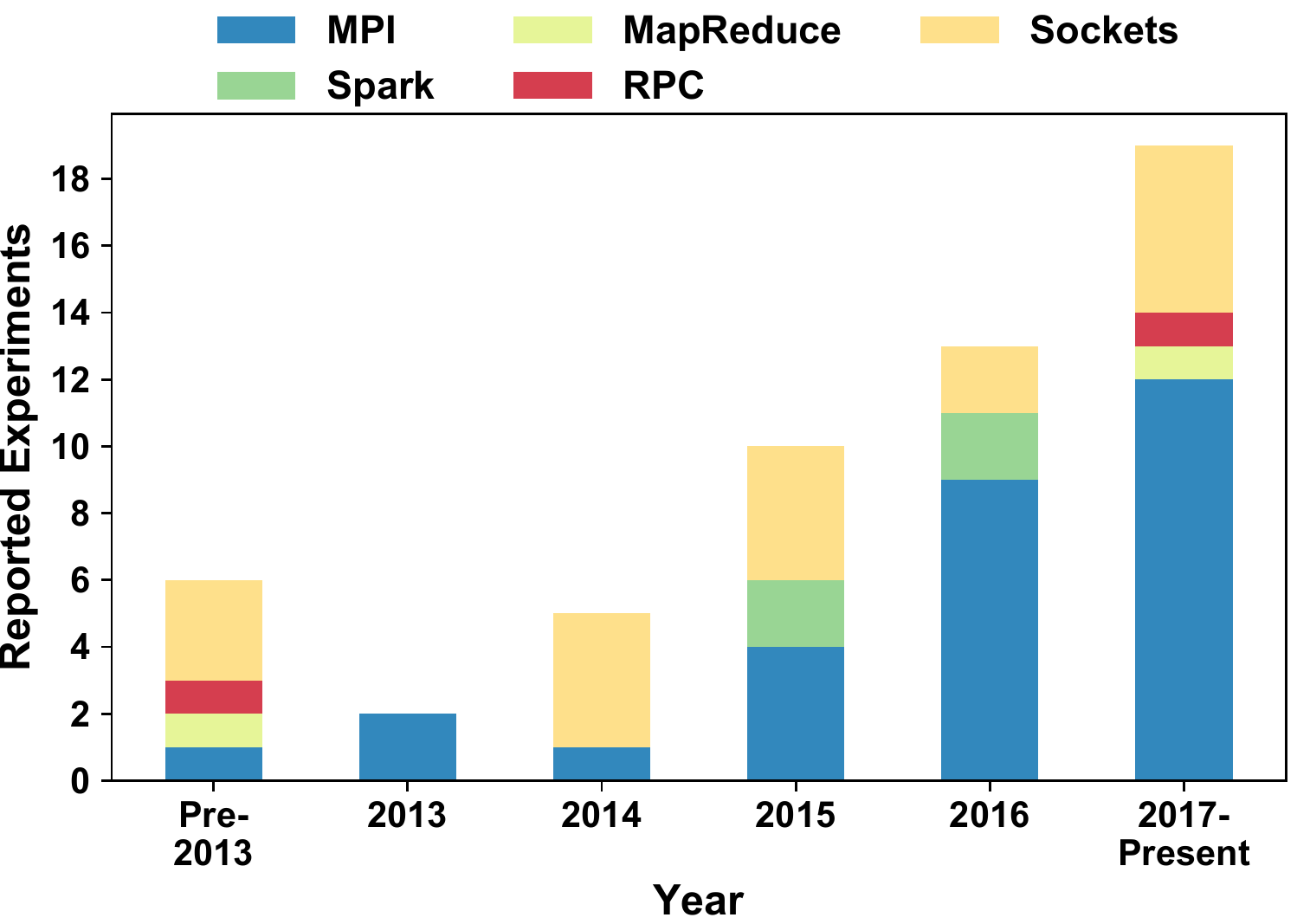}
		\caption{Communication Layer}
		\label{fig:char:commlayer}
	\end{subfigure}
	\vspace{-1em}
	\caption{Characteristics of Deep Learning Clusters}
	\vspace{-1.5em}
	\label{fig:char}
\end{figure}

Fig.~\ref{fig:char:number} shows a breakdown of the number of nodes used in deep
learning research over the years. It started very high with the large-scale
DistBelief run, reduced slightly with the introduction of powerful accelerators
and is on a quick rise again since 2015 with the advent of large-scale deep
learning.
Out of the 240 reviewed papers, 73 make use of distributed-memory systems and
provide details about their hardware setup. We observe that large-scale setups,
similar to HPC machines, are commonplace and essential in today's training. 

\subsection{Parallel Programming}
\label{sec:term:parprog}

Programming techniques to implement parallel learning algorithms on
parallel computers depend on the target architecture. They range from
simple threaded implementations to OpenMP on single machines.
Accelerators are usually programmed with special languages such as
NVIDIA's CUDA, OpenCL, or in the case of FPGAs using hardware design
languages. Yet, the details are often hidden behind library calls
(e.g., cuDNN or MKL-DNN) that implement the time-consuming primitives.

On multiple machines with distributed memory, one can either use simple
communication mechanisms such as TCP/IP or Remote Direct Memory Access
(RDMA). On distributed memory machines, one can also use more
convenient libraries such as the Message Passing Interface (MPI) or
Apache Spark. MPI is a low level library focused on providing portable
performance while Spark is a higher-level framework that focuses more
on programmer productivity.

Fig.~\ref{fig:char:commlayer} shows a breakdown of the different
communication mechanisms that were specified in 55 of the 73
papers using multi-node parallelism.
It shows how the community quickly recognized that deep learning has very
similar characteristics than large-scale HPC applications. Thus, beginning from
2016, the established MPI interface became the de-facto portable communication
standard in distributed deep learning.

\subsection{Parallel Algorithms}
\label{sec:term:paralg}

We now briefly discuss some key concepts in parallel computing that are
needed to understand parallel machine learning. Every computation on a
computer can be modeled as a directed acyclic graph (DAG). The vertices
of the DAG are the computations and the edges are the data dependencies
(or data flow). The computational parallelism in such a graph can be
characterized by two main parameters: the graph's work $\mathbf{W}$, which
corresponds to the total number of vertices, and the graph's depth $\mathbf{D}$,
which is the number of vertices on any longest path in the DAG. These
two parameters allow us to characterize the computational complexity on
a parallel system. For example, assuming we can process one operation
per time unit, then the time needed to process the graph on a single
processor is $T_1 = \mathbf{W}$ and the time needed to process the graph on an
infinite number of processes is $T_\infty = \mathbf{D}$.  The average parallelism
in the computation is $\mathbf{W}/\mathbf{D}$, which is often a \emph{good} number of
processes to execute the graph with. Furthermore, we can show that the
execution time of such a DAG on $p$ processors is bounded by:
$\min\{\mathbf{W}/p, \mathbf{D}\} \leq T_p \leq
\mathcal{O}(\mathbf{W}/p+\mathbf{D})$~\cite{Brent:1974:PEG:321812.321815,Arora:1998:TSM:277651.277678}.

Most of the operations in learning can be modeled as operations on
tensors (typically tensors as a parallel programming model~\cite{1512.00066}).
Such operations are highly data-parallel and only summations
introduce dependencies. Thus, we will focus
on parallel reduction operations in the following. 

\begin{figure}[t]
	\centering
	\begin{minipage}{.45\textwidth}
		\centering
		\begin{subfigure}[t]{\linewidth}
			\centering
			\includegraphics[width=1in]{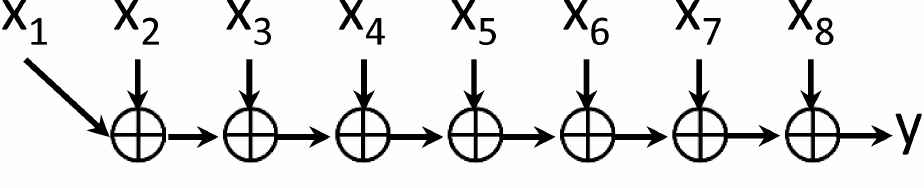}
			\caption{Linear-Depth Reduction}
			\label{fig:linred}
		\end{subfigure}\vspace{0.1in} \\
		\begin{subfigure}[t]{\linewidth}
			\centering
			\includegraphics[width=1in]{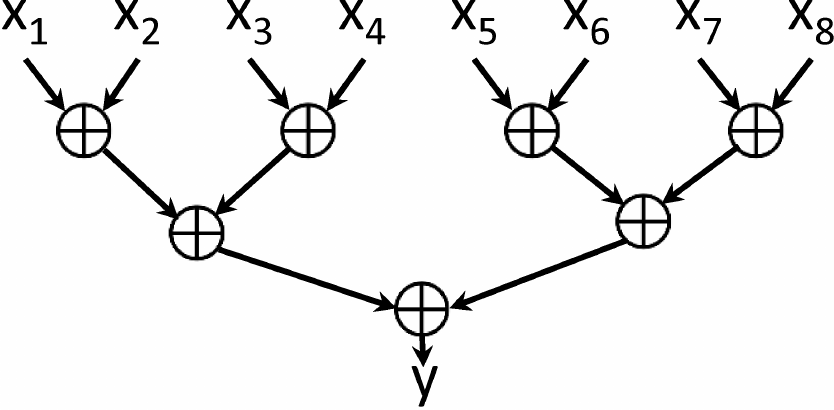}
			\caption{Tree Reduction}
			\label{fig:treered}
		\end{subfigure}
	\end{minipage}%
	\begin{minipage}{.45\textwidth}
		\centering
		\begin{subfigure}[t]{\linewidth}
			\centering
			\includegraphics[height=1.05in,trim={8cm 11cm 18cm 2.25cm},clip]{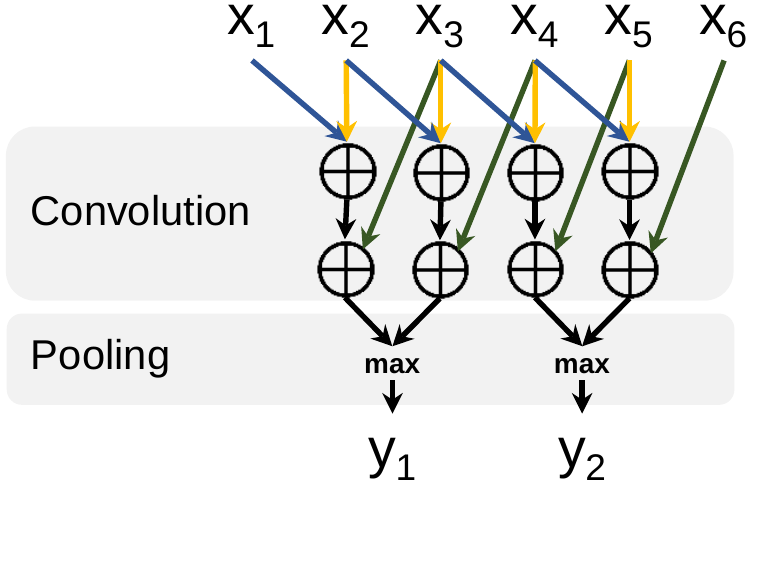}
			\caption{Convolution Downscaling in DNNs}
			\label{fig:convcomm}
		\end{subfigure}
	\end{minipage}
	\vspace{-1em}
	\caption{Reduction Schemes}
	\vspace{-2em}
\end{figure}

In a reduction, we apply a series of binary operators $\oplus$ to
combine $n$ values into a single value, e.g., $y=x_1 \oplus x_2 \oplus
x_3 \cdots \oplus x_{n-1} \oplus x_{n}$. If the operation $\oplus$ is
associative then we can change its application, which changes the DAG
from a linear-depth line-like graph as shown in Fig.~\ref{fig:linred}
to a logarithmic-depth tree graph
as shown in Fig.~\ref{fig:treered}. It is simple to show that the work and depth
for reducing $n$ numbers is $\mathbf{W}=n-1$ and $\mathbf{D}=\lceil\log_2 n\rceil$, respectively. In
deep learning, one often needs to reduce (sum) large tables of $m$
independent parameters and return the result to all processes. This is
called \emph{allreduce} in the MPI
specification~\cite{mpi-3.1,UsingAdvancedMPI}.

In multi-machine environments, these tables are distributed across the
machines which participate in the overall reduction operation. Due to
the relatively low bandwidth between the machines (compared to local
memory bandwidths), this operation is often most critical for
distributed learning. We analyze the algorithms in a simplified LogP
model~\cite{Culler:1993:LTR:155332.155333}, where we ignore injection
rate limitations ($o=g=0$), which makes it similar to the simple
$\alpha$-$\beta$ model: $L=\alpha$ models the point-to-point latency in
the network, $G=\beta$ models the cost per byte, and $P\leq p$ is the
number of networked machines. Based on the DAG model from above, it is
simple to show a lower bound for the reduction time $T_r \geq L
\log_2(P)$ in this simplified model. Furthermore, because each element
of the table has to be sent at least once, the second lower bound is
$T_r \geq \gamma m G$, where $\gamma$ represents the size of a single
data value and $m$ is the number of values sent.  This bound can be
strengthened to $T_r \geq L \log_2(P) + 2\gamma mG(P-1)/P$ if we
disallow redundant computations~\cite{Chan:2007:CCT:1285358.1285359}. 

Several practical algorithms exist for the parallel allreduce operation
in different environments and the best algorithm depends on the system,
the number of processes, and the message size. We refer to Chan et
al.~\cite{Chan:2007:CCT:1285358.1285359} and Hoefler and
Moor~\cite{hoefler-moor-collectives} for surveys of collective
algorithms. Here, we summarize key algorithms that have been
rediscovered in the context of parallel learning. The simplest algorithm
is to combine two trees, one for summing the values to one process,
similar to Fig.~\ref{fig:treered}, and one for broadcasting the values
back to all processes; its complexity is $T_{\mathit{tree}}=2\log_2(P)(L + \gamma
mG)$. Yet, this algorithm is inefficient and can be optimized with a
simple butterfly pattern, reducing the time to $T_{\mathit{bfly}}=\log_2(P)(L +
\gamma mG)$. The butterfly algorithm is efficient (near-optimal) for
small $\gamma m$. For large $\gamma m$ and small $P$, a simple linear
pipeline that splits the message into $P$ segments is bandwidth-optimal
and performs well in practice, even though it has a linear component in
$P$: $T_{\mathit{pipe}}=2(P-1)(L+\gamma \frac{m}{P}G)$.  For most ranges
of $\gamma m$ and $P$, one could use Rabenseifner's
algorithm~\cite{rabenseifner2004optimization}, which combines
reduce-scatter with gather, running in time
$T_{\mathit{rabe}}=2L\log_2(P) + 2\gamma mG(P-1)/P$. This algorithm
achieves the lower bound but may be harder to implement and tune.

Other communication problems needed for convolutions and pooling, illustrated
in Fig.~\ref{fig:convcomm}, exhibit high spatial locality due to strict
neighbor interactions. They can be optimized using well-known HPC techniques
for stencil computations such as MPI Neighborhood
Collectives~\cite{hoefler-neighborcolls-optimization} (formerly known as sparse
collectives~\cite{hoefler-sparsecolls-hips}) or optimized Remote Memory Access
programming~\cite{notified-access}. In general, exploring different low-level
communication, message scheduling, and topology mapping~\cite{hoefler-topomap}
strategies that are well-known in the HPC field could significantly speed up
the communication in distributed deep learning.

\section{The Efficiency Tradeoff: Generalization vs. Utilization}
\label{sec:tradeoff}

\begin{figure}[t]
	\centering
	\begin{subfigure}[t]{0.45\textwidth}
		\centering
		\includegraphics[height=1.1in,clip,trim={0cm 9.5cm 20.5cm 0cm}]{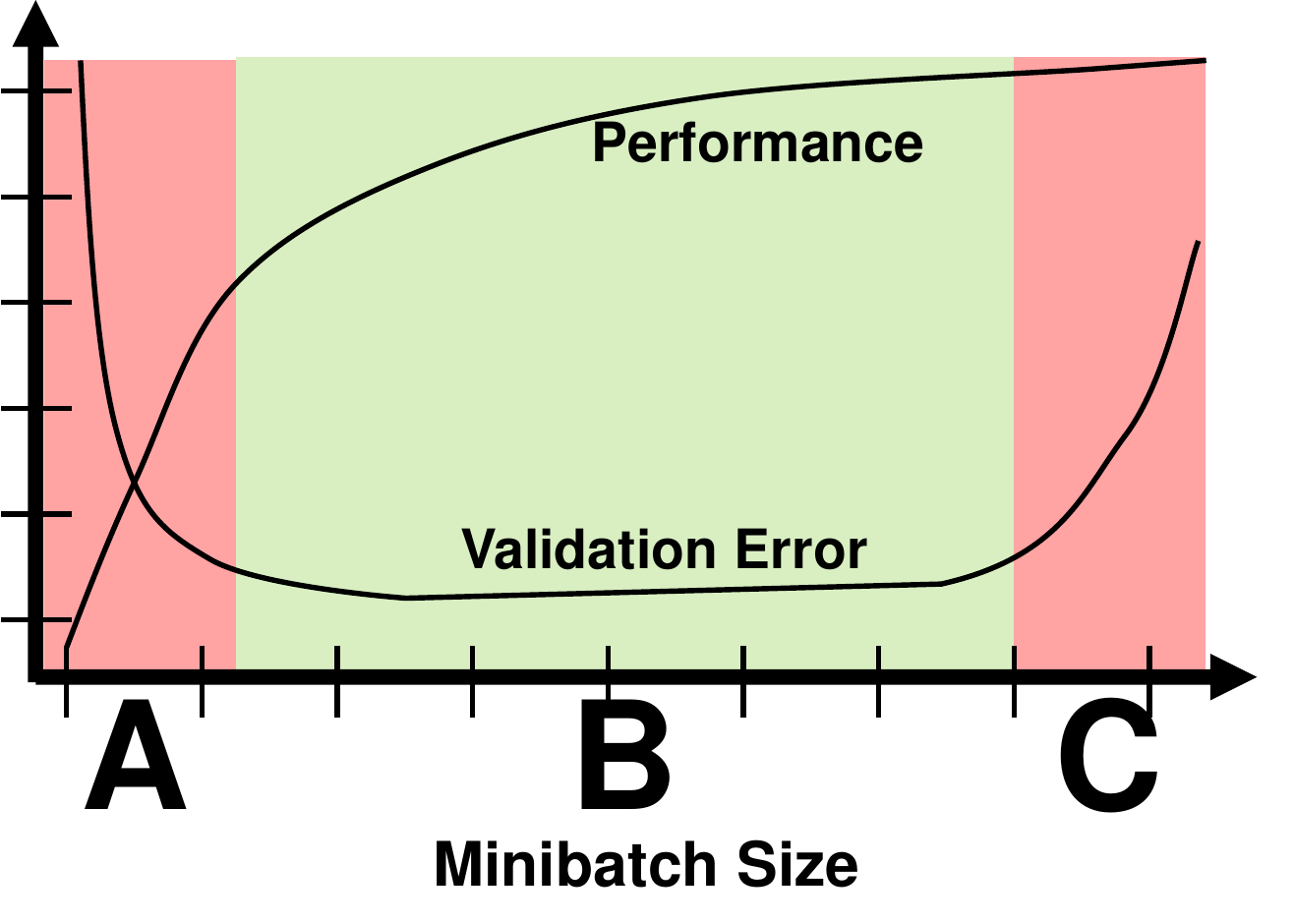}
		\caption{Performance and accuracy of minibatch SGD after a fixed number of epochs (Illustration).}
		\label{fig:minibatch:perf}
	\end{subfigure}
	\qquad
	\begin{subfigure}[t]{0.45\textwidth}
		\centering
		\includegraphics[height=1.1in]{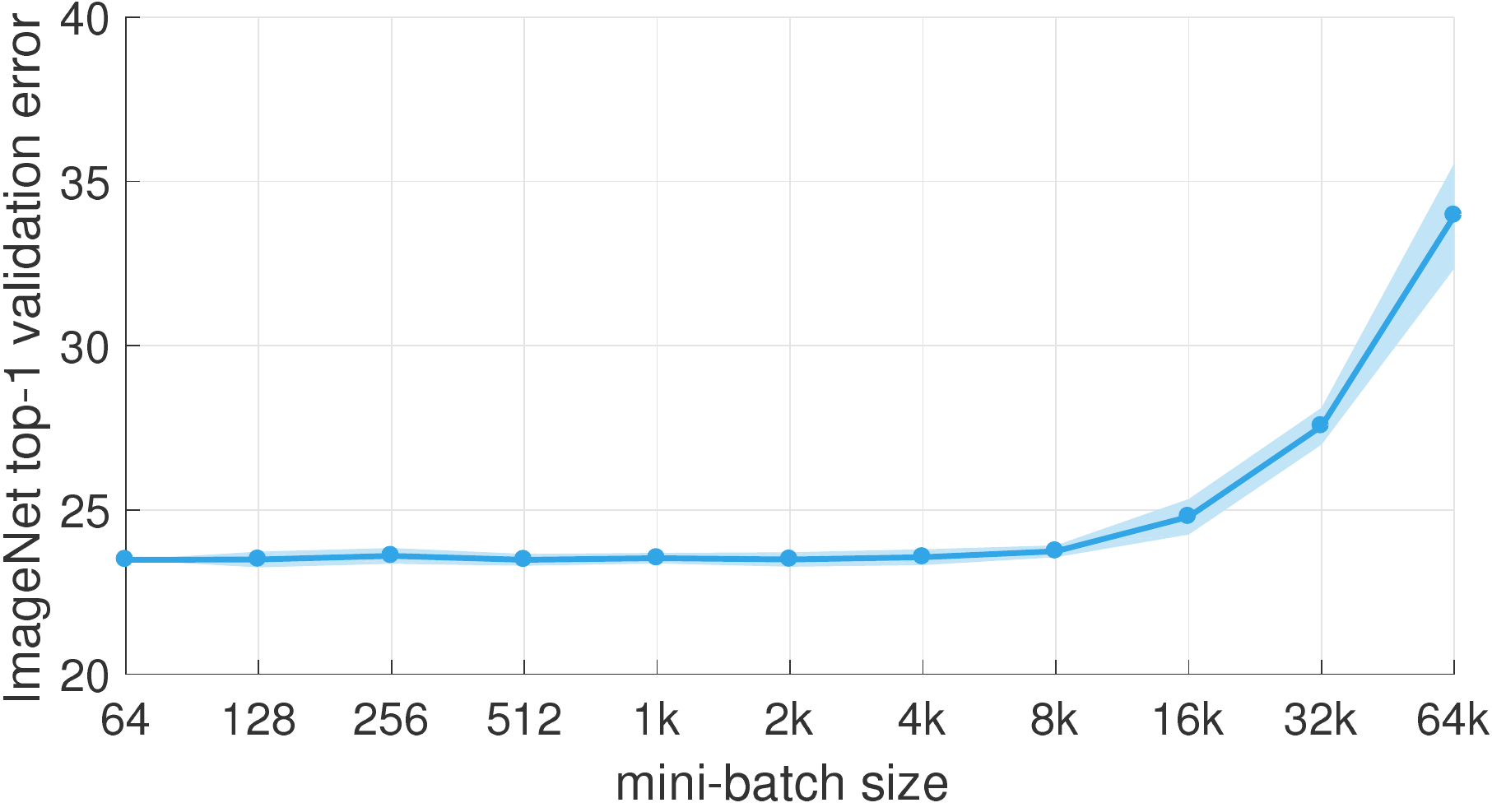}
		\caption{Empirical accuracy (ResNet-50, figure adapted from \cite{goyal17}, lower is better).}
		\label{fig:minibatch:acc}
	\end{subfigure}
	\vspace{-1em}
	\caption{Minibatch Size Effect on Accuracy and Performance}
	\label{fig:minibatch}
	\vspace{-1.25em}
\end{figure}

In the previous section, we mentioned that SGD can be executed  concurrently through the use of minibatches. However, setting the minibatch size is a complex optimization space on its own merit, as it affects both statistical accuracy (generalization) and hardware efficiency (utilization) of the model. As illustrated in Fig. \ref{fig:minibatch:perf}, minibatches should not be too small (region A), so as to harness inherent concurrency in evaluation of the loss function; nor should they be too large (region C), as the quality of the result decays once increased beyond a certain point. 

We can show the existence of region C by combining SGD with the descent lemma for a function $f$ with $L$-Lipschitz gradient: 
$
\e_z\left[f(w^{(t+1)})\right]\le 
f(w^{(t)}) -
\eta_t\left\|\nabla f(w^{(t)})\right\|^2 + 
\eta^2_t\frac{L}{2}\e_z\left[\left\|\nabla f_z(w^{(t)})\right\|^2\right]
$, where $z\sim \mathcal{D}$ and $\nabla f_z$ is the stochastic subgradient for $z$. This indicates that a large minibatch (with adjusted learning rate) can increase the convergence rate (negative term), but along with it the gradient variance and learning rate, which causes the last term to hinder convergence.
 
Indeed, the illustrated behavior is empirically shown for larger minibatch sizes in Fig. \ref{fig:minibatch:acc}, and typical sizes lie between the orders of 10 and 10,000. Also, large-batch methods only converge and generalize when: (a) learning rates are adjusted statically \cite{krizhevsky2014one,goyal17} or adaptively \cite{b32k}; (b) using a ``warmup'' phase \cite{goyal17}; (c) using the batch size to control gradient variance \cite{friedlander11}; (d) adaptively increasing minibatch size during training \cite{batchsize}; or (e) when using specific learning rate schedules \cite{sgdr}. Overall, such works \textit{increase} the upper bound on feasible minibatch sizes, but do not \textit{remove} it.

\section{Deep Neural Networks}
\label{sec:dnn}

\begin{figure}[b]
	\centering
	\begin{subfigure}[t]{0.27\textwidth}
		\centering
		\includegraphics[height=1.2in,trim={4.5cm 7cm 16cm 0.5cm},clip]{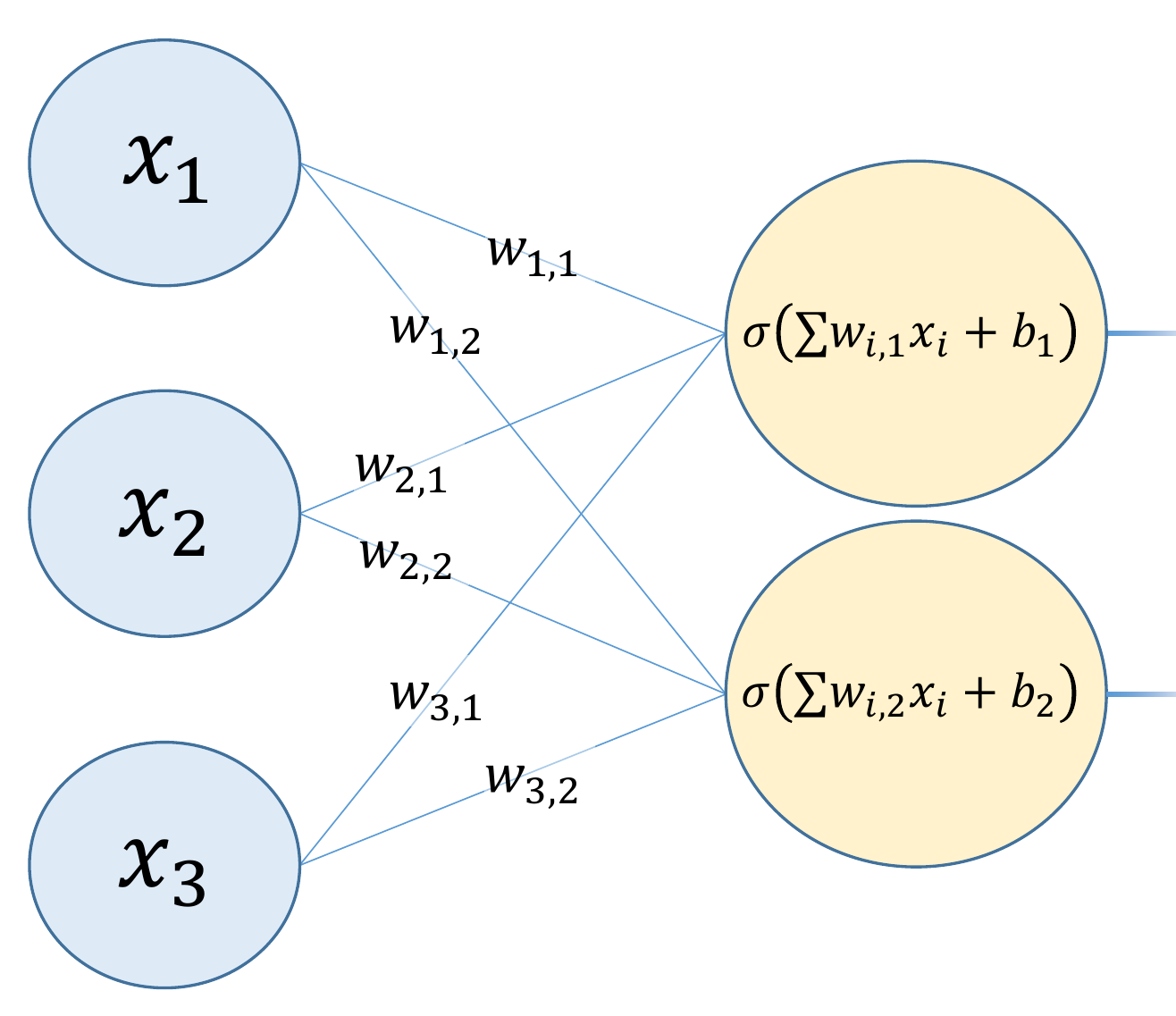}
		\caption{Neural Network Operator}
		\label{fig:layer}
	\end{subfigure}
	\qquad
	\begin{subfigure}[t]{0.53\textwidth}
		\centering
		\includegraphics[height=1.2in]{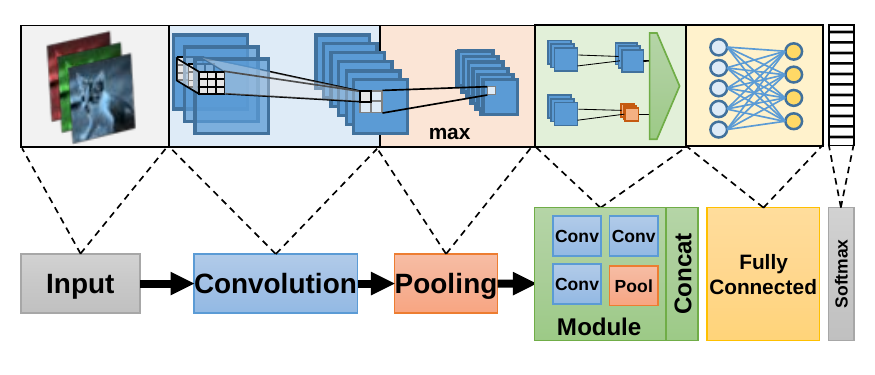}
		\caption{Deep Network}
		\label{fig:dag}
	\end{subfigure}
	\vspace{-1em}
	\caption{Deep Neural Network Architecture}
	\label{fig:dnn}
\end{figure}

We now describe the anatomy of a Deep Neural Network (DNN). In Fig. \ref{fig:dnn}, we see a DNN in two scales: the single operator (Fig. \ref{fig:layer}, also ambiguously called layer) and the composition of such operators in a layered deep network (Fig. \ref{fig:dag}). In the rest of this section, we describe popular operator types and their properties, followed by the computational description of deep networks and the backpropagation algorithm. Then, we study several examples of popular neural networks, highlighting the computational trends driven by their definition.

\subsection{Neurons}
\label{sec:dnn:layers}

The basic building block of a deep neural network is the \textit{neuron}. Modeled after the brain, an artificial neuron (Fig. \ref{fig:layer}) accumulates signals from other neurons connected by \textit{synapses}. An \textit{activation function} (or \textit{axon}) is applied on the accumulated value, which adds nonlinearity to the network and determines the signal this neuron ``fires'' to its neighbors. In \textit{feed-forward neural networks}, the neurons are grouped to \textit{layers} strictly connected to neurons in subsequent layers. In contrast, \textit{recurrent neural networks} allow back-connections within the same layer.

\subsubsection{Feed-Forward Operators}

\begin{figure}[t]
	\centering
	\begin{subtable}[b]{0.45\linewidth}
		\centering
		\tiny
		\begin{tabular}{ l p{0.7\linewidth} }
			\toprule
			\bf Name & \bf Description \\\midrule
			$N$& Minibatch size\\
			$C$& Number of channels, features, or neurons\\
			$H$& Image Height\\
			$W$& Image Width\\
			$K_x$ & Convolution kernel width\\
			$K_y$ & Convolution kernel height\\		
			\bottomrule
		\end{tabular}
		\caption{Data Dimensions}
		\label{tbl:dims}
	\end{subtable}
	\qquad
	\begin{subfigure}[b]{0.45\linewidth}
		\centering
		\includegraphics[height=0.9in,trim={0cm 11cm 17cm 0cm},clip]{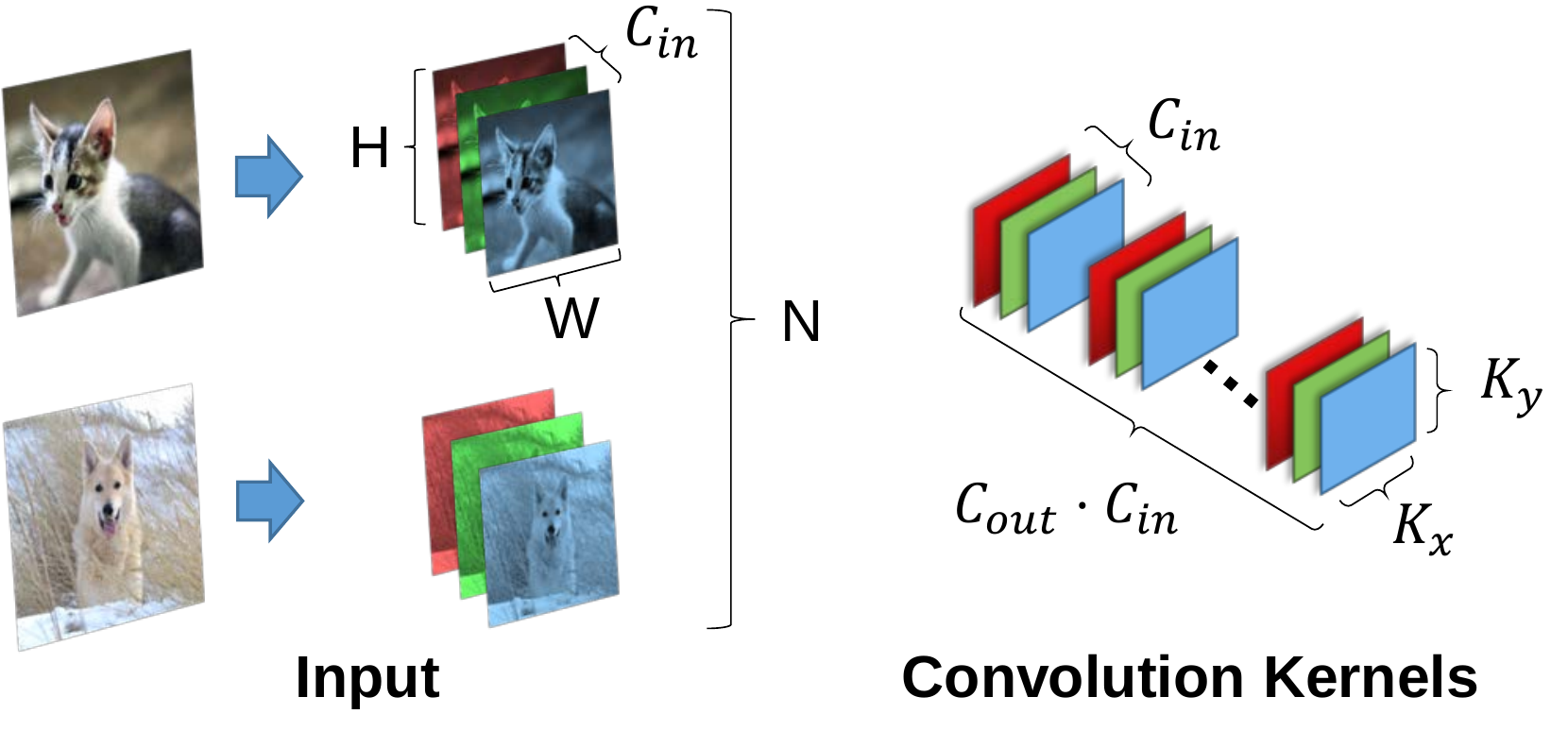}
		\caption{Convolution Dimensions}
	\end{subfigure}
	\vspace{-1em}
	\caption{Summary of Data Dimensions in Operators}
	\label{fig:dnn:tbls}
	\vspace{-1em}
\end{figure}

Neural network operators are implemented as weighted sums, using the synapses as weights. Activations (denoted $\sigma$) can be implemented as different functions, such as Sigmoid, Softmax, hyperbolic tangents, Rectified Linear Units (ReLU), or variants thereof \cite{prelu}. When color images are used as input (as is commonly the case in computer vision), they are usually represented as a 4-dimensional tensor sized $N\times$$C\times$$H\times$$W$. As shown in Fig. \ref{fig:dnn:tbls}, $N$ is number of images in the minibatch, where each $H\times$$W$ image contains $C$ channels (e.g., image RGB components). If an operator disregards spatial locality in the image tensor (e.g., a fully connected layer), the dimensions are flattened to $N\times(C\cdot H\cdot W)$.
In typical DNN and CNN constructions, the number of features (channels in subsequent layers), as well as the width and height of an image, change from layer to layer using the operators defined below. We denote the input and output features of a layer by $C_{in}$ and $C_{out}$ respectively. 

A fully connected layer (Fig. \ref{fig:layer}) is defined on a group of neurons $x$ (sized $N\times C_{in}$, disregarding spatial properties) by $y_{i,*} = \sigma(wx_{i,*}+b)$, where $w$ is the weight matrix (sized $C_{in}\times C_{out}$) and $b$ is a per-layer trainable bias vector (sized $C_{out}$). While this inner product is usually implemented with multiplication and addition, some works use other operators, such as similarity \cite{simnets}.

Not all operators in a neural network are fully connected. Sparsely connecting neurons and sharing weights is beneficial for reducing the number of parameters; as is the case in the popular \textit{convolutional} operator. In a convolutional operator, every 3D tensor $x$ (i.e., a slice of the 4D minibatch tensor representing one image) is convolved with $C_{out}$ kernels of size $C_{in}\times$$K_y\times$$K_x$, where the base formula for a minibatch is given by:
\small
\begin{equation}\label{eq:conv}
y_{i,j,k,l}=\sum_{m=0}^{C_{in}-1}{\sum_{k_y=0}^{K_y-1}{\sum_{k_x=0}^{K_x-1}{x_{i,m,k+k_y,l+k_x} \cdot w_{j,m,k_y,k_x}}}},
\end{equation}
\normalsize
where $y$'s dimensions are $N\times C_{out}\times H'\times W'$, $H'=H-K_y+1$, and $W'=W-K_x+1$, accounting for the size after the convolution, which does not consider cases where the kernel is out of the image bounds. Note that the formula omits various extensions of the operator \cite{dlconv}, such as variable stride, padding, and dilation \cite{dilated}, each of which modifies the accessed indices and $H',W'$. The two inner loops of Eq. \ref{eq:conv} are called the \textit{convolution kernel}, and the kernel (or filter) size is $K_x\times K_y$.

While convolutional operators are the most computationally demanding in CNNs, other operator types are prominently used in networks. Two such operators are \textit{pooling} and \textit{batch normalization}. The former reduces an input tensor in the width and height dimensions, performing an operation on contiguous sub-regions of the reduced dimensions, such as maximum (called max-pooling) or average, and is given by:
\small
\begin{equation*}
y_{i,j,k,l}= \max_{k_x\in [0,K_x),k_y\in [0,K_y)}{x_{i, j, k+k_x, l+k_y}}.
\end{equation*}
\normalsize
The goal of this operator is to reduce the size of a tensor by sub-sampling it while emphasizing important features. Applying subsequent convolutions of the same kernel size on a sub-sampled tensor enables learning high-level features that correspond to larger regions in the original data.

Batch Normalization (BN) \cite{bn} is an example of an operator that creates inter-dependencies between samples in the same minibatch. Its role is to center the samples around a zero mean and a variance of one, which, according to the authors, reduces the internal covariate shift. BN is given by the following transformation:
\small
\begin{equation*}
y_{i,j,k,l} = \left(\frac{x_{i,j,k,l} - \e\left[x_{*,j,k,l}\right]}{\sqrt{\Var\left[x_{*,j,k,l}\right]+\epsilon}}\right)\cdot \gamma + \beta,
\end{equation*}
\normalsize
where $\gamma,\beta$ are scaling factors, and $\epsilon$ is added to the denominator for numerical stability.

\subsubsection{Recurrent Operators}
Recurrent Neural Networks (RNNs) \cite{rnn} enable connections from a layer's output to its own inputs. These connections create ``state'' in the neurons, retaining persistent information in the network and allowing it to process data sequences instead of a single tensor. We denote the input tensor at time point $t$ as $x^{(t)}$.

\begin{figure}[t]
	\centering
	\begin{subfigure}[t]{0.27\textwidth}
		\centering
		\includegraphics[page=1,height=1in,trim={7cm 2.25cm 10cm 4.5cm},clip]{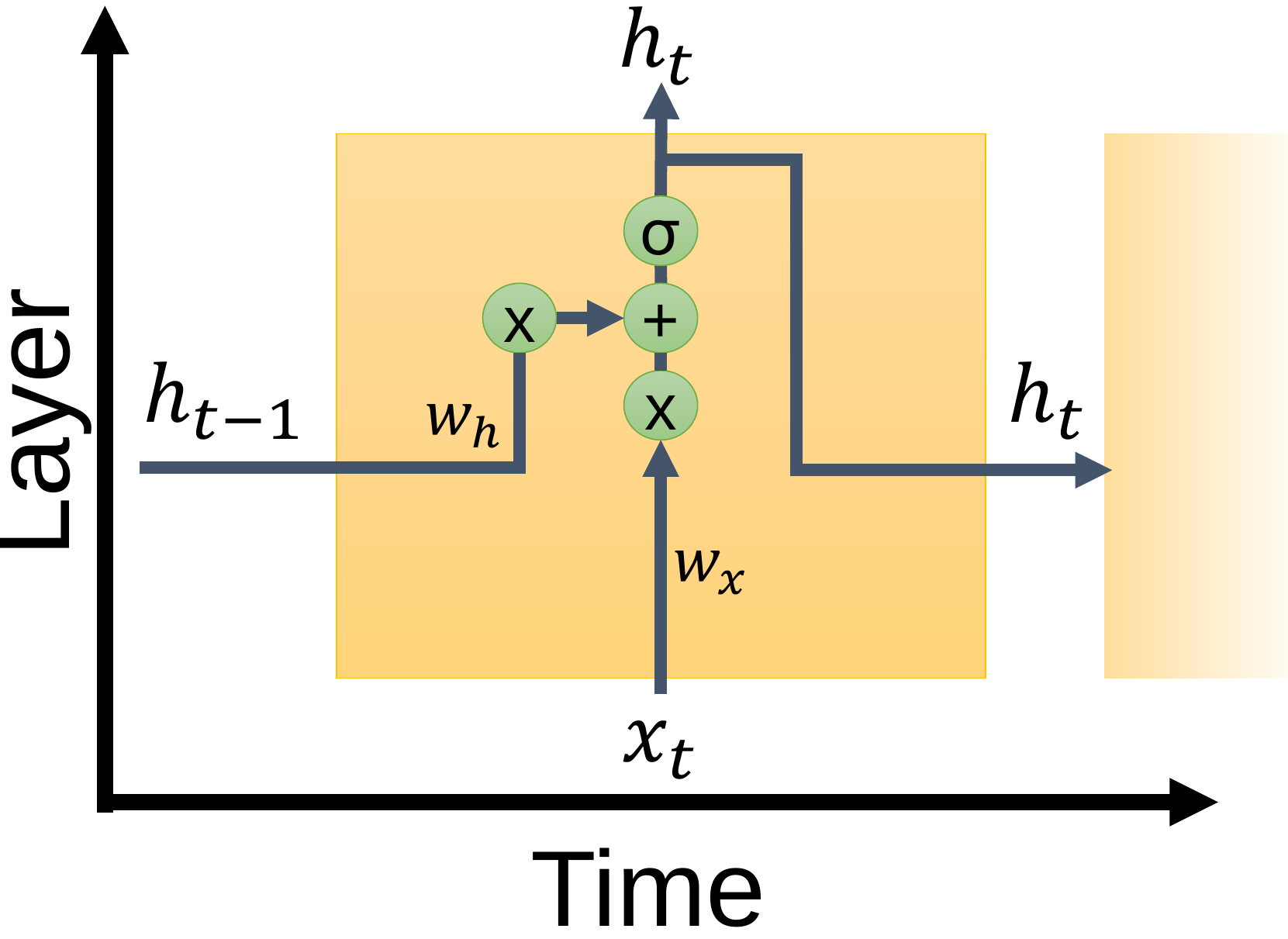}
		\caption{Recurrent Units}
		\label{fig:rnn:rnn}
	\end{subfigure}
	\qquad
	\begin{subfigure}[t]{0.27\textwidth}
		\centering
		\includegraphics[page=2,height=1in,trim={7cm 2.25cm 10cm 4.5cm},clip]{figures/rnn}
		\caption{Long Short-Term Memory}
		\label{fig:rnn:lstm}
	\end{subfigure}
	\qquad
	\begin{subfigure}[t]{0.27\textwidth}
		\centering
		\includegraphics[page=3,height=1in,trim={7cm 2.25cm 10cm 4.5cm},clip]{figures/rnn}
		\caption{Gated Recurrent Unit}
		\label{fig:rnn:gru}
	\end{subfigure}
	\vspace{-1em}
	\caption{Recurrent Neural Network (RNN) Layers. Sub-figures (b) and (c) adapted from \cite{lstmblog}.}
	\label{fig:rnn}
	\vspace{-1em}
\end{figure}

The standard Elman RNN layer is defined as $y^{(t)}=w_y\cdot \left(w_h\cdot h_{t-1} + w_x\cdot x^{(t)}\right)$ (omitting bias, illustrated in Fig. \ref{fig:rnn:rnn}), where $h_t$ represents the ``hidden'' data at time-point $t$ and is carried over to the next time-point.
Despite the initial success of these operators, it was found that they tend to ``forget'' information quickly (as a function of sequence length) \cite{bengio94}. To address this issue, Long-Short Term Memory (LSTM) \cite{lstm} (Fig. \ref{fig:rnn:lstm}) units redesign the structure of the recurrent connection to resemble memory cells.
Several variants of LSTM exist, such as the Gated Recurrent Unit (GRU) \cite{gru} (Fig. \ref{fig:rnn:gru}), which simplifies the LSTM gates to reduce the number of parameters.

\subsection{Deep Networks}
\label{sec:dnn:bprop}

According to the definition of a fully connected layer, the expressiveness of a ``shallow'' neural network is limited to a separating hyperplane, skewed by the nonlinear activation function. When composing layers one after another, we create deep networks (as shown in Fig. \ref{fig:dag}) that can approximate arbitrarily complex continuous functions. While the exact class of expressible functions is currently an open problem, results \cite{delalleau11,cohen16expressive} show that neural network depth can reduce breadth requirements exponentially with each additional layer.

A Deep Neural Network (DNN) can be represented as a function composition, e.g., $\ell(L_M(w_M,\cdots$ $L_2(w_2, L_1(w_1, x))))$, where each function $L_i$ is an operator, and each vector $w_i$ represents operator $i$'s weights (parameters). In addition to direct composition, a DNN DAG might reuse the output values of a layer in multiple subsequent layers, forming \textit{shortcut} connections \cite{resnet,densenet}.

\begin{figure}[b]
	\centering		
	\vspace{-0.8em}
	\includegraphics[height=1in, trim={0cm 12.5cm 12.25cm 0cm},clip]{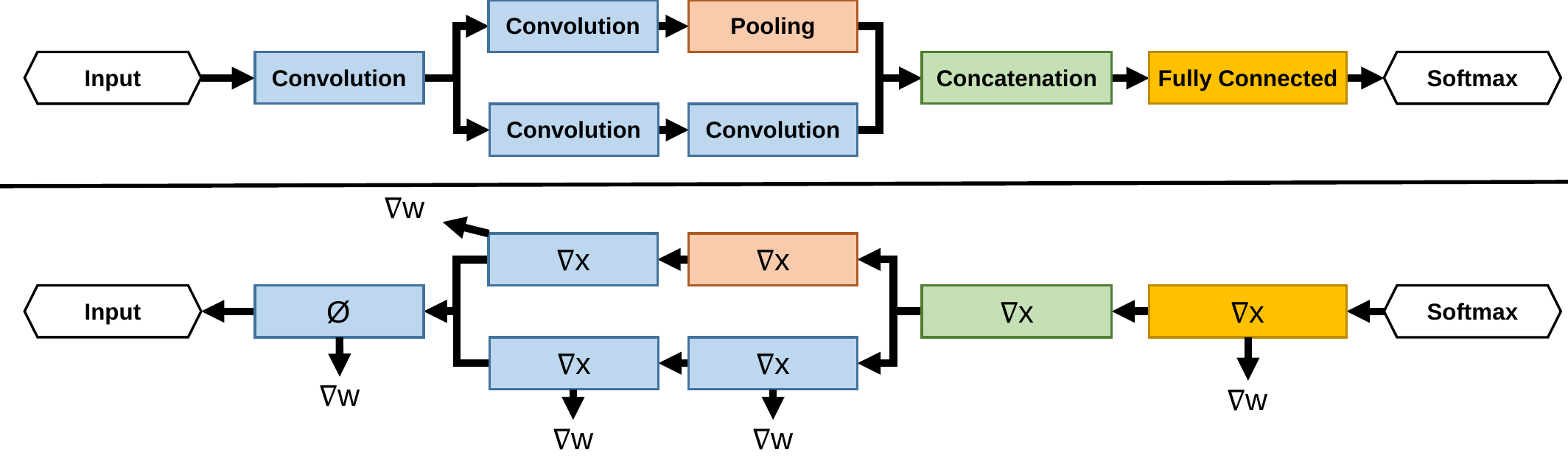}
	\caption{The Backpropagation Algorithm}
	\label{fig:backprop}
\end{figure}

Computation of the DNN loss gradient $\nabla \ell$, which is necessary for SGD, can be performed by repeatedly applying the chain rule in a process commonly referred to as \textit{backpropagation}.
As shown in Fig. \ref{fig:backprop}, the process of obtaining $\nabla\ell(w,x)$ is performed in two steps. First, $\ell(w,x)$ is computed by forward evaluation (top portion of the figure), computing each layer of operators after its dependencies in a topological ordering. After computing the loss, information is propagated backward through the network (bottom portion of the figure), computing two gradients --- $\nabla x$ (w.r.t. input data), and $\nabla w_i$ (w.r.t. layer weights). Note that some operators do not maintain mutable parameters (e.g., pooling, concatenation), and thus $\nabla w_i$ is not always computed. 

In terms of concurrency, we use the Work-Depth (W-D) model to formulate the costs of computing the forward evaluation and backpropagation of different layer types. Table \ref{tbl:wdlayers} shows that the work ($\mathbf{W}$) performed in each layer asymptotically dominates the maximal operation dependency path ($\mathbf{D}$), which is at most logarithmic in the parameters. This result reaffirms the state of the practice, in which parallelism plays a major part in the feasibility of evaluating and training DNNs.

\begin{table}[t]
	\centering
	\caption{Asymptotic Work-Depth Characteristics of DNN Operators}
	\vspace{-1em}
	\label{tbl:wdlayers}
	\tiny
	\renewcommand{\arraystretch}{1.2}
	\begin{tabular}{ l l l l }
		\toprule
		\bf Operator Type & \bf Eval.
		& \bf Work ($\mathbf{W}$) & \bf Depth ($\mathbf{D}$)\\
		\midrule
		Activation & $y$ &
		$\mathcal{O}(NCHW)$&$\mathcal{O}(1)$\\&$\nabla w$&
		$\mathcal{O}(NCHW)$&$\mathcal{O}(1)$\\&$\nabla x$&
		$\mathcal{O}(NCHW)$&$\mathcal{O}(1)$ \\\addlinespace
		Fully Connected&$y$&$\mathcal{O}(C_{out}\cdot C_{in} \cdot N)$&$\mathcal{O}(\log C_{in})$ \\&$\nabla w$&
		$\mathcal{O}(C_{in}\cdot N \cdot C_{out})$&$\mathcal{O}(\log N)$ \\&$\nabla x$&
		$\mathcal{O}(C_{in} \cdot C_{out} \cdot N)$&$\mathcal{O}(\log C_{out})$
		\\\addlinespace
		Convolution (Direct)&$y$&$\mathcal{O}(N\cdot C_{out}\cdot C_{in}\cdot H'\cdot W'\cdot K_x\cdot K_y)$&$\mathcal{O}(\log K_x + \log K_y + \log C_{in})$\\&$\nabla w$&
		$\mathcal{O}(N\cdot C_{out}\cdot C_{in}\cdot H'\cdot W'\cdot K_x\cdot K_y)$&$\mathcal{O}(\log K_x + \log K_y + \log C_{in})$\\&$\nabla x$&
		$\mathcal{O}(N\cdot C_{out}\cdot C_{in}\cdot H\cdot W\cdot K_x\cdot K_y)$&$\mathcal{O}(\log K_x + \log K_y + \log C_{in})$\\\addlinespace
		Pooling&$y$&$\mathcal{O}(NCHW)$&$\mathcal{O}(\log K_x+\log K_y)$\\&$\nabla w$&
		---&---\\&$\nabla x$&
		$\mathcal{O}(NCHW)$&$\mathcal{O}(1)$\\\addlinespace
		Batch Normalization&$y$&$\mathcal{O}(NCHW)$&$\mathcal{O}(\log N)$\\&$\nabla w$&
		$\mathcal{O}(NCHW)$&$\mathcal{O}(\log N)$\\&$\nabla x$&
		$\mathcal{O}(NCHW)$&$\mathcal{O}(\log N)$\\
		\bottomrule										
	\end{tabular}
	\renewcommand{\arraystretch}{1}
\end{table}

As opposed to feed-forward networks, RNNs contain self-connections and thus cannot be trained with backpropagation alone. The most popular way to solve this issue is by applying \textit{backpropagation through time} (BPTT) \cite{bptt}, which unrolls the recurrent layer up to a certain amount of sequence length, using the same weights for each time-point. This creates a larger, feed-forward network that can be trained with the usual means.


\subsection{Trends in DNN Characteristics}
\label{sec:dnn:casestudies}

To understand how successful neural architectures orchestrate the aforementioned operators, we discuss five influential convolutional networks and highlight trends in their characteristics over the past years. Each of the networks, listed in Table \ref{tbl:popdnns}, has achieved state-of-the-art performance upon publication. The table summarizes these networks, their concurrency characteristics, and their achieved test accuracy on the ImageNet \cite{imagenet} (1,000 class challenge) and CIFAR-10 \cite{cifar} datasets. More detailed analysis of these networks can be found in Appendix \ref{app:dnntrends}.

\begin{table}[t]
	\caption{Popular Neural Network Characteristics}
	\vspace{-1em}
	\label{tbl:popdnns}
	\scriptsize
	\begin{tabular}{ l r r r r r }
		\toprule
		\bf Property & \bf LeNet \cite{lecun98} & \bf AlexNet \cite{alexnet} & \bf GoogLeNet \cite{inception}  & \bf ResNet \cite{resnet} & \bf DenseNet \cite{densenet} \\\midrule
		$|w|$ & 60K & 61M & 6.8M & 1.7M--60.2M & $\sim$15.3M--30M \\
		Layers ($\propto \mathbf{D}$) & 7 & 13 & 27 & 50--152 & 40--250 \\
		Operations ($\propto \mathbf{W}$, ImageNet-1k) & N/A & 725M & 1566M & $\sim$1000M--2300M & $\sim$600M--1130M \\
		Top-5 Error (ImageNet-1k) & N/A & 15.3\% & 9.15\%  & 5.71\% & 5.29\%\\
		Top-1 Error (CIFAR-10) & N/A & N/A    & N/A  & 6.41\% & 3.62\% \\\bottomrule
	\end{tabular}
\end{table}

The listed networks, as well as other works \cite{coates13, dean12,nin,vgg, deepcompression,squeezenet,nasnet,dpn}, indicate three periods in the history of classification neural networks:  experimentation ($\sim$1985--2010), growth (2010--2015), and resource conservation (2015--today).

In the experimentation period, different types of neural network structures (e.g., Deep Belief Networks \cite{bengio07}) were researched, and the methods to optimize them (e.g., backpropagation) were developed. Once the neural network community has converged on deep feed-forward networks (with the success of AlexNet cementing this decision), research during the growth period yielded networks with larger sizes and more operations, in an attempt to both increase model parallelism and solve increasingly complex problems. This trend was supported by the advent of GPUs and other large computational resources (e.g., the Google Brain cluster), increasing the available processing elements towards the average parallelism ($\mathbf{W}/\mathbf{D}$). 

However, as over-parameterization leads to overfitting, and since the resulting networks were too large to fit into consumer devices, efforts to decrease resource usage started around 2015, and so did the average parallelism (see table). Research has since focused on increasing expressiveness, mostly by producing deeper networks, while also reducing the number of parameters and operations required to forward-evaluate the given network. Parallelization efforts have thus shifted towards concurrency within minibatches (data parallelism, see Section \ref{sec:par}). By reducing memory and increasing energy efficiency, the resource conservation trend aims to move neural processing to the end user, i.e., to embedded and mobile devices. At the same time, smaller networks are faster to prototype and require less information to communicate when training on distributed platforms.

\section{Concurrency in Operators}
\label{sec:layercomp}

Given that neural network layers operate on 4-dimensional tensors (Fig. \ref{tbl:dims}) and the high locality of the operations, there are several opportunities for parallelizing layer execution. In most cases, computations (e.g., in the case of pooling operators) can be directly parallelized. However, in order to expose parallelism in other operator types, computations have to be reshaped. Below, we list efforts to model DNN performance, followed by a concurrency analysis of three popular operators.

\subsection{Performance Modeling}

\begin{figure}[t]
	\vspace{-1em}
	\centering
	\includegraphics[height=0.9in]{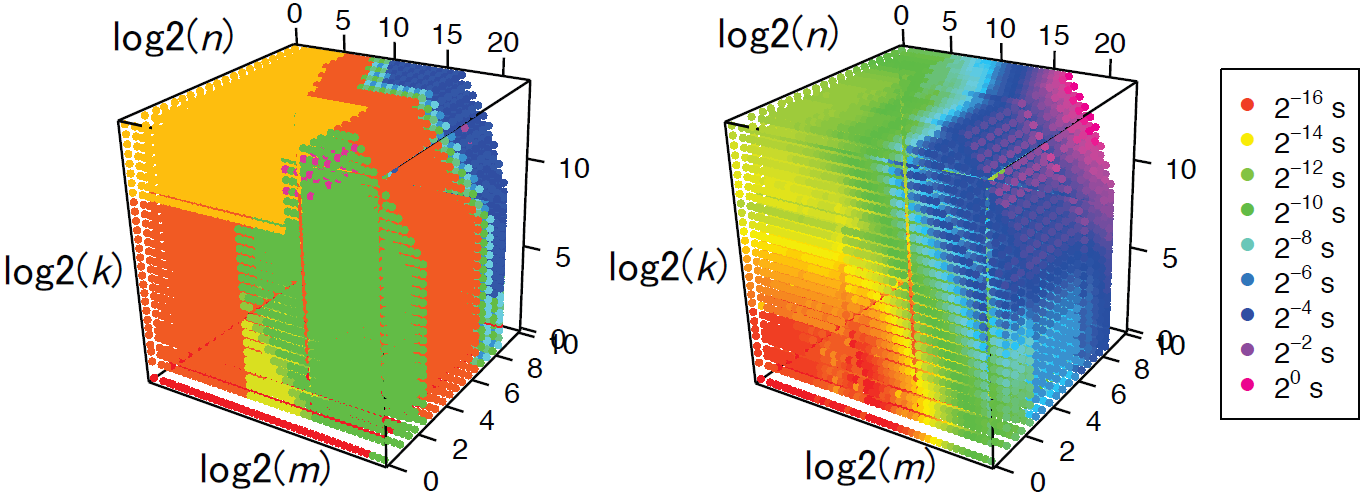}
	\caption{Performance of \texttt{cublasSgemm} on a Tesla K80 GPU for various matrix sizes (adapted from \cite{oyama16}).}
	\label{fig:perfmodel}
\end{figure}

Even with work and depth models, it is hard to estimate the runtime of a single DNN operator, let alone an entire network. Fig. \ref{fig:perfmodel} presents measurements of the performance of the highly-tuned matrix multiplication implementation in the NVIDIA CUBLAS library \cite{cublas}, which is at the core of nearly all operators. The figure shows that as the dimensions are modified, the performance does not change linearly, and that in practice the system internally chooses from one of 15 implementations for the operation, where the left-hand side of the figure depicts the segmentation.

In spite of the above observation, other works still manage to approximate the runtime of a given DNN with performance modeling. Using the values in the figure as a lookup table, it was possible to predict the time to compute and backpropagate through minibatches of various sizes with $\sim$5--19\% error, even on clusters of GPUs with asynchronous communication \cite{oyama16}. The same was achieved for CPUs in a distributed environment \cite{yan15}, using a similar approach, and for Intel Xeon Phi accelerators \cite{viebke2017} strictly for training time estimation (i.e., not individual layers or DNN evaluation). Paleo \cite{paleo} derives a performance model from operation counts alone (with 10--30\% prediction error), and Pervasive CNNs \cite{song17} uses performance modeling to select networks with decreased accuracy to match real-time requirements from users. To further understand the performance characteristics of DNNs, Demmel and Dinh \cite{demmel18} provide lower bounds on communication requirements for the convolution and pooling operators.

\subsection{Fully Connected Layers}

As described in Section \ref{sec:dnn:layers}, a fully connected layer can be expressed and modeled (see Table \ref{tbl:wdlayers}) as a matrix-matrix multiplication of the weights and the neuron values (column per minibatch sample). To that end, efficient linear algebra libraries, such as CUBLAS \cite{cublas} and MKL \cite{mkl}, can be used. The BLAS \cite{blas} GEneral Matrix-Matrix multiplication (GEMM) operator, used for this purpose, also includes scalar factors that enable matrix scaling and accumulation, which can be used when batching groups of neurons.

Vanhoucke et al.~\cite{optcpu} present a variety of methods to further optimize CPU implementations of fully connected layers. In particular, the paper shows efficient loop construction, vectorization, blocking, unrolling, and batching. The paper also demonstrates how weights can be quantized to use fixed-point math instead of floating point.

\subsection{Convolution}

\begin{figure}[t]
	\centering
	\begin{subfigure}[t]{0.3\textwidth}
		\centering
		\includegraphics[height=1.2in]{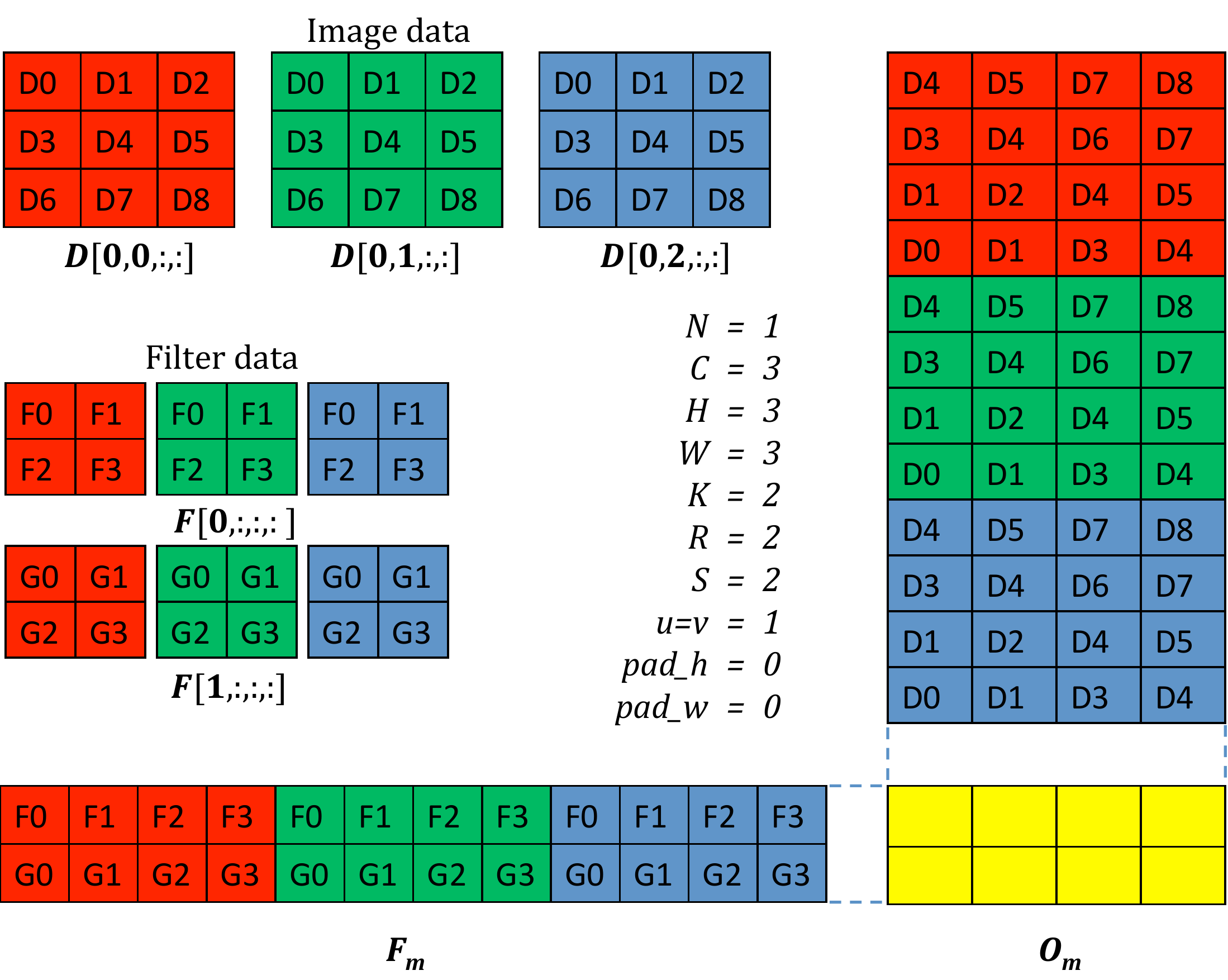}
		\caption{\texttt{im2col} (adapted from \cite{cudnn})}
		\label{fig:conv:im2col}
	\end{subfigure}
	\quad
	\begin{subfigure}[t]{0.3\textwidth}
		\centering
		\includegraphics[height=1.2in,trim={5cm 0.5cm 5.75cm 0cm},clip]{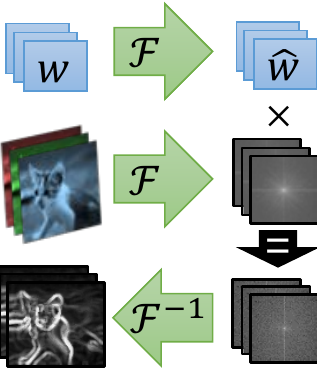}
		\caption{FFT}
		\label{fig:conv:fft}
	\end{subfigure}
	\quad
	\begin{subfigure}[t]{0.3\textwidth}
		\centering
		\includegraphics[height=1.2in,trim={0cm 5cm 24cm 0cm},clip]{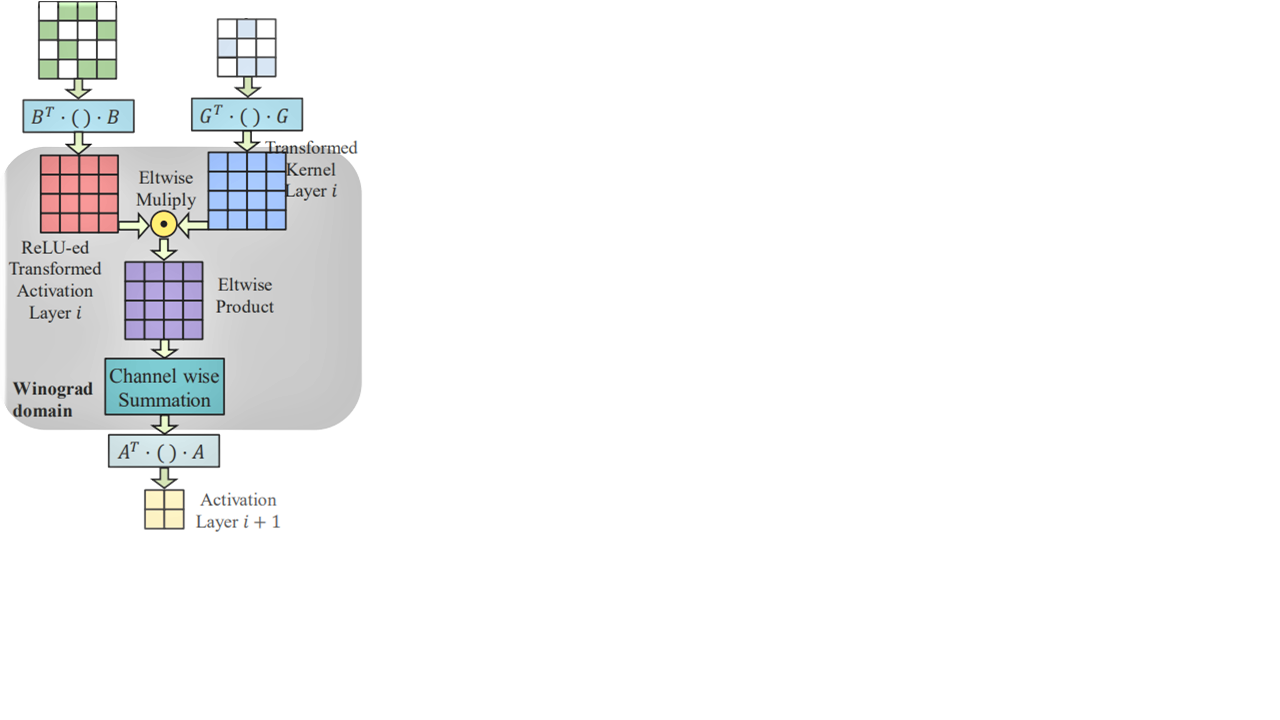}
		\caption{Winograd (adapted from \cite{liu18winograd})}
		\label{fig:conv:winograd}
	\end{subfigure}
	\vspace{-1em}
	\caption{Computation Methods for Convolutional Operators}
	\label{fig:conv}
	\vspace{-1.5em}
\end{figure}

Convolutions constitute the majority of computations involved in training and inference of DNNs. As such, the research community and the industry have invested considerable efforts into optimizing their computation on all platforms. Fig. \ref{fig:conv} depicts the convolution methods detailed below, and Table \ref{tbl:wdconv} summarizes their work and depth characteristics (see Appendix \ref{app:conv} for detailed analyses).

While a convolution operator (Eq. \ref{eq:conv}) can be computed directly, it will not fully utilize the resources of vector processors (e.g., Intel's AVX registers) and many-core architectures (e.g., GPUs), which are geared towards many parallel multiplication-accumulation operations. It is possible, however, to increase the utilization by ordering operations to maximize data reuse \cite{demmel18}, introducing data redundancy, or via basis transformation. 

The first algorithmic change proposed for convolutional operators was the use of the well-known technique to transform a discrete convolution into matrix multiplication, using \textit{Toeplitz matrices} (colloquially known as \textit{im2col}). The first occurrence of unrolling convolutions in CNNs \cite{im2col1} used both CPUs and GPUs for training (since the work precedes CUDA, it uses Pixel Shaders for GPU computations). The method was subsequently popularized by Coates et al.~\cite{coates13}, and consists of reshaping the images in the minibatch from 3D tensors to 2D matrices. Each 1D row in the matrix contains an unrolled 2D patch that would usually be convolved (possibly with overlap), generating redundant information (see Fig. \ref{fig:conv:im2col}). The convolution kernels are then stored as a 2D matrix, where each column represents an unrolled kernel (one convolution filter). Multiplying those two matrices results in a matrix that contains the convolved tensor in 2D format, which can be reshaped to 3D for subsequent operations. Note that this operation can be generalized to 4D tensors (an entire minibatch), converting it into a single matrix multiplication. Alternatively, the kernels can be unrolled to rows (\textit{kn2row}) for the matrix multiplication \cite{kn2row}.

While processor-friendly, the GEMM method (as described above) consumes a considerable amount of memory, and thus was not scalable. Practical implementations of the GEMM method, such as in CUDNN \cite{cudnn}, implement ``implicit GEMM'', in which the Toeplitz matrix is never materialized. It was also reported \cite{cong2014strassen} that the Strassen matrix multiplication \cite{strassen} can be used for the underlying computation, reducing the number of operations by up to 47\%.

A second method to compute convolutions is to make use of the Fourier domain, in which convolution is defined as an element-wise multiplication \cite{fft,fft2}. In this method, both the data and the kernels are transformed using FFT, multiplied, and the inverse FFT is applied on the result: 
\small
\[
y_{i,j,*,*}=\mathcal{F}^{-1}\left(\sum_{m=0}^{C_{in}}\mathcal{F}\left(x_{i,m,*,*}\right)\circ\mathcal{F}\left(w_{j,m,*,*}\right)\right)
\]
\normalsize
where $\mathcal{F}$ denotes the Fourier Transform and $\circ$ is element-wise multiplication. Note that for a single minibatch, it is enough to transform $w$ once and reuse the results.

Experimental results \cite{fft2} have shown that the larger the convolution kernels are, the more beneficial FFT becomes, yielding up to 16$\times$ performance over the GEMM method, which has to process patches of proportional size to the kernels. Additional optimizations were made to the FFT and IFFT operations \cite{fft2}, using DNN-specific knowledge: (a) The process uses decimation-in-frequency for FFT and decimation-in-time for IFFT in order to mitigate bit-reversal instructions; (b) multiple FFTs with sizes $\le$32 are batched together and performed at the warp-level on the GPU; and (c) pre-computation of twiddle factors.

Working with DNNs, FFT-based convolution can be optimized further. In ZNNi \cite{znni}, the authors observed that due to zero-padding, the convolutional kernels, which are considerably smaller than the images, mostly consist of zeros. Thus, pruned FFT \cite{fftpruning} can be executed for transforming the kernels, reducing the number of operations by 3$\times$. In turn, the paper reports 5$\times$ and 10$\times$ speedups for CPUs and GPUs, respectively.

The prevalent method used today to perform convolutions is Winograd's algorithm for minimal filtering \cite{winograd}. First proposed by Lavin and Gray \cite{lavin16}, the method modifies the original algorithm for multiple filters (as is the case in convolutions), performing the following computation for one tile:
\small
\[
y_{i,j,*,*}=A^T\left(\sum_{m=0}^{C_{in}}Gw_{j,m,*,*}G^T\circ B^T x_{i,m,*,*} B\right)A,
\]
\normalsize
with the matrices $A,G,B$ constructed as in Winograd's algorithm (implementation in Appendix \ref{app:conv}).

Since the number of operations in Winograd convolutions grows quadratically with filter size, the convolution is decomposed into a sum of tiled, small convolutions, and the method is strictly used for small kernels (e.g., 3$\times$3). Additionally, because the magnitude of elements in the expression increases with filter size, the numerical accuracy of Winograd convolution is generally lower than the other methods, and decreases as larger filters are used.

Table \ref{tbl:wdconv} lists the concurrency characteristics of the aforementioned convolution implementations, using the Work-Depth model. From the table, we can see that each method exhibits different behavior, where the average parallelism ($\mathbf{W}/\mathbf{D}$) can be determined by the kernel size or by image size (e.g., FFT). This coincides with experimental results \cite{fft2,lavin16,cudnn}, which show that there is no ``one-size-fits-all'' convolution method. We can also see that the Work and Depth metrics are not always sufficient to reason about absolute performance, as the Direct and im2col methods exhibit the same concurrency characteristics, even though \texttt{im2col} is faster in many cases, due to high processor utilization and memory reuse (e.g., caching) opportunities.

Data layout also plays a role in convolution performance. Li et al.~\cite{layout16} assert that convolution and pooling operators can be computed faster by transposing the data from $N$$\times$$C$$\times$$H$$\times$$W$ tensors to $C$$\times$$H$$\times$$W$$\times$$N$. The paper reports up to 27.9$\times$ performance increase over the state-of-the-art for a single operator, and 5.6$\times$ for a full DNN (AlexNet). The paper reports speedup even in the case of transposing the data during the computation of the DNN, upon inputting the tensor to the operator.

DNN primitive libraries, such as CUDNN \cite{cudnn} and MKL-DNN \cite{mkldnn}, provide a variety of convolution methods and data layouts. In order to assist users in a choice of algorithm, such libraries provide functions that choose the best-performing algorithm given tensor sizes and memory constraints. Internally, the libraries may run all methods and pick the fastest one.

\begin{table}[t]
	\caption{Work-Depth Analysis of Convolution Implementations}
	\vspace{-1em}
	\label{tbl:wdconv}
	\scriptsize
	\begin{tabular}{l l l}
		\toprule
		\bf Method & \bf Work ($\mathbf{W}$) & \bf Depth ($\mathbf{D}$)\\
		\midrule
		Direct & $N\cdot C_{out}\cdot H'\cdot W'\cdot C_{in}\cdot K_y\cdot K_x$ & $\left\lceil\log_2 C_{in}\right\rceil+\left\lceil\log_2 K_y\right\rceil+\left\lceil\log_2 K_x\right\rceil$ \\\addlinespace
		im2col & $N\cdot C_{out}\cdot H'\cdot W'\cdot C_{in}\cdot K_y\cdot K_x$ & $\left\lceil\log_2 C_{in}\right\rceil+\left\lceil\log_2 K_y\right\rceil+\left\lceil\log_2 K_x\right\rceil$ \\\addlinespace
		FFT & $c\cdot HW\log_2 (HW)\cdot (C_{out}\cdot C_{in} +$ & $2\left\lceil\log_2 HW\right\rceil+\left\lceil\log_2 C_{in}\right\rceil$\\ 
		& $N\cdot C_{in} + N\cdot C_{out}) + H W N \cdot C_{in} \cdot C_{out}$ &\\\addlinespace
		Winograd & \multirow{2}{*}{$\alpha (r^2 + \alpha r + 2\alpha^2+ \alpha m + m^2) + C_{out}\cdot C_{in}\cdot P$} & \multirow{2}{*}{$2\left\lceil\log_2 r\right\rceil+4\left\lceil\log_2 \alpha\right\rceil+\left\lceil\log_2 C_{in}\right\rceil$}\\
		($m\times m$ tiles,&&\\
		$r\times r$ kernels)&$\left(\alpha\equiv m-r+1,\quad P\equiv N\cdot \left\lceil H/m \right\rceil\cdot \left\lceil W/m \right\rceil\right)$&\\
		\bottomrule
	\end{tabular}
	\vspace{-1em}
\end{table}

\subsection{Recurrent Units}

The complex gate systems that occur within RNN units (e.g., LSTMs, see Fig. \ref{fig:rnn:lstm}) contain multiple operations, each of which incurs a small matrix multiplication or an element-wise operation. Due to this reason, these layers were traditionally implemented as a series of high-level operations, such as GEMMs. However, further acceleration of such layers is possible. Moreover, since RNN units are usually chained together (forming consecutive recurrent layers), two types of concurrency can be considered: within the same layer, and between consecutive layers.

Appleyard et al. \cite{cudnnrnn} describe several optimizations that can be implemented for GPUs. The first optimization fuses all computations (GEMMs and otherwise) into one function (kernel), saving intermediate results in scratch-pad memory. This both reduces the kernel scheduling overhead, and conserves round-trips to the global memory, using the multi-level memory hierarchy of the massively parallel GPU. Other optimizations include pre-transposition of matrices and enabling concurrent execution of independent recurrent units on different multi-processors on the GPU. 

Inter-layer concurrency is achieved through pipeline parallelism, with which Appleyard et al. implement stacked RNN unit computations, immediately starting to propagate through the next layer once its data dependencies have been met. Overall, these optimizations result in $\sim$11$\times$ performance increase over the high-level implementation.

\begin{figure}[t]
	\centering
	\begin{subfigure}[b]{0.56\linewidth}
		\centering
		\includegraphics[height=1in]{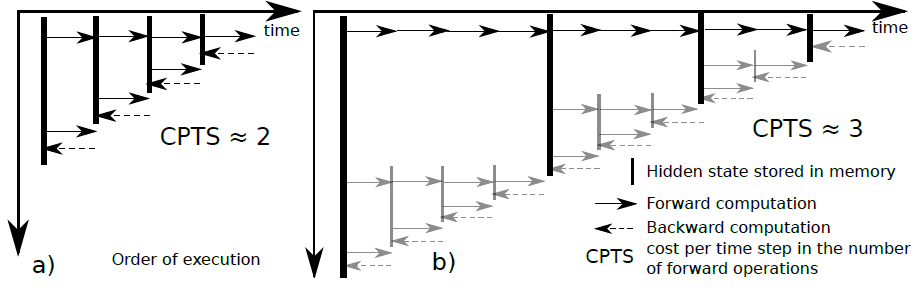}
		\caption{Dynamic Programming BPTT \cite{gruslys16}}		
		\label{fig:rnnopt:dp}
	\end{subfigure}
	\qquad
	\begin{subfigure}[b]{0.35\linewidth}
		\centering
		\includegraphics[height=1in]{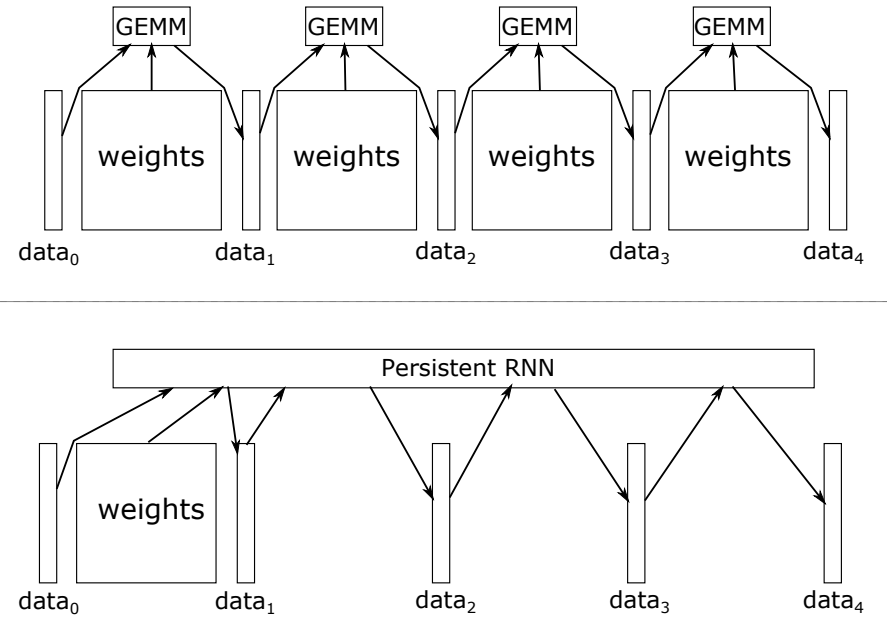}
		\caption{Persistent RNNs \cite{diamos16}}
		\label{fig:rnnopt:prnn}
	\end{subfigure}
	\vspace{-1em}
	\caption{RNN Optimizations}
	\label{fig:rnnopt}
	\vspace{-1em}
\end{figure}

From the memory consumption perspective, dynamic programming was proposed \cite{gruslys16} for RNNs (see Fig. \ref{fig:rnnopt:dp}) in order to balance between caching intermediate results and recomputing forward inference for backpropagation. For long sequences (1000 time-points), the algorithm conserves 95\% memory over standard BPTT, while adding $\sim$33\% time per iteration. A similar result has been achieved when re-computing convolutional operators as well \cite{sublinearmem}, yielding memory costs sublinear in the number of layers.
 
Persistent RNNs \cite{diamos16} are an optimization that addresses two limitations of GPU utilization: small minibatch sizes and long sequences of inputs. By caching the weights of standard RNN units on the GPU registers, they optimize memory round-trips between timesteps ($x^{(t)}$) during training (Fig. \ref{fig:rnnopt:prnn}). In order for the registers not to be scheduled out, this requires the GPU kernels that execute the RNN layers to be ``persistent'', performing global synchronization on their own and circumventing the normal GPU programming model. The approach attains up to $\sim$30$\times$ speedup over previous state-of-the-art for low minibatch sizes, performing on the order of multiple TFLOP/s per-GPU, even though it does not execute GEMM operations and loads more memory for each multi-processor. Additionally, the approach reduces the total memory footprint of RNNs, allowing users to stack more layers using the same resources.

\section{Concurrency in Networks}
\label{sec:par}

\begin{figure}[t]
	\centering
	\vspace{-1em}
	\begin{subfigure}{0.26\textwidth}
		\centering
		\includegraphics[page=1,height=1in,clip,trim={0cm 0.5cm 9.5cm 0cm}]{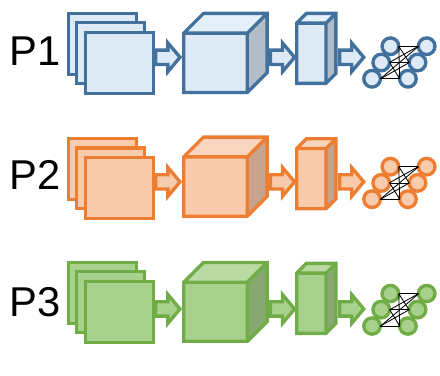}
		\caption{Data Parallelism}
		\label{fig:parallelism:data}
	\end{subfigure}
	\quad
	\begin{subfigure}{0.32\textwidth}
		\centering
		\includegraphics[page=2,height=1in,clip,trim={0cm 0.8cm 9.5cm 0.55cm}]{figures/parallelism}
		\caption{Model Parallelism}
		\label{fig:parallelism:model}		
	\end{subfigure}
	\quad
	\begin{subfigure}{0.32\textwidth}
		\centering
		\includegraphics[page=3,height=1in,clip,trim={0cm 0.8cm 9.5cm 0.55cm}]{figures/parallelism}
		\caption{Layer Pipelining}
		\label{fig:parallelism:pipelining}		
	\end{subfigure}
	\vspace{-1em}
	\caption{Neural Network Parallelism Schemes}
	\label{fig:parallelism}

\end{figure}

The high average parallelism ($\mathbf{W}/\mathbf{D}$) in neural networks may not only be harnessed to compute individual operators efficiently, but also to evaluate the whole network concurrently with respect to different dimensions. Owing to the use of minibatches, the breadth ($\propto \mathbf{W}$) of the layers, and the depth of the DNN ($\propto \mathbf{D}$), it is possible to partition both the forward evaluation and the backpropagation phases (lines \ref{alg:sgd:fwd}--\ref{alg:sgd:bwd} in Algorithm \ref{alg:sgd}) among parallel processors. Below, we discuss three prominent partitioning strategies, illustrated in Fig. \ref{fig:parallelism}: partitioning by input samples (\textbf{data parallelism}), by network structure (\textbf{model parallelism}), and by layer (\textbf{pipelining}).

\subsection{Data Parallelism}
\label{sec:par:data}

In minibatch SGD (Section \ref{sec:term:sgd}), data is processed in increments of $N$ samples. As most of the operators are independent with respect to $N$ (Section \ref{sec:dnn}), a straightforward approach for parallelization is to partition the work of the minibatch samples among multiple computational resources (cores or devices). This method (initially named pattern parallelism, as input samples were called patterns), dates back to the first practical implementations of artificial neural networks \cite{nips89datapar}.

It could be argued that the use of minibatches in SGD for neural networks was initially driven by data parallelism. Farber and Asanovi\'{c} \cite{spert} used multiple vector accelerator microprocessors (Spert-II) to parallelize error backpropagation for neural network training. To support data parallelism, the paper presents a version of delayed gradient updates called ``bunch mode'', where the gradient is updated several times prior to updating the weights, essentially equivalent to minibatch SGD.

One of the earliest occurrences of mapping DNN computations to  data parallel architectures (e.g., GPUs) were performed by Raina et al.~\cite{gpudp}. The paper focuses on the problem of training Deep Belief Networks \cite{dbn}, mapping the unsupervised training procedure to GPUs by running minibatch SGD. The paper shows speedup of up to 72.6$\times$ over CPU when training Restricted Boltzmann Machines. Today, data parallelism is supported by the vast majority of deep learning frameworks, using a single GPU, multiple GPUs, or a cluster of multi-GPU nodes.

The scaling of data parallelism is naturally defined by the minibatch size (Table \ref{tbl:wdlayers}). Apart from Batch Normalization (BN) \cite{bn}, all operators mentioned in Section \ref{sec:dnn} operate on a single sample at a time, so forward evaluation and backpropagation are almost completely independent. In the weight update phase, however, the results of the partitions have to be averaged to obtain the gradient w.r.t. the whole minibatch, which potentially induces an allreduce operation. Furthermore, in this partitioning method, all DNN parameters have to be accessible for all participating devices, which means that they should be replicated. 

\subsubsection{Neural Architecture Support for Large Minibatches}

By applying various modifications to the training process, recent works have successfully managed to increase minibatch size to 8k samples \cite{goyal17}, 32k samples \cite{b32k}, and even 64k \cite{batchsize} without losing considerable accuracy. While the generalization issue still exists (Section \ref{sec:tradeoff}), it is not as severe as claimed in prior works \cite{seide14}. One bottleneck that hinders scaling of data parallelism, however, is the BN operator, which requires a full synchronization point upon invocation. Since BN recurs multiple times in some DNN architectures \cite{resnet}, this is too costly. Thus, popular implementations of BN follow the approach driven by large-batch papers \cite{goyal17,gbn,b32k}, in which small subsets (e.g., 32 samples) of the minibatch are normalized independently. If at least 32 samples are scheduled to each processor, then this synchronization point is local, which in turn increases scaling.

Another approach to the BN problem is to define a different operator altogether. Weight Normalization (WN) \cite{wn} proposes to separate the parameter ($w$) norm from its directionality by way of re-parameterization. In WN, the weights are defined as $w = \left(\frac{g}{\|v\|}\right) \cdot v$, where $g$ represents weight magnitude and $v$ a normalized direction (as changing the magnitude of $v$ will not introduce changes in $\nabla\ell$). WN decreases the depth ($\mathbf{D}$) of the operator from $\mathcal{O}(\log N)$ to $\mathcal{O}(1)$, removing inter-dependencies within the minibatch. According to the authors, WN reduces the need for BN, achieving comparable accuracy using a simplified version of BN (without variance correction).

\subsubsection{Coarse- and Fine-Grained Data Parallelism}

Additional approaches for data parallelism were proposed in literature. 
In ParallelSGD \cite{parsgd}, SGD is run (possibly with minibatches) $k$ times in parallel, dividing the dataset among the processors. After the convergence of all SGD instances, the resulting weights are aggregated and averaged to obtain $\overline{w}$, exhibiting coarse-grained parallelism. 

ParallelSGD \cite{parsgd}, as well as other deep learning implementations \cite{le11,dbnmapreduce,dynsgd17}, were designed with the MapReduce \cite{mapreduce} programming paradigm. Using MapReduce, it is easy to schedule parallel tasks onto multiple processors, as well as distributed environments. Prior to these works, the potential scaling of MapReduce was studied \cite{mrml} on a variety of machine learning problems, including NNs, promoting the need to shift from single-processor learning to distributed memory systems. 

\begin{figure}[t]
	\centering
	\begin{subfigure}[b]{0.49\linewidth}
		\includegraphics[height=0.95in]{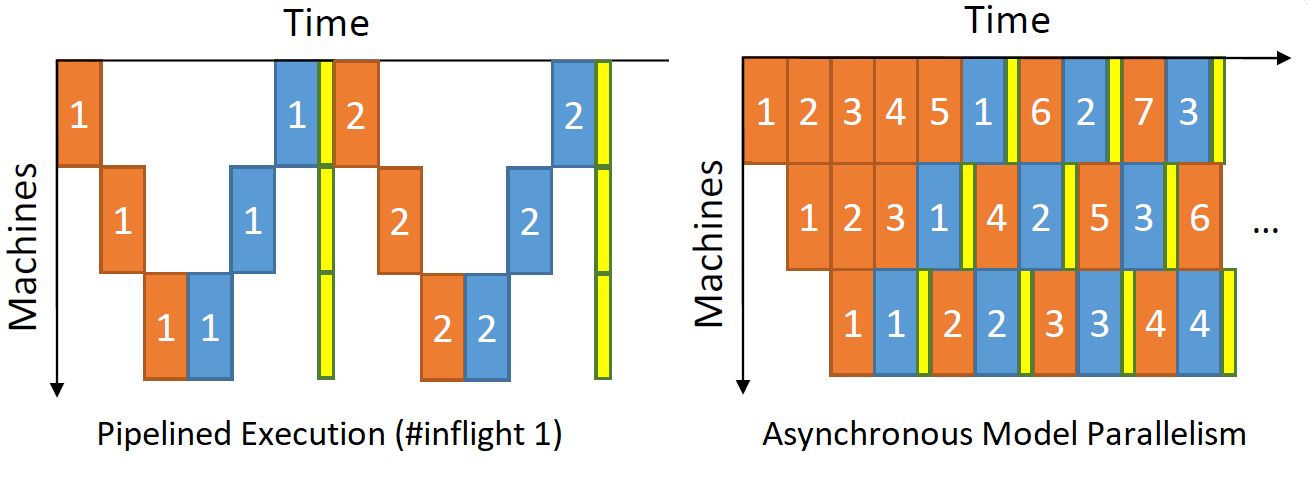}
		\caption{Pipelined Asynchronous Execution \cite{ampnet}}
		\label{fig:ampnet}
	\end{subfigure}
	\qquad
	\begin{subfigure}[b]{0.4\linewidth}
		\includegraphics[height=1in]{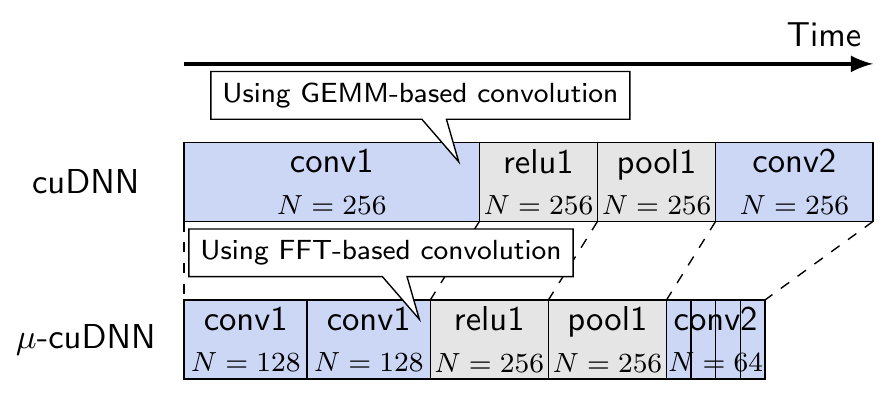}
		\caption{Convolution Decomposition  \cite{microbatch2}}
		\label{fig:microbatch}
	\end{subfigure}
	\vspace{-1em}
	\caption{Data Parallelism Schemes}
	\vspace{-1.35em}
\end{figure}

While the MapReduce model was successful for deep learning at first, its generality hindered DNN-specific optimizations. Therefore, current implementations make use of high-performance communication interfaces (e.g., MPI) to implement fine-grained parallelism features, such as reducing latencies via asynchronous execution and pipelining \cite{ampnet} (Fig. \ref{fig:ampnet}), sparse communication (see Section \ref{sec:dist:compression}), and exploiting parallelism within a given computational resource \cite{znni,microbatch2}. In the last category, minibatches are fragmented into micro-batches (Fig. \ref{fig:microbatch}) that are decomposed \cite{znni} or computed sequentially \cite{microbatch2}. This reduces the required memory footprint, thus making it possible to choose faster methods that require more memory, as well as enabling hybrid CPU-GPU inference.

\subsection{Model Parallelism}
\label{sec:par:model}

The second partitioning strategy for DNN training is model parallelism (also known as network parallelism). This strategy divides the work according to the neurons in each layer, namely the $C$, $H$, or $W$ dimensions in a 4-dimensional tensor. In this case, the sample minibatch is copied to all processors, and different parts of the DNN are computed on different processors, which can conserve memory (since the full network is not stored in one place) but incurs additional communication after every layer. 

Since the minibatch size does not change in model parallelism, the utilization vs. generalization tradeoff (Section \ref{sec:tradeoff}) does not apply. Nevertheless, the DNN architecture creates layer interdependencies, which, in turn, generate communication that determines the overall performance. Fully connected layers, for instance, incur all-to-all communication (as opposed to allreduce in data parallelism), as neurons connect to all the neurons of the next layer.

To reduce communication costs in fully connected layers, it has been proposed \cite{music} to introduce redundant computations to neural networks. In particular, the proposed method partitions an NN such that each processor will be responsible for twice the neurons (with overlap), and thus would need to compute more but communicate less.

Another method proposed for reducing communication in fully connected layers is to use Cannon's matrix multiplication algorithm, modified for DNNs \cite{cannon}. The paper reports that Cannon's algorithm produces better efficiency and speedups over simple partitioning on small-scale multi-layer fully connected networks.

\begin{figure}[t]
	\centering
	\begin{subfigure}[t]{0.35\linewidth}
		\centering
		\includegraphics[height=1in]{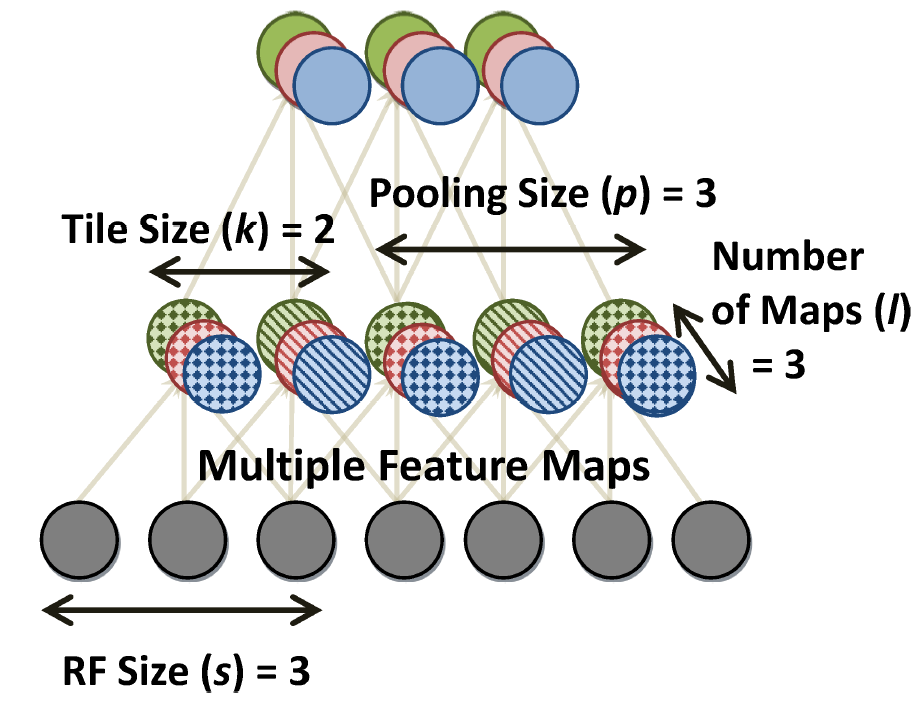}
		\caption{Locally Connected Networks \cite{tiledcnn}}
		\label{fig:lcn}
	\end{subfigure}
	\qquad
	\begin{subfigure}[t]{0.45\linewidth}
		\centering
		\includegraphics[height=1in]{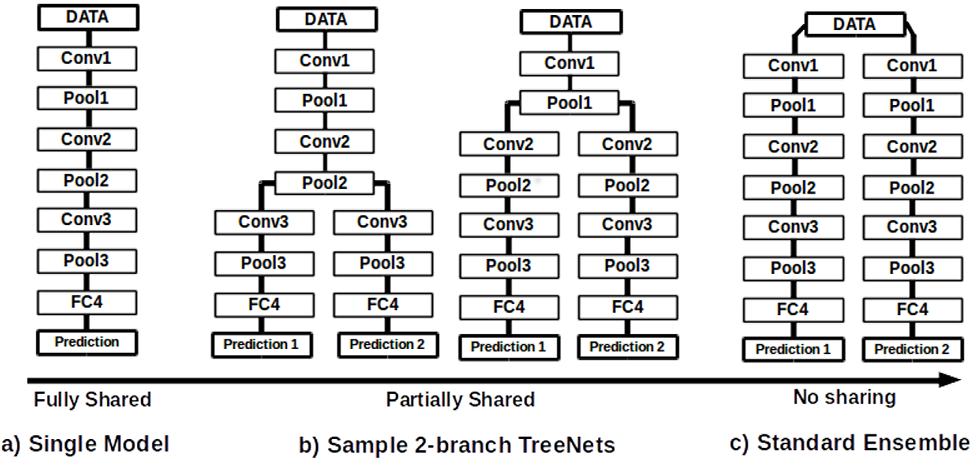}
		\caption{TreeNets \cite{treenets}}
		\label{fig:treenets}
	\end{subfigure}
	\vspace{-1em}
	\caption{Model Parallelism Schemes}
	\vspace{-1.25em}
\end{figure}

As for CNNs, using model parallelism for convolutional operators is relatively inefficient. If samples are partitioned across processors by feature (channel), then each convolution would have to obtain all results from the other processors to compute its result, as the operation sums over all features. To mitigate this problem, Locally Connected Networks (LCNs) \cite{tiledcnn} were introduced. While still performing convolutions, LCNs define multiple local filters for each region (Fig. \ref{fig:lcn}), enabling partitioning by the $C,H,W$ dimensions that does not incur all-to-all communication. 

Using LCNs and model parallelism, the work presented by Coates et al. \cite{coates13} managed to outperform a CNN of the same size running on 5,000 CPU nodes with a 3-node multi-GPU cluster. Due to the lack of weight sharing (apart from spatial image boundaries), training is not communication-bound, and scaling can be achieved. Successfully applying the same techniques on CNNs requires fine-grained control over parallelism, as we shall show in Section \ref{sec:par:hybrid}. Unfortunately, weight sharing is an important part of CNNs, contributing to memory footprint reduction as well as improving generalization, and thus standard convolutional operators are used more frequently than LCNs.

A second form of model parallelism is the replication of DNN elements. In TreeNets \cite{treenets}, the authors study ensembles of DNNs (groups of separately trained networks whose results are averaged, rather than their parameters), and propose a mid-point between ensembles and training a single model: a certain layer creates a ``junction'', from which multiple copies of the network are trained (see Fig. \ref{fig:treenets}). The paper defines ensemble-aware loss functions and backpropagation techniques, so as to regularize the training process. The training process, in turn, is parallelized across the network copies, assigning each copy to a different processor. The results presented in the paper for three datasets indicate that TreeNets essentially train an ensemble of expert DNNs.

\subsection{Pipelining}
\label{sec:par:pipeline}

In deep learning, pipelining can either refer to overlapping computations, i.e., between one layer and the next (as data becomes ready); or to partitioning the DNN according to depth, assigning layers to specific processors (Fig. \ref{fig:parallelism:pipelining}). Pipelining can be viewed as a form of data parallelism, since elements (samples) are processed through the network in parallel, but also as model parallelism, since the length of the pipeline is determined by the DNN structure.
 
The first form of pipelining can be used to overlap forward evaluation, backpropagation, and weight updates. This scheme is widely used in practice \cite{tensorflow2015,torch,jia2014caffe}, and increases utilization by mitigating processor idle time.
In a finer granularity, neural network architectures can be designed around the principle of overlapping layer computations, as is the case with Deep Stacking Networks (DSN) \cite{deepstack}. In DSNs, each step computes a different fully connected layer of the data. However, the results of all previous steps are concatenated to the layer inputs (see Fig. \ref{fig:deepstack}). This enables each layer to be partially computed in parallel, due to the relaxed data dependencies. 

As for layer partitioning, there are several advantages for a multi-processor pipeline over both data and model parallelism: (a) there is no need to store all parameters on all processors during forward evaluation and backpropagation (as with model parallelism); (b) there is a fixed number of communication points between processors (at layer boundaries), and the source and destination processors are always known. Moreover, since the processors always compute the same layers, the weights can remain cached to decrease memory round-trips. Two disadvantages of pipelining, however, are that data (samples) have to arrive at a specific rate in order to fully utilize the system, and that latency proportional to the number of processors is incurred.

In the following section, we discuss two implementations of layer partitioning --- DistBelief \cite{dean12} and Project Adam \cite{adam} --- which combine the advantages of pipelining with data and model parallelism.

\subsection{Hybrid Parallelism}
\label{sec:par:hybrid}

\begin{figure}[t]
	\centering
	\begin{subfigure}[t]{0.29\linewidth}
		\centering
		\includegraphics[height=1in,clip,trim={6cm 0cm 10cm 0cm}]{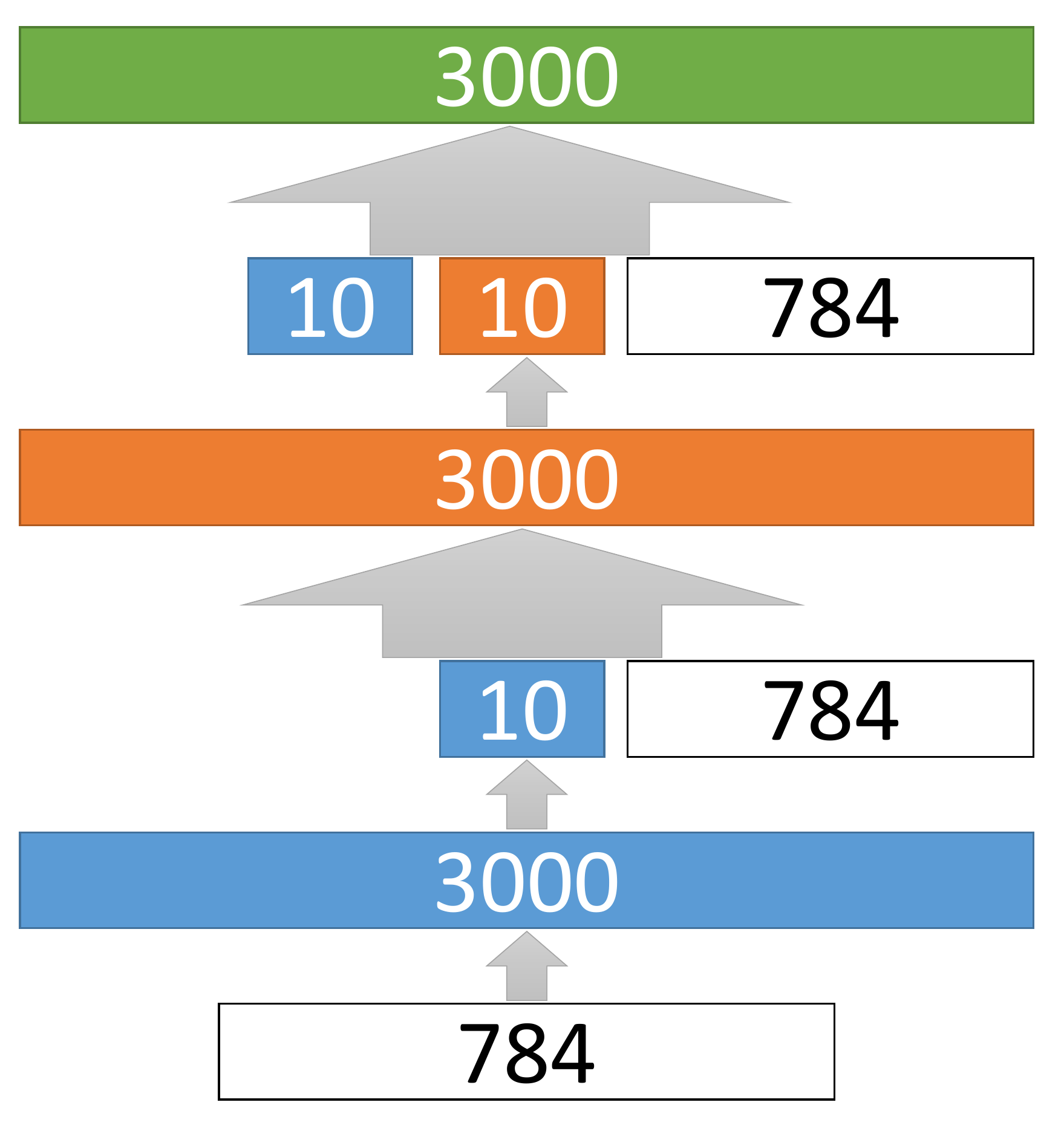}
		\caption{Deep Stacking Network~\cite{deepstack}}
		\label{fig:deepstack}
	\end{subfigure}
	\qquad
	\begin{subfigure}[t]{0.29\linewidth}
		\centering
		\includegraphics[page=4,height=1in,clip,trim={0cm 0.5cm 9.25cm 0cm}]{figures/parallelism}
		\caption{Hybrid Parallelism}
		\label{fig:hybriddm}
	\end{subfigure}
	\qquad
	\begin{subfigure}[t]{0.27\linewidth}
		\centering
		\includegraphics[height=1in]{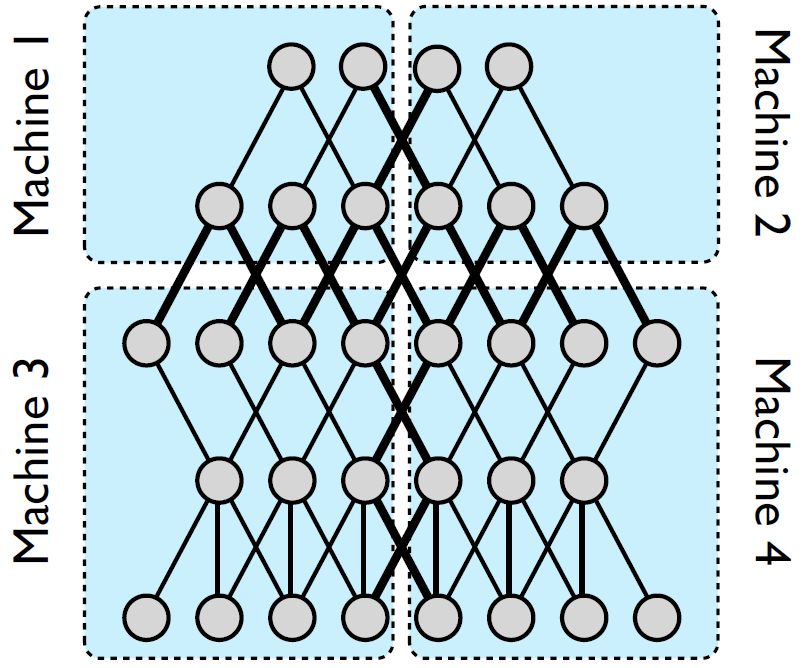}
		\caption{DistBelief Replica~\cite{dean12}}
		\label{fig:dbreplica}
	\end{subfigure}
	\vspace{-1em}
	\caption{Pipelining and Hybrid Parallelism Schemes}
	\vspace{-1.25em}
\end{figure}

The combination of multiple parallelism schemes can overcome the drawbacks of each scheme. Below we overview successful instances of such hybrids.

In AlexNet, most of the computations are performed in the convolutional layers, but most of the parameters belong to the fully connected layers. When mapping AlexNet to a multi-GPU node using data or model parallelism separately, the best reported speedup for 4 GPUs over one is $\sim$2.2$\times$ \cite{mgpu}. One successful example \cite{krizhevsky2014one} of a hybrid scheme applies data parallelism to the convolutional layer, and model parallelism to the fully connected part (see Fig. \ref{fig:hybriddm}). Using this hybrid approach, a speedup of up to 6.25$\times$ can be achieved for 8 GPUs over one, with less than 1\% accuracy loss (due to an increase in minibatch size). These results were also reaffirmed in other hybrid implementations \cite{bnsc15}, in which 3.1$\times$ speedup was achieved for 4 GPUs using the same approach, and derived theoretically using communication cost analysis \cite{gholami18}, promoting 1.5D matrix multiplication algorithms for integrated data/model parallelism.

AMPNet \cite{ampnet} is an asynchronous implementation of DNN training on CPUs, which uses an intermediate representation to implement fine-grained model parallelism. In particular, internal parallel tasks within and between layers are identified and scheduled asynchronously. Additionally, asynchronous execution of dynamic control flow enables pipelining the tasks of forward evaluation, backpropagation, and weight update (Fig. \ref{fig:ampnet}, right). The main advantage of AMPNet is in recurrent, tree-based, and gated-graph neural networks, all of which exhibit heterogeneous characteristics, i.e., variable length for each sample and dynamic control flow (as opposed to homogeneous CNNs). The paper shows speedups of up to 3.94$\times$ over the TensorFlow \cite{tensorflow2015} framework.

Lastly, the DistBelief \cite{dean12} distributed deep learning system combines all three parallelism strategies. In the implementation, training is performed on multiple model replicas simultaneously, where each replica is trained on different samples (data parallelism). Within each replica (shown in Fig. \ref{fig:dbreplica}), the DNN is distributed both according to neurons in the same layer (model parallelism), and according to the different layers (pipelining). Project Adam \cite{adam} extends upon the ideas of DistBelief and exhibits the same types of parallelism. However, in Project Adam pipelining is restricted to different CPU cores on the same node.

\section{Concurrency in Training}
\label{sec:dist}

\begin{figure}[t]
	\begin{minipage}[b]{.63\textwidth}
	\scriptsize
	\begin{tabular}{ l p{2.15in} }
		\toprule
		\bf Category & \bf Method\\\midrule\addlinespace
		\multicolumn{2}{ l }{\bf Model Consistency} \\
		Synchronization & Synchronous \cite{coates13, mgpu, seide14onebit, strom15, sparknet, firecaffe} \\
		& Stale-Synchronous \cite{ssp13, asysg, zhang16, rudra16,dynsgd17} \\
		& Asynchronous \cite{hogwild, dean12, dogwild, paine13, zhang13, keuper15,tamingthewild} \\
		& Nondeterministic Comm. \cite{ram09,gossipdl,gossipgrad}\\\addlinespace
		\multicolumn{2}{ l }{\bf Parameter Distribution and Communication} \\
		Centralization & Parameter Server (PS) \cite{distps,geeps,deepspark,firecaffe} \\
		& Sharded PS \cite{le12,dean12,adam,petuum,poseidon15,poseidon17,kurth17,dynsgd17} \\
		& Hierarchical PS \cite{rudra16,yu16supercomputer,gaia}\\
		& Decentralized \cite{decentralized,gossipdl,scaffe17,gossipgrad}\\\addlinespace
		Compression & Quantization \cite{flexpoint,eightbit,qnn,gupta15,xnornet,dorefanet,twn,seide14onebit,binaryconnect,tamingthewild,diannao,alistarh17,deepcompression,courbariaux16,terngrad}\\
		& Sparsification \cite{strom15,privpres,graddrop,adacomp,dryden16,lin18,sparcml}\\
		& Other Methods \cite{sfb,poseidon15,squeezenet,deeprebirth,xception,paramcomp,mobilenets} \\\addlinespace
		\multicolumn{2}{ l }{\bf Training Distribution} \\
		Model Consolidation 
		& Ensemble Learning \cite{instantlearning, gap, treenets} \\
		& Knowledge Distillation \cite{ba14mimic,hinton15distill} \\
		&  Model Averaging: Direct \cite{parsgd,miao14,bmuf,byzantine}, \\
		&  \hspace{3em} Elastic \cite{easgd,gossipdl,decentralized,you17}, Natural Gradient \cite{kfac,natgrad} \\\addlinespace
		Optimization Algorithms
		& First-Order \cite{lecun98,le11,admmddl,svrg,chung17,psodl,proggan}\\
		& Second-Order \cite{hessianfree,disthf,byrd16slbfgs,moritz16slbfgs,dean12,neumannopt,kfac}\\
		& Evolutionary \cite{such17,geneticcnn,hyperneat,evolvingdnns} \\
		& Hyper-Parameter Search \cite{snoek12,klein16,baker17,hazan18,omnivore,yellowfin,psodl,evolvingdnns,pbt} \\
		& Architecture Search: Reinforcement \cite{nasnet,nasrl,enas,metaqnn,zhong17},\\
		& \hspace{8.35em} Evolutionary \cite{real17evo,regevo,geneticcnn,liu18,young17},\\
		& \hspace{8.35em} SMBO \cite{nash,pnasnet,negrinho17,darts,smash}\\
		\bottomrule
	\end{tabular}
	\caption{Overview of Distributed Deep Learning Methods}
	\label{tbl:ddl}
\end{minipage}
\hfil
\begin{minipage}[b]{.3\textwidth}
	\centering
	\includegraphics[width=0.75\textwidth,clip,trim={0cm 3cm 0cm 0cm}]{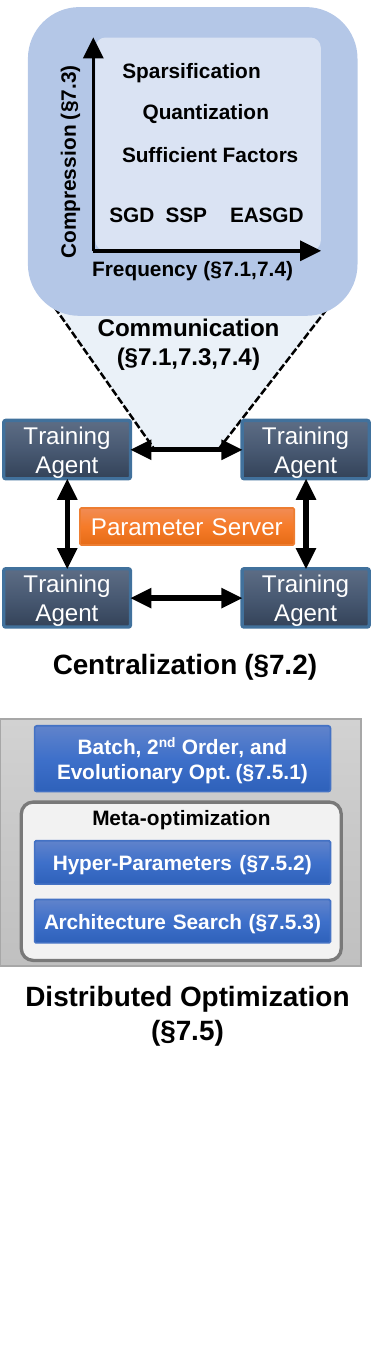}
	\caption{Section Overview}
	\label{fig:distoverview}
\end{minipage}
\vspace{-1em}
\end{figure}

So far we have discussed training algorithms where there is only one copy of $w$, and its up-to-date value is directly visible to all processors. In distributed environments, there may be multiple instances of SGD (\textit{training agents}) running independently, and thus the overall algorithm has to be adapted. 
Distribution schemes for deep learning can be categorized along three axes: \textbf{model consistency}, \textbf{parameter distribution}, and \textbf{training distribution}; where Figures \ref{tbl:ddl} and \ref{fig:distoverview} summarize the applied techniques and optimizations.

\subsection{Model Consistency}
\label{sec:dist:consistency}

\begin{figure}[t]
	\centering
	\begin{tabular}{ c c }
		\begin{subfigure}[b]{0.45\linewidth}
			\centering
			\includegraphics[height=1.2in,trim={0cm 0cm 6.5cm 0cm},clip,page=1]{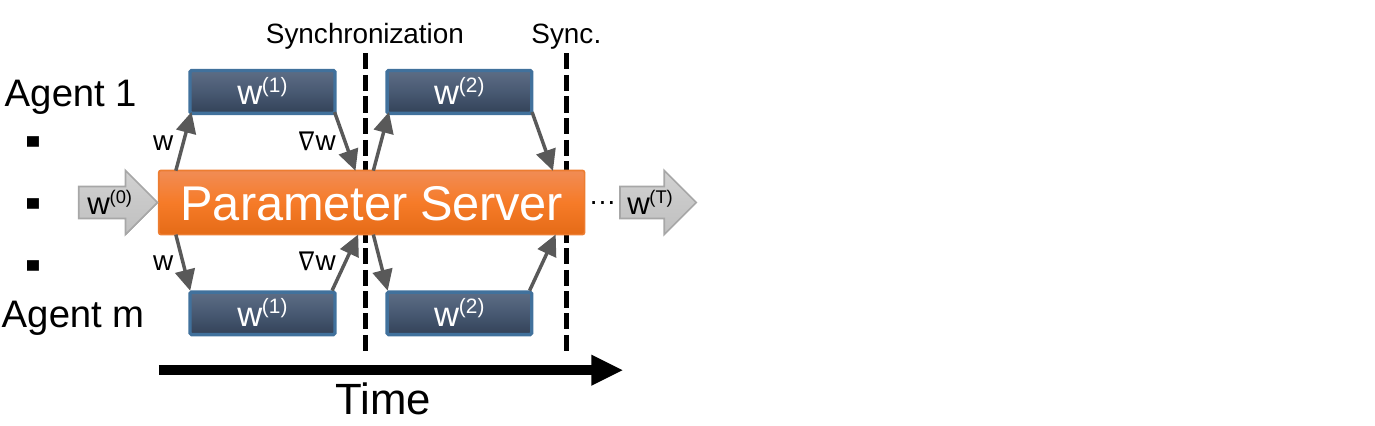}
			\caption{Synchronous, Parameter Server}
			\label{fig:dist:cons-ps}
		\end{subfigure}
		&
		\begin{subfigure}[b]{0.45\linewidth}
			\centering
			\includegraphics[height=1.2in,trim={0cm 0cm 7.25cm 0cm},clip,page=2]{figures/dist.pdf}
			\caption{Synchronous, Decentralized}
			\label{fig:dist:cons-decent}
		\end{subfigure}\\
		\begin{subfigure}[b]{0.45\linewidth}
			\centering
			\includegraphics[height=1.2in,trim={0cm 0cm 6.5cm 0cm},clip,page=3]{figures/dist.pdf}
			\caption{Asynchronous, Parameter Server}
			\label{fig:dist:incons-async}
		\end{subfigure}
		&
		\begin{subfigure}[b]{0.45\linewidth}
			\centering
			\includegraphics[height=1.2in,trim={0cm 0cm 6.5cm 0cm},clip,page=4]{figures/dist.pdf}
			\caption{Stale-Synchronous, Decentralized}
			\label{fig:dist:incons-ssp}
		\end{subfigure}
	\end{tabular}
	\vspace{-1em}
	\caption{Training Distribution in Deep Learning (Model Consistency, Centralization)}
	\label{fig:dist}
	\vspace{-0.9em}
\end{figure}

We denote training algorithms in which the up-to-date $w$ is observed by everyone as \textit{consistent model} methods (See Figures \ref{fig:dist:cons-ps} and \ref{fig:dist:cons-decent}). Directly dividing the computations among nodes creates a distributed form of data parallelism (Section \ref{sec:par}), where all nodes have to communicate their updates to the others before fetching a new minibatch. To support distributed, data parallel SGD, we can modify Algorithm \ref{alg:sgd} by changing lines \ref{alg:sgd:readweights} and \ref{alg:sgd:writeweights} to read (write) weights from (to) a parameter store, which may be centralized or decentralized (see Section \ref{sec:dist:centralization}). This incurs a substantial overhead on the overall system, which hinders training scaling.

Recent works relax the synchronization restriction, creating an \textit{inconsistent model} (Fig. \ref{fig:dist:incons-async}). As a result, a training agent $i$ at time $t$ contains a copy of the weights, denoted as $w^{(\tau,i)}$ for $\tau\le t$, where $t-\tau$ is called the \textit{staleness} (or lag). A well-known instance of inconsistent SGD is the HOGWILD shared-memory algorithm \cite{hogwild}, which allows training agents to read parameters and update gradients at will, overwriting existing progress. HOGWILD has been proven to converge for sparse learning problems \cite{hogwild}, where updates only modify small subsets of $w$, and generally \cite{tamingthewild}. Based on foundations of distributed asynchronous SGD \cite{distasgd}, the proofs impose that (a) write-accesses (adding gradients) are always atomic; (b) Lipschitz continuous differentiability and strong convexity on $f_w$; and (c) that the staleness, i.e., the maximal number of iterations between reading $w$ and writing $\nabla w$, is bounded.

The HOGWILD algorithm was originally designed for shared-memory architectures, but has since been extended \cite{dogwild,dean12} to distributed-memory systems, in which it still attains convergence for deep learning problems. To mitigate the interference effect of overwriting $w$ at each step, the implementation transfers the gradient $\nabla w$ instead of $w$ from the training agents. Asymptotically, the lack of synchronization in HOGWILD and its gradient-communicating variants admits an optimal SGD convergence rate of $\mathcal{O}(1/\sqrt{mT})$ for $m$ participating nodes \cite{agarwalduchi,dekel12,asysg}, as well as linear scaling, as every agent can train almost independently. 

To provide correctness guarantees in spite of asynchrony, Stale-Synchronous Parallelism (SSP) \cite{ssp13} proposes a compromise between consistent and inconsistent models. In SSP (Fig. \ref{fig:dist:incons-ssp}), the gradient staleness is enforced to be bounded by performing a global synchronization step after a maximal staleness may have been reached by one of the nodes. This approach works especially well in heterogeneous environments, where lagging agents (stragglers) are kept in check. To that end, distributed asynchronous processing has the additional advantage of adding and removing nodes on-the-fly, allowing users to add more resources, introduce node redundancy, and remove straggling nodes \cite{dean12,ppedl}.

In practical implementations, the prominently-used model consistency approaches are synchronous for up to 32--50 nodes \cite{goyal17,baiduhpc}, where the allreduce operation still scales nearly linearly; and asynchronous/SSP for larger clusters and heterogeneous environments \cite{dynsgd17,gaia,oyama16,zhang16}.

\subsection{Centralization}
\label{sec:dist:centralization}

\begin{figure}[t]
	\centering
	\begin{subfigure}[b]{0.3\textwidth}
		\centering
		\includegraphics[height=1in]{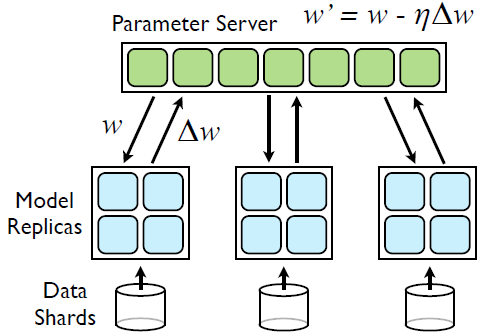}		
		\caption{DistBelief Sharded PS \cite{dean12}}
		\label{fig:ps:sharded}
	\end{subfigure}
	\quad
	\begin{subfigure}[b]{0.3\textwidth}
		\centering
		\includegraphics[height=1in]{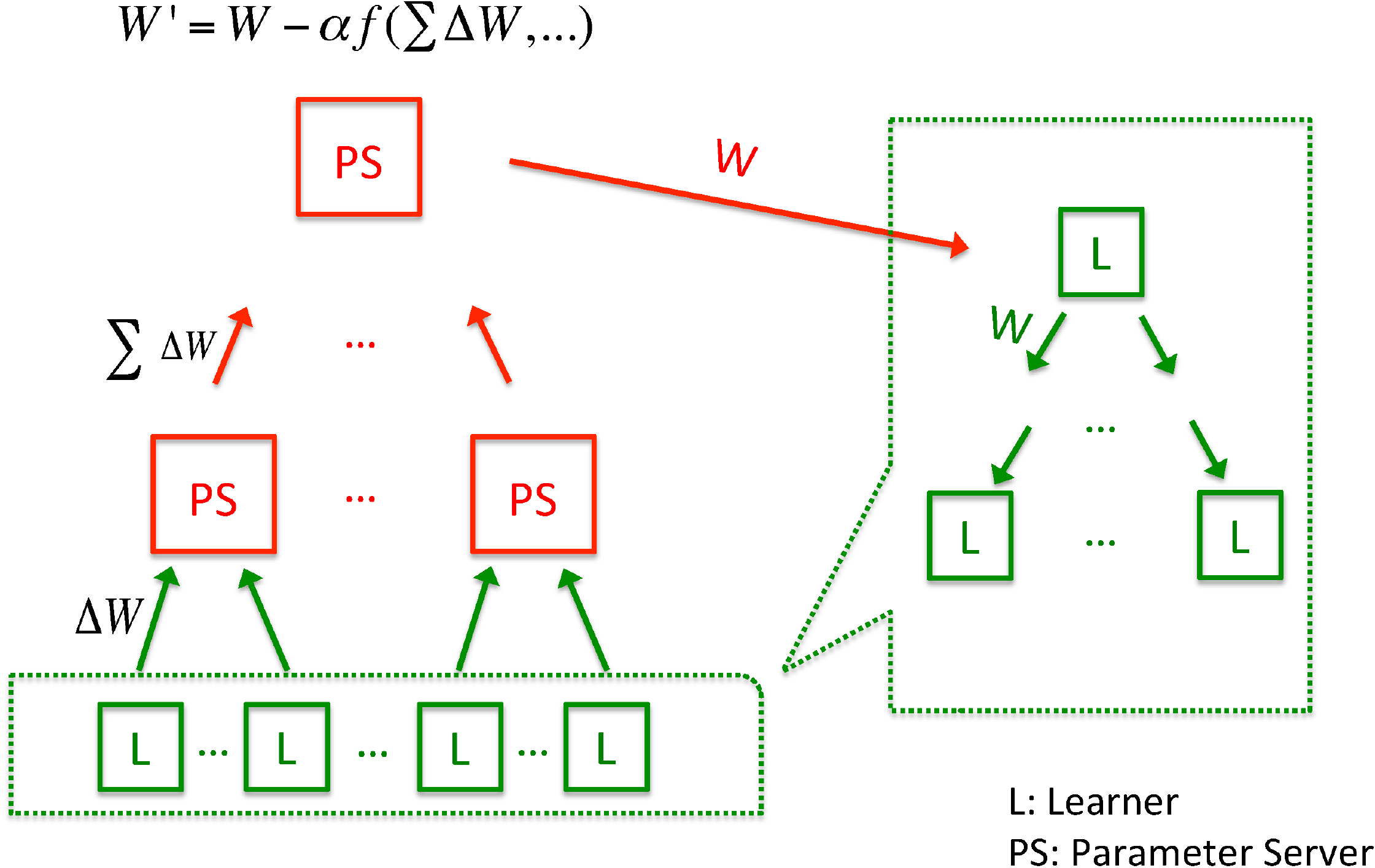}
		\caption{Rudra Hierarchical PS \cite{rudra16}}
		\label{fig:ps:hierarchy}
	\end{subfigure}
	\quad
	\begin{subfigure}[b]{0.3\textwidth}
		\centering
		\includegraphics[height=1in]{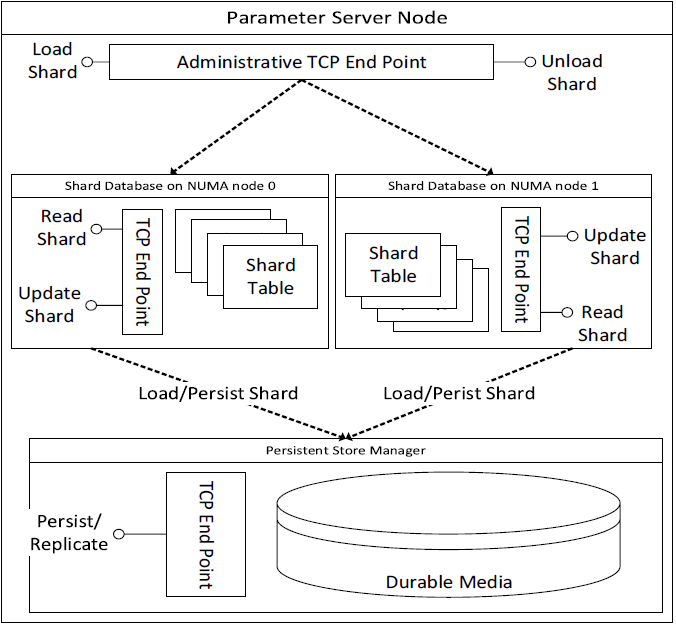}
		\caption{Project Adam PS \cite{adam}}
		\label{fig:ps:ft}
	\end{subfigure}
	\vspace{-1em}
	\caption{Parameter Server Infrastructures}
	\label{fig:ps}
	\vspace{-1em}
\end{figure}

The choice between designing a centralized and a decentralized network architecture for DNN training depends on multiple factors \cite{decentralized}, including the network topology, bandwidth, communication latency, parameter update frequency, and desired fault tolerance. A centralized network architecture would typically include a \textit{parameter server} (PS) infrastructure (e.g., Figures \ref{fig:dist:cons-ps}, \ref{fig:dist:incons-async}, \ref{fig:ps}), which may consist of one or more specialized nodes; whereas a decentralized architecture (Figures \ref{fig:dist:cons-decent}, \ref{fig:dist:incons-ssp}) would rely on allreduce to communicate parameter updates among the nodes. Following communication, centralized parameter update is performed by the PS, whereas the decentralized update is computed by each node separately. In the latter case, every node creates its own optimizer.

The tradeoff between using either distribution scheme can be modeled by the communication cost per global update. While the allreduce operation can be implemented efficiently for different message sizes and nodes (Section \ref{sec:term:paralg}), the PS scheme requires each training agent to send and receive information to/from the PS nodes. Thus, not all network routes are used, and in terms of communication the operation is equivalent to a reduce-then-broadcast implementation of allreduce, taking $T_{\mathit{tree}}$ time. On the other hand, the PS can keep track of a ``global view'' of training, averaging the gradients at one location and enabling asynchronous operation of the agents. This, in turn, allows nodes to communicate less information by performing some of the computations on the PS \cite{adam}, as well as increases fault tolerance by dynamic spin-up and removal of nodes during training.

The PS infrastructure is an abstract concept, and is not necessarily represented by one physical server. Sharded parameter servers \cite{dean12,adam} divide the ownership of $w$ over multiple nodes, each containing a segment of its elements. In conjunction with model parallelism and layer pipelining (Sections \ref{sec:par:model} and \ref{sec:par:pipeline}), this alleviates some of the congestion at the PS, as shown in Fig. \ref{fig:ps:sharded}, in which each portion of a ``model replica'' (training agent) transmits its gradients and receives its weights from a different shard. Hierarchical parameter servers \cite{rudra16, yu16supercomputer} (Fig. \ref{fig:ps:hierarchy}) further alleviate resource contention by assigning training agents with PS ``leaves'', propagating weights and gradients from specific agent groups up to the global parameter store. Rudra \cite{rudra16} also studies the tradeoff in allowed staleness, number of agents, and minibatch size, showing that SSP performs better, but requires adapting the learning rate accordingly.

A PS infrastructure is not only beneficial for performance, but also for fault tolerance. The simplest form of fault tolerance in machine learning is checkpoint/restart, in which $w^{(t)}$ is periodically synchronized and persisted to a non-volatile data store (e.g., a hard drive). This is performed locally in popular deep learning frameworks, and globally in frameworks such as Poseidon \cite{poseidon17}. Besides checkpoints, fault tolerance in distributed deep learning has first been tackled by DistBelief \cite{le12,dean12}. In the system, training resilience is increased by both introducing computational redundancy in the training agents (using different nodes that handle the same data), as well as replicating parameter server shards. In the former, an agent, which is constructed from multiple physical nodes in DistBelief via hybrid parallelism (Section \ref{sec:par:hybrid}), is assigned multiple times to separate groups of nodes. Allocating redundant agents enables handling slow and faulty replicas (``stragglers'') by cancelling their work upon completion of the faster counterpart. As for the latter resilience technique, in DistBelief and Project Adam \cite{adam}, the parameters on the PS are replicated and persisted on non-volatile memory using a dedicated manager, as can be seen in Fig. \ref{fig:ps:ft}. Project Adam further increases the resilience of distributed training by using separate communication endpoints for replication and using Paxos consensus between PS nodes.

Applying weight updates in a distributed environment is another issue to be addressed. In Section \ref{sec:term:sl}, we establish that all popular weight rules are first-order with respect to the required gradients (Table \ref{tbl:wupdate}). As such, both centralized and decentralized schemes can perform weight updates by storing the last gradient and parameter values. Since GPUs are commonly used when training DNNs (Fig. \ref{fig:devs}), frameworks such as GeePS \cite{geeps} implement a specialized PS for accelerator-based training agents. In particular, GeePS incorporates additional components over a general CPU PS, including CPU-GPU memory management components for weight updates. 

In addition to reducing local (e.g., CPU-GPU) memory copies, PS infrastructures enable reducing the amount of information communicated over the network. Project Adam utilizes the fact that the PS is a compute-capable node to offload computation in favor of communicating less. In particular, it implements two different weight update protocols. For convolutional operators, in which the weights are sparse, gradients are communicated directly. However, in fully connected layers, the output of the previous layer $x\in X^{C_{in}\times N}$ and error $\frac{\partial \ell}{\partial y}\in X^{C_{out}\times N}$ are transmitted instead, and $\nabla w$ is computed on the PS. 
Therefore, with $m$ nodes communication is modified from $m\cdot C_{out}\cdot C_{in}$ to $m\cdot N\cdot (C_{out} + C_{in})$, which may be significantly smaller, and balances the load between the agents and the normally under-utilized PS. 

Parameter servers also enable handling heterogeneity, both in training agents \cite{dynsgd17} and in network settings (e.g., latency) \cite{gaia}. The former work models distributed SGD over clusters with heterogeneous computing resources, and proposes two distributed algorithms based on stale-synchronous parallelism. Specifically, by decoupling global and local learning rates, unstable convergence caused by stragglers is mitigated. The latter work \cite{gaia} acknowledges that training may be geo-distributed, i.e., originating from different locations, and proposes a hierarchical PS infrastructure that only synchronizes ``significant'' (large enough gradient) updates between data centers. To support this, the Approximate Synchronous Parallel model is defined, proven to retain convergence for SGD, and empirically shown to converge up to 5.6$\times$ faster with GoogLeNet.

In a decentralized setting, load balancing can be achieved using asynchronous training. However, performing the parameter exchange cannot use the allreduce operation, as it incurs global synchronization. One approach to inconsistent decentralized parameter update is to use Gossip Algorithms \cite{boyd05}, in which each node communicates with a fixed number of random nodes (typically in the order of 3). With very high probability \cite{drezner86}, after communicating for $1.639\cdot log_2 m$ time-steps, where $m$ is the number of nodes, the data will have been disseminated to the rest of the nodes. As strong consistency is not required for distributed deep learning, this method has shown marginal success for SGD \cite{ram09,gossipdl,gossipgrad}, attaining both convergence and faster performance than allreduce SGD up to 32 nodes. On larger systems, the resulting test accuracy degrades. One approach to improve this could be to employ deterministic correction protocols \cite{corrgoss}.

\subsection{Parameter and Gradient Compression}
\label{sec:dist:compression}

The distributed SGD algorithm requires global reduction operations to converge. As discussed above, reducing the \textit{number} of messages (via an inconsistent view of $w$ or efficient collective operations) is possible. Here, we discuss reducing the \textit{size} of each message.

There are two general ways to conserve communication bandwidth in distributed deep learning: compressing the parameters with efficient data representations, and avoiding sending unnecessary information altogether, resulting in communication of sparse data structures. While the methods in the former category are orthogonal to the network infrastructure, the methods applied in the latter category differ when implemented using centralized (PS) and decentralized topologies.

\subsubsection{Quantization}

\begin{figure}[t]
	\centering
	\begin{subfigure}[b]{0.44\textwidth}
		\centering
		\includegraphics[height=1.2in]{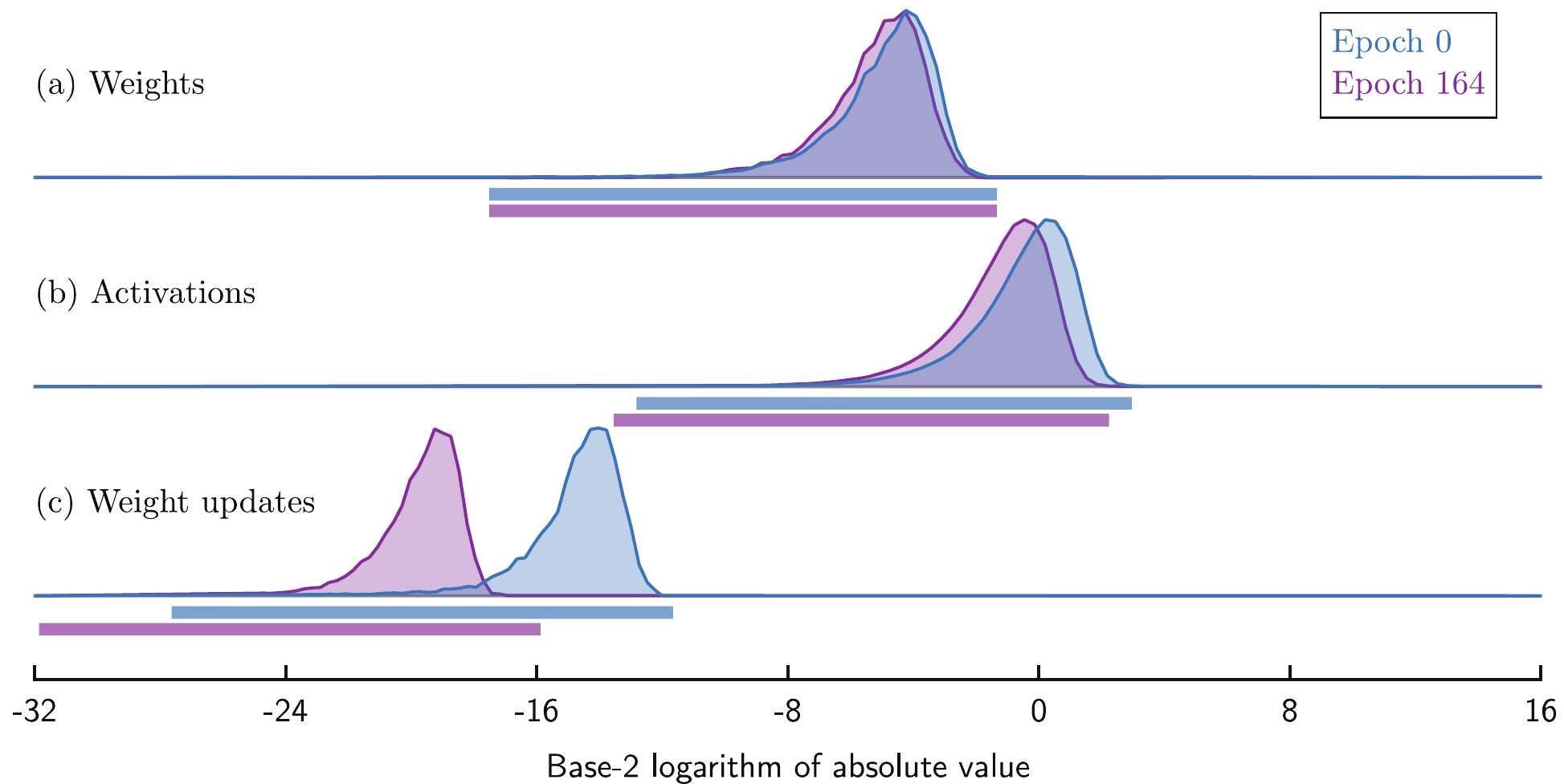}
		\caption{Parameter and Gradient Value Distribution (ResNet on CIFAR-10, adapted from \cite{flexpoint})}
		\label{fig:quant:flexpoint}
	\end{subfigure}
	\quad
	\begin{subfigure}[b]{0.52\textwidth}
		\centering
		\includegraphics[width=.3\linewidth,clip,trim={0cm 0cm 0cm 0cm}]{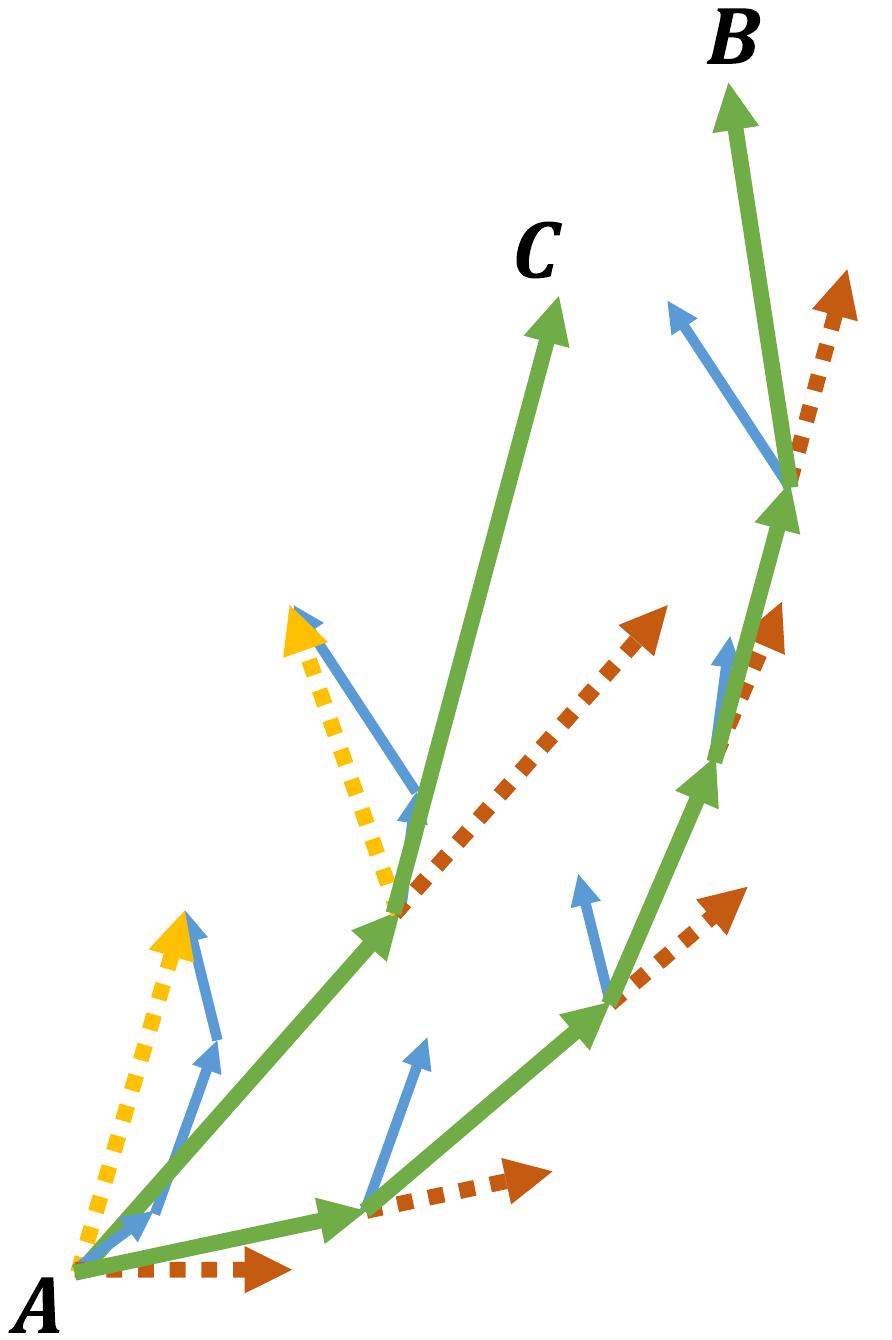}
		\includegraphics[width=.3\linewidth,clip,trim={0cm 0cm 0cm 0cm}]{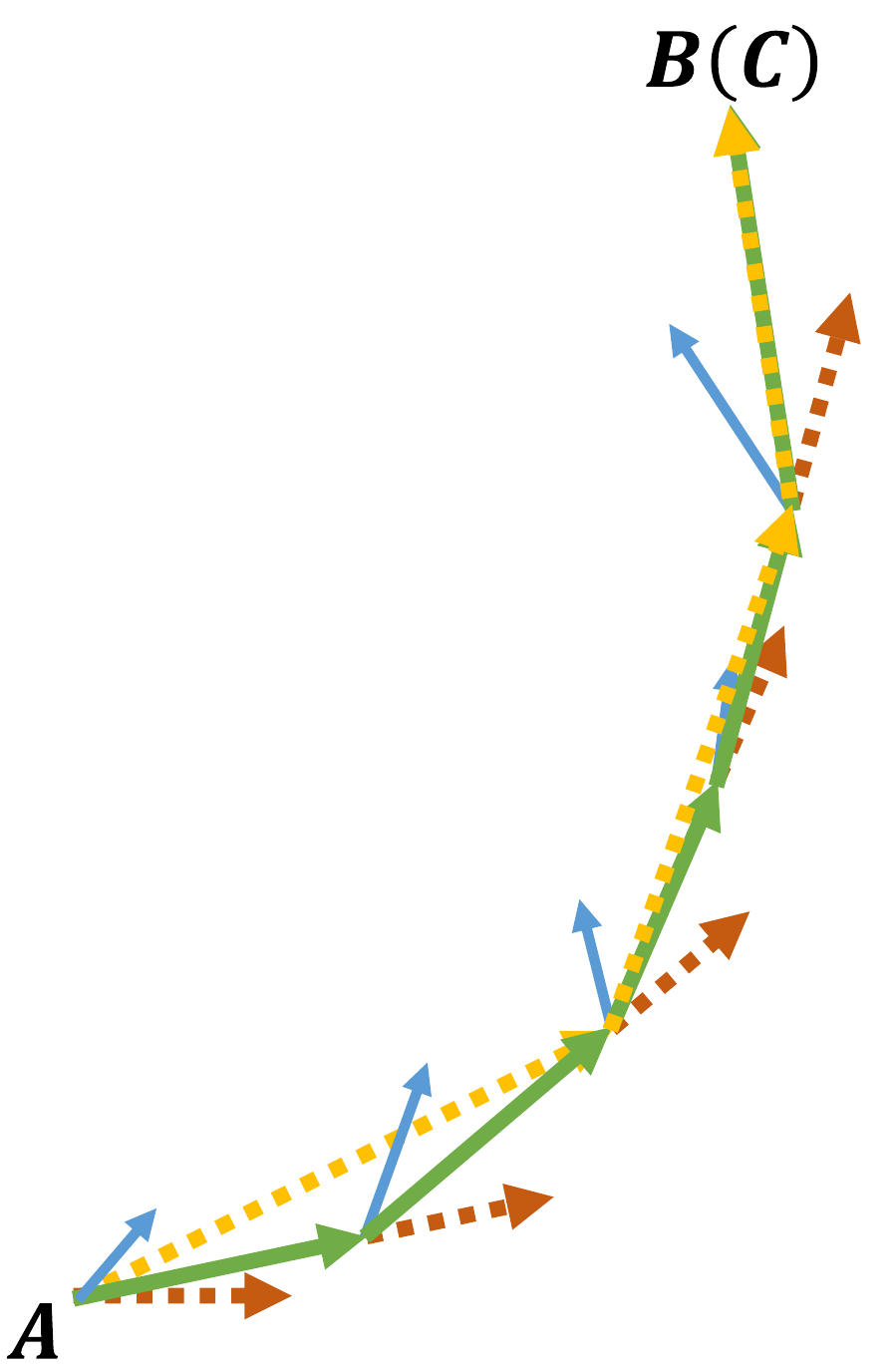}
		\includegraphics[height=0.6in,clip,trim={0cm 10.5cm 21.6cm 0cm}]{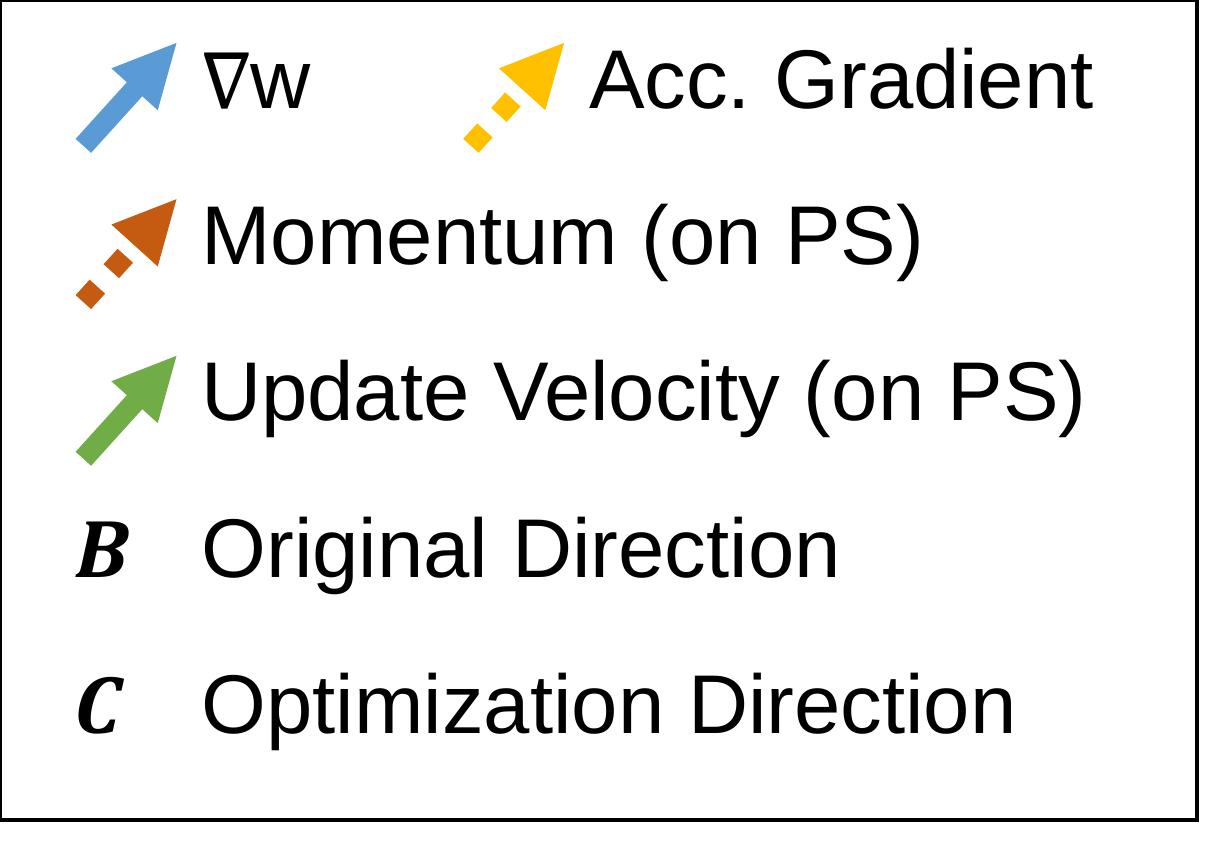}
		\caption{Local Gradient Accumulation without (Left) and with (Right) Momentum Correction (adapted from \cite{lin18})}
		\label{fig:quant:momacc}
	\end{subfigure}
	\vspace{-1em}
	\caption{Parameter and Gradient Quantization}
	\label{fig:quant}
	\vspace{-1em}	
\end{figure}

A prominent data representation for gradient (or parameter) compression is quantization, i.e., mapping continuous information into buckets that represent sets of values (usually ranges). It has been shown \cite{flexpoint} that the distributions of parameter and gradient values are narrowly dispersed (Fig. \ref{fig:quant:flexpoint}), thus these methods are effective in representing the working range to reduce the number of bits per parameter. This method has been successfully utilized in deep learning, both during training \cite{eightbit,qnn,gupta15} and for inference, where values are quantized post-training \cite{xnornet,dorefanet}. Some papers go so far as to quantize gradients to binary \cite{seide14onebit, binaryconnect} or ternary \cite{twn} values, while still attaining convergence with marginally reduced accuracy.

Quantization is commonly performed by way of reducing floating-point dynamic range \cite{gupta15,eightbit,tamingthewild}. In particular, such methods represent IEEE 754 32-bit floating-point values with fewer bits. While already applied to inference hardware \cite{diannao}, the first successful instance of reduced precision for training \cite{gupta15} was performed with IEEE 754 16-bit float values (``half precision''). As evaluated in the paper, quantized training does not work ``out-of-the-box'' for lossy compression such as reduced precision. Rather, it depends on rounding the parameters in a way that preserves the expected value ($\e$) of the parameters. To resolve this issue, the paper proposes Stochastic Rounding \cite{gupta15}, which randomly rounds numbers down or up, providing correct values in expectation.

Other forms of quantization extend upon these ideas. QSGD \cite{alistarh17} generalizes stochastic rounding to stochastic quantization, and proposes multi-level gradient quantization schemes. Deep Compression \cite{deepcompression} also employs the lossless Huffman Coding \cite{huffman} to further increase storage efficiency without impairing convergence. Binarized Neural Networks (BNNs) \cite{courbariaux16}, Ternary Weight Networks \cite{twn},  TernGrad \cite{terngrad}, and DoReFa-Net \cite{dorefanet} quantize networks to binary parameters, ternary parameters, ternary gradients, and binary parameters+ternary gradients, respectively. Both BNNs (in some cases) and TernGrad use stochastic rounding to lower the input representation. Lastly, FlexPoint \cite{flexpoint} implements block floating-point arithmetic \cite{blockfp}, which computes the mantissa portion of the floating-point values as fixed-point math, and shares the exponent part among multiple parameters/gradients. To accommodate changes to the exponents, predictive analysis is used for estimating subsequent values.

Essential to the convergence of SGD with lossy quantization is local gradient accumulation. Particularly in distributed environments, where the updates are inconsistent, it is important to carry the quantization error to the next gradient, accumulating error to avoid drift. The idea originates from Sigma-Delta Modulation \cite{seide14onebit}, and has proven to be successful in many cases. Deep Gradient Compression \cite{lin18} extends this idea by correcting momentum as well (Fig. \ref{fig:quant:momacc}), further decreasing the loss in accuracy to become non-negligible, and even resulting in a minor accuracy increase.

\subsubsection{Sparsification}

DNNs (and CNNs in particular) exhibit sparse gradients
during parameter updates. This is primarily due to the very large number
of parameters that do not necessarily change all at once; and operators
such as convolutions, in which the optimization process may
improve the accuracy of certain convolution kernels.  Therefore, the
full gradient is not necessary to retain convergence, and various
methods that leverage this feature have been proposed.

The first application of gradient sparsification \cite{strom15} prunes
gradient values using a static threshold, below which an element should
not be sent. Results show up to 54$\times$ speedup for 80 nodes and even
an up to 1.8\% reduction in error. The authors achieved a compression
ratio (which also includes 32-bit fixed point quantization) of
846--2,871$\times$ for a non-convolutional DNN. Subsequent works propose
relative  (e.g., top 1\%) \cite{privpres,graddrop,adacomp} and adaptive
thresholds \cite{dryden16} to transmit only the ``important'' gradients,
based on their absolute value. To counter the accuracy loss as a result
of sparsification, some works suggest to condition gradient values by
changing the DNN architecture, adding various normalization operators
\cite{graddrop}; whereas others \cite{lin18} propose local
gradient clipping (Section \ref{sec:dnn:bprop}) and warm-up training. 

In a centralized setting (Section \ref{sec:dist:centralization}), distributing sparse gradients is straightforward --- sparse
messages are sent between the training agents and the PS. 
However, implementing the necessary allreduce in a decentralized setting
is not as simple because each agent may contribute different non-zero
indices (dimensions) in its gradient. 
Kylix \cite{kylix} implements sparse allreduce in two steps, first
exchanging the indices and then the data.  While this is desirable for
systems where the sparsity pattern per node does not change, in deep
learning the gradient indices differ with each iteration. 
SparCML~\cite{sparcml} targets the specifics of deep learning explicitly by
supporting arbitrarily changing indices in a framework for sparse
allreduce operations. SparCML combines sending only the top-k most
significant indices with quantization and supports sparse vectors of
heterogeneous sizes. The system switches between a sparse and dense
representation automatically, informed by a simple performance model.
SparCML achieves a speedup of more than 20$\times$ over a well-tuned CNTK
implementation on Ethernet.

\subsubsection{Other Techniques}

In Section \ref{sec:dist:centralization}, we discuss Project Adam sending activations and errors instead of parameters, decreasing the overall footprint for fully connected layers in favor of redundant computation on the PS. The Poseidon (formerly Petuum) framework \cite{petuum,poseidon15,poseidon17} extends the idea of transmitting decomposed outer products $u\cdot v^T$ of $w$, generalizing the concept to other fields in machine learning as Sufficient Factor Broadcasting (SFB) \cite{sfb}. With SFB, the activations are not sent to the PS, but rather broadcast between the training agents for local recomposition.
SFB should work best in centralized topologies, as recomposing the gradients in a decentralized environment causes each agent to process $m-1$ additional outer products with each step, where $m$ is the number of agents. However, the authors claim \cite{sfb} that the cost of recomposition is negligible compared to communication. 

Since the decomposed weights are not additive, as opposed to gradients, SFB incurs all-to-all communication between training agents. To overcome scalability issues, the paper suggests partial broadcasting \cite{sfb}, where nodes communicate with a predetermined subset of the other nodes. By trading off gradient update latency ($\propto\mathbf{D}$) for bandwidth ($\propto\mathbf{W}$), the paper shows that convergence can still be attained, equating the delayed updates with stale gradients (Section \ref{sec:dist:consistency}).

A different approach to reduce DNN memory footprints is to design them specifically for that purpose \cite{squeezenet,deeprebirth,xception,paramcomp}. Such works make use of memory-efficient operators and techniques, mostly applied to convolutions, to train networks that fit on devices such as FPGAs and mobile phones. Applied techniques include layers constructed from a multitude of $1\times 1$ convolutions \cite{squeezenet}, reshaping \cite{deeprebirth} or applying Tucker Decomposition \cite{paramcomp} on convolution tensors, and separable convolutions (sequential application of reduced-dimension convolutions) \cite{mobilenets,xception}. The papers show that DNNs can decrease in size (up to 50$\times$) and evaluation time (6.13$\times$), exhibiting minor reduction in accuracy.

\subsection{Model Consolidation}

\begin{figure}[t]
	\centering
	\includegraphics[width=\linewidth,clip,trim={0cm 10cm 0cm 5cm}]{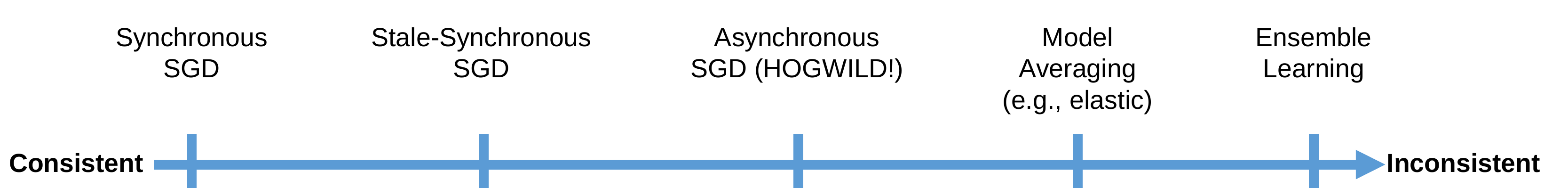}
	\caption{Parameter Consistency Spectrum}
	\label{fig:consistency}
	\vspace{-1.25em}
\end{figure}

In this section, we discuss the far (inconsistent) end of the parameter consistency spectrum (shown in Fig. \ref{fig:consistency}) in distributed deep learning. In such cases, parameter updates are highly infrequent (or nonexistent), and thus precautions must be taken with the received values.
In particular, rather than running data-parallel SGD on multiple nodes, distributed deep learning can be achieved by assigning training agents with different copies of $w$ and combining the resulting models, either post-training or several times during training. While the latter can be seen as a generalization of an inconsistent view of $w$, the former may entirely change the training and inference processes, depending on the method.

\subsubsection{Ensemble Learning and Knowledge Distillation}

A widely-used technique for post-training consolidation is \textit{ensemble learning} \cite{instantlearning, gap, treenets}. With ensembles, multiple instances of $w$ are trained separately on the same dataset, and the overall prediction is the average of the predictions of the ensemble members, i.e., $f(x)=\frac{1}{m}\sum_{i=0}^{m}f_{w^{(T,i)}}(x)$. Ensemble learning has been used extensively in machine learning before the deep learning era \cite{ensemble} as a form of boosting, and typically increases the overall accuracy over a single model. Thus, it is routinely applied in machine learning competitions such as ILSVRC \cite{imagenet} and in industrial applications.
Distributed training of ensembles is a completely parallel process, requiring no communication between the agents. However, works such as TreeNets \cite{treenets} (Section \ref{sec:par:model}) combine ensemble learning with custom (ensemble-aware) loss functions to promote diversity between ensemble members.

Given that ensembles consume a factor of $m$ more memory and compute power, another post-training model consolidation technique is to reduce the size of a DNN using \textit{knowledge distillation} \cite{ba14mimic,hinton15distill}. In this scheme, training is performed in two steps: in the first step, a large network or an ensemble is trained normally; and the second step trains a single neural network to mimic the output of the large ensemble. Results \cite{hinton15distill} show that the second network is easier to train on the ensemble than on a labeled dataset, attaining the same word error rate as an ensemble of 10 DNNs.

\subsubsection{Model Averaging}

Another technique for consolidating models is \textit{model averaging} \cite{avgsgd}. Such methods may separately run $m$ SGD instances on different machines, aggregating the parameters only once (post-training) \cite{parsgd} or every few iterations \cite{miao14,bmuf}. While these methods are proven to converge, applying stale-synchronous SGD (Section \ref{sec:dist:consistency}) leads to higher overall accuracy. 

To overcome accuracy degradation as a result of infrequent averaging, more sophisticated consolidation methods include Elastic Averaging SGD (EASGD) \cite{easgd} and Natural Gradient Descent \cite{natgrad}. EASGD is based on a centralized environment (i.e., including a PS), extending direct averaging by using elastic forces between the training agents' view of $w$ ($w^{(t,i)}$) and the PS's view ($\bar{w}$). This allows the agents to ``explore'' further by increasing the possible distance of each agent from the average, and also allows to communicate sparsely with respect to time (iterations). EASGD was reported \cite{easgd} to outperform the DistBelief \cite{dean12} SGD method in terms of accuracy, shown to be tolerant in terms of update delay, and was used successfully in practice for communication reduction by other works \cite{decentralized,you17}.

Natural Gradient Descent (NG-SGD) can also be used to deal with diverging parameters in different agents \cite{natgrad}. NG-SGD modifies SGD to define learning rate matrices, approximating the inverse Fisher information matrix and thus natural gradients. By averaging agent parameters only every $k$ samples (typically in the order of hundreds of thousands), the algorithm allows agents to gradually diverge and synchronize less than traditional SGD. Natural Gradients were also approximated for distributed deep learning using Kronecker Factorization (K-FAC) \cite{kfac}, where the work is divided between gradient- and statistics-computing agents (for Fisher matrix blocks).

In distributed settings, algorithms are also inspected w.r.t. fault tolerance. Krum \cite{byzantine} is a Byzantine fault-tolerant \cite{bgp} SGD algorithm, allowing up to $f$ Byzantine training agents. In particular, the paper shows that any gradient aggregation rule based on linear combination cannot sustain a single Byzantine agent. By combining specific $m-f$ gradients (that are closest to each other), Krum is able to overcome adversarial gradient inputs from the network.

\subsection{Optimization Algorithms and Architecture Search}
\label{sec:dist:opt}

As training in deep learning is a nonlinear optimization problem, other algorithms that exhibit concurrency can be used in SGD's stead \cite{bottou16opt}. Furthermore, it is possible to use excess computational power to perform meta-optimization, searching for better hyper-parameters and DNN architectures.

\subsubsection{Parameter Search}

Supervised learning can either be viewed as a stochastic optimization process that uses one or a minibatch of samples at a time, or it can be expressed as a batch optimization problem, where the entire dataset is necessary to obtain gradients for descent. Batch optimization has been used for deep learning since the inception of DNNs \cite{lecun98}, using first- and second-order methods \cite{nocedal06numopt} such as Levenberg-Marquardt, Conjugate Gradient (CG), and L-BFGS. Although considerably more computationally expensive than SGD, there are several advantages to such approaches, including increased concurrency (as data-parallelism increases) and better theoretical convergence guarantees (e.g., second-order methods converge locally at a quadratic rate). As mentioned in Sections \ref{sec:tradeoff} and \ref{sec:par:data}, large-minibatch training represents a middle ground between SGD and batch methods. Such methods combine the ``best of both worlds'' --- on one hand they exhibit increased inherent concurrency (as higher-order methods); and on the other hand they employ stochasticity, which, despite the sublinear rate of convergence, works well in practice.

For distributed deep learning, batch methods \cite{le11} (specifically CG and L-BFGS) and Hessian-free second-order optimization \cite{hessianfree,disthf,chung17} have initially been favored due to their apparent scalability compared to traditional SGD (Algorithm \ref{alg:ssgd}). However, due to the superior generalization properties of first-order stochastic optimization, and the successful DistBelief \cite{dean12} implementation of inconsistent SGD (called Downpour SGD, based on HOGWILD \cite{hogwild}); it was found that the quadratic increase of batch methods in memory, communication, and computations due to high dimensionality is not desirable. To overcome these issues, stochastic variants of L-BFGS have been proposed \cite{byrd16slbfgs,moritz16slbfgs} that estimate the inverse Hessian matrix and proven to converge at a linear rate in strongly-convex, Lipschitz-continuous settings \cite{moritz16slbfgs}.

Other optimization algorithms applied to deep learning attempt to: (a) reduce the variance of SGD incurred by random sampling \cite{svrg}, (b) use the Alternating Direction Method of Multipliers (ADMM) \cite{admm} to skip the backpropagation altogether \cite{admmddl}, and (c) use the Neumann series expansion to approximate the Hessian matrix \cite{neumannopt}, scaling to large minibatch sizes (32k) with no accuracy loss or substantial computational overhead.

Gradient-free evolutionary algorithms have also been employed for deep learning, where examples include Genetic Algorithms \cite{such17,geneticcnn}, Neuro-Evolution \cite{hyperneat,evolvingdnns}, and Particle-Swarm Optimization \cite{psodl}. Apart from recombination/evolution steps, training behavior is similar to ensemble learning, and thus these algorithms are more amenable to parallelism than traditional gradient descent. As we show in the rest of this section, the gradient-independent nature of such algorithms enable their use for meta-optimization of both hyper-parameters and DNN architectures.

\subsubsection{Hyper-Parameter Search}
\label{sec:training:hyperparams}

The multitude of hyper-parameters in SGD (e.g., learning rate, momentum, maximal staleness) and their adverse effect on the resulting accuracy hinders research efforts into new techniques in machine learning. Until recently, the prominent method for hyper-parameter search was to perform parameter sweeps (i.e., grid search over feasible ranges). Since this method increases the overall time exponentially with the number of hyper-parameters, its effectiveness is limited by the availability of computing power. 

Several methods try to expand beyond simple parameter sweeps by making educated guesses and tuning hyper-parameters during training. In the former class, methods include Bayesian optimization \cite{snoek12}, predictive analysis of the learning curves (e.g., training error, validation error) \cite{klein16,baker17} for dropping undesirable configurations, and sampling the hyper-parameter space efficiently using spectral methods such as Compressed Sensing \cite{hazan18}.

As for tuning hyper-parameters during training, Omnivore \cite{omnivore} employs predictive analysis and grid searches every predetermined number of minutes to modify the momentum and a hyper-parameter controlling local gradient staleness. The paper shows that in distributed environments, controlling the synchronous SGD node-group size during training can increase both accuracy and performance. YellowFin \cite{yellowfin} uses the local gradient curvature and variance to tune momentum, working especially well on LSTM-based models and asynchronous environments, performing up to 3.28$\times$ faster than the Adam optimizer (Table \ref{tbl:wupdate}).

Metaheuristic optimization algorithms can inherently integrate hyper-parameter tuning with training, and are thus used for DNNs. Such methods include Particle Swarm Optimization (PSO) based deep learning \cite{psodl}; and CoDeepNEAT \cite{evolvingdnns}, a modification of the NEAT algorithm that simultaneously searches for hyper-parameter and architecture configurations (see below). Such methods
scale almost linearly, due to abundance of independent computations.

Lastly, Population-Based Training \cite{pbt} (PBT) uses a reinforcement learning approach to ``explore'' and ``exploit'' the hyper-parameter space. In particular, each training agent independently samples (exploits) information from other agents every few SGD iterations. The information is then used to select the best configuration (e.g., using a t-test), and hyper-parameters are in turn perturbed (explored) to continue learning. This creates a decentralized topology where communication is nondeterministic (i.e., exploitation is performed with a randomly sampled agent), which may scale better as the number of training agents increases.

\subsubsection{Architecture Search}

Like feature engineering before the era of deep learning, manually crafting DNN architectures is naturally limited by human resourcefulness and creativity. This limitation promoted a recent rise of research into automated neural \textit{architecture search}.
Architecture search can be categorized into three approaches: Sequential Model-Based Optimization (SMBO), Reinforcement Learning (RL), and Evolutionary Algorithms (EA).

\begin{figure}[t]
	\centering
	\begin{subfigure}[t]{0.25\textwidth}
		\centering
		\includegraphics[height=1in,clip,trim={0cm 0cm 0.8cm 0cm}]{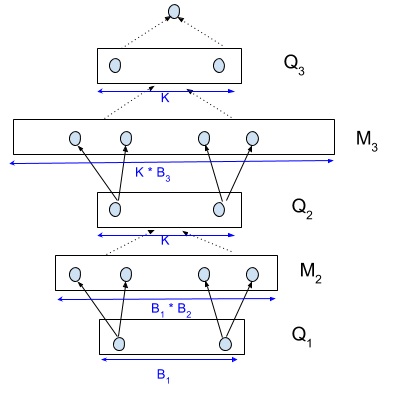}
		\caption{PNASNet Search-Space Traversal \cite{pnasnet}}
		\label{fig:archsearch:pnasnet}
	\end{subfigure}
	\quad
	\begin{subfigure}[t]{0.4\textwidth}
		\centering
		\includegraphics[width=\linewidth]{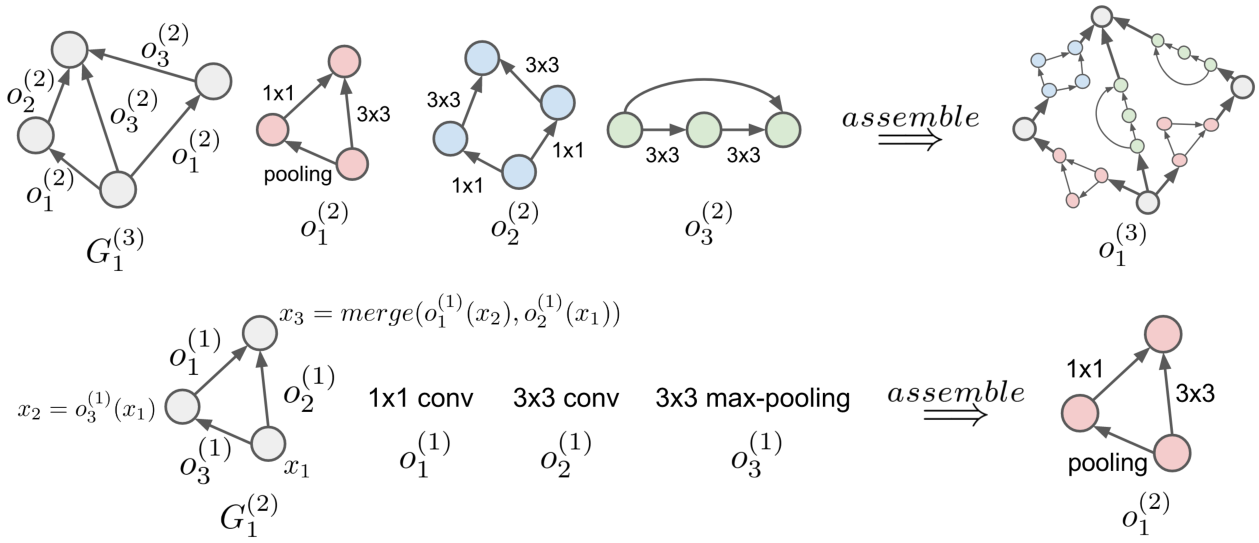}
		\caption{Hierarchical Representation \cite{liu18}}
		\label{fig:archsearch:hierarchical}
	\end{subfigure}
	\quad
	\begin{subfigure}[t]{0.25\textwidth}
		\centering
		\includegraphics[height=0.8in]{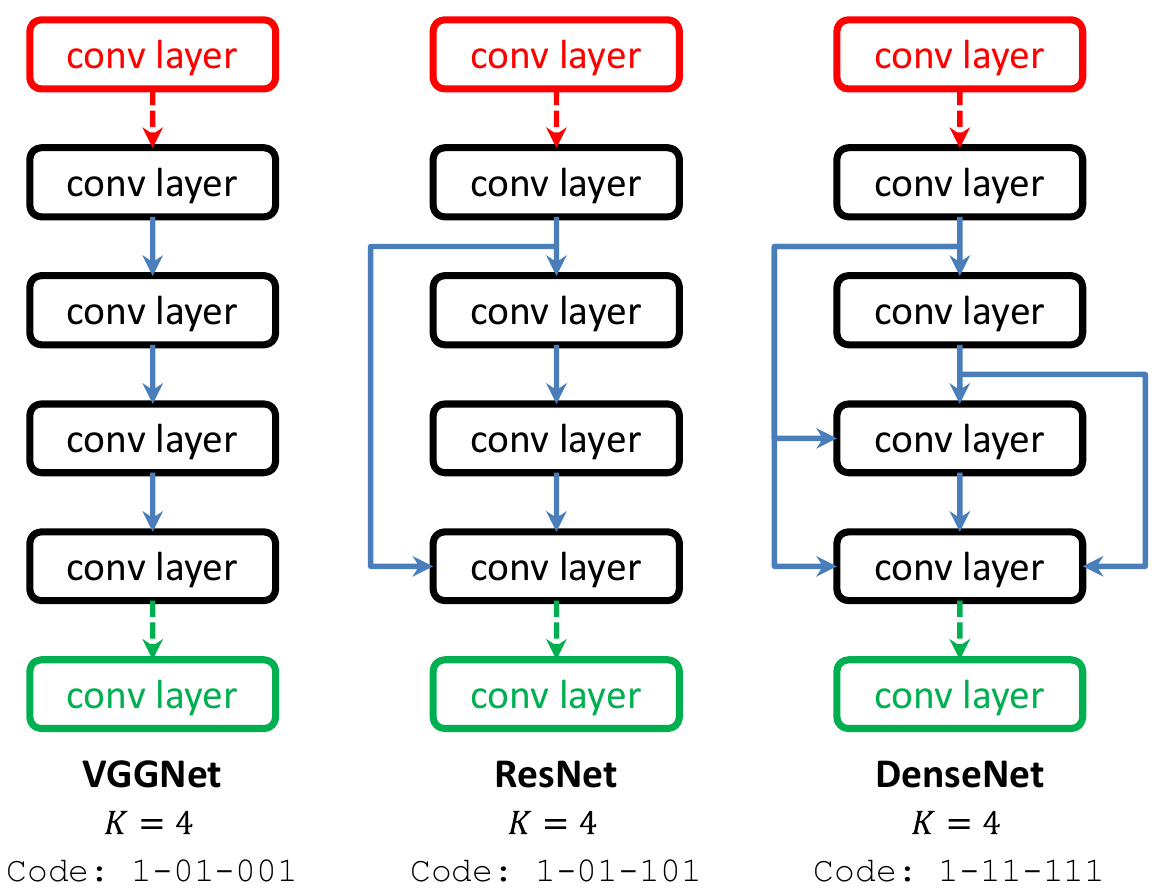}
		\caption{Genetic Encoding of DNN Connections \cite{geneticcnn}}
		\label{fig:archsearch:genecoding}
	\end{subfigure}
	\vspace{-1em}
	\caption{Methods for Automated Architecture Search}
	\label{fig:archsearch}
	\vspace{-2em}
\end{figure}

SMBO-based search methods rely on optimizing an architecture candidate, defining a finite set of states to explore (e.g., search tree children), and traversing those sets. As a result, concurrency depends on the number of points in the search space at a given time. Examples of SMBO include DeepArchitect \cite{negrinho17}, which proposes a DNN definition language that allows programmers to explicitly define the space; PNASNet \cite{pnasnet}, which searches for networks ordered by increasing complexity using a search algorithm based on A* (see Fig. \ref{fig:archsearch:pnasnet}), conserving half the evaluated models compared to an equivalent RL approach \cite{nasnet}; SMASH \cite{smash}, which assesses optimality (fitness) of candidate networks using another CNN that maps the given DNN architecture to weights for testing; and DARTS \cite{darts}, which formulates architecture search as a bi-level, differentiable optimization problem.

Many recent DNN architectures (Section \ref{sec:dnn:casestudies}) exhibit self-similarity and repeating sub-units (modules). 
This observation can be leveraged to dramatically reduce the number of explored architectures, composing networks hierarchically out of modules and basic blocks (e.g., convolution) as can be seen in Fig. \ref{fig:archsearch:hierarchical}. This approach has been used successfully in the community, creating new candidates for both CNN modules \cite{evolvingdnns,nasnet,zhong17,pnasnet,regevo,liu18} and RNN units \cite{nasrl,enas}.

RL-based architecture search uses the accuracy of the resulting network as a reward function, whereas modifications to the DNN or its hyper-parameters are actions. In Neural Architecture Search (NAS) \cite{nasrl}, the parameters of each layer can be modified, but the number of layers is fixed. A sharded PS-based distributed system, in conjunction with policy gradient optimization \cite{Williams1992}, is used for training. Other examples include MetaQNN \cite{metaqnn} and BlockQNN \cite{zhong17}, which operate similarly to NAS, but use Q-learning for optimization; and ENAS \cite{enas}, which significantly reduces computational time over NAS (by three orders of magnitude) by sharing parameters across children DNNs (i.e., networks in the immediate search space).

Evolutionary Algorithms (EA) are advantageous for architecture search, as any function (not necessarily differentiable) can be optimized using these methods. HyperNEAT was the first EA successfully applied \cite{hyperneat} to deep learning, used for training weights and DNN architecture at the same time; and  CoDeepNEAT \cite{evolvingdnns} defines a variant of the NEAT algorithm to optimize hyper-parameters and architecture, using the self-similarity feature of DNNs by optimizing ``blueprints'' that are composed of modules. Genetic CNNs \cite{geneticcnn} uses Genetic Algorithms (GAs) by encoding the DNN connections as binary genes (as required in GAs, shown in Fig. \ref{fig:archsearch:genecoding}), and training the population of DNNs with every time-step, using the final accuracy as the fitness function. GAs are highly amenable to parallelism, and have been successfully used for very large-scale training \cite{young17}, where 18,000 nodes were used on the Titan supercomputer for 24 hours to obtain state-of-the-art accuracy for segmentation and reconstruction problems.

Large-Scale Evolution \cite{real17evo} also uses GAs, but defines a set of specific mutations (e.g., insert convolution, alter stride) that can be applied. Large-Scale Evolution outperforms some existing RL-based methods in terms of accuracy, as well as in terms of scalability, as GAs can run the entire population in parallel (where accuracy increases with population size in expectation). However, in the general case GA requires synchronous reductive communication between time-steps for selection of the fittest candidates. To overcome this issue, the paper employs \textit{tournament selection} \cite{goldberg91}, which only performs pairwise comparisons between population members. 

Additional GA architecture search methods include the use of multi-level hierarchical representations of DNNs \cite{liu18} (Fig. \ref{fig:archsearch:hierarchical}), which implement an asynchronous distributed tournament selection (centralized, queue-based implementation) with specialized mutation. Regularized Evolution (AmoebaNets) \cite{regevo} further extends GA with tournament selection by removing the oldest sample from the population each iteration (akin to death in nature), thus regularizing the optimization process. AmoebaNets outperform all existing methods, including manually engineered DNNs and RL-based searches, with 3.8\% error for ImageNet and 2.13\% error for CIFAR-10 (compared to 5.29\% and 3.62\% on the best instances of DenseNet, see Table \ref{tbl:popdnns}).

\section{Concluding Remarks}
\label{sec:conclusions}

The world of deep learning is brimming with concurrency. Nearly every aspect of training, from the computation of a convolution to the meta-optimization of DNN architectures, is inherently parallel. Even if an aspect is sequential, its consistency requirements can be reduced, due to the robustness of nonlinear optimization, to increase concurrency while still attaining reasonable accuracy, if not better. In this paper, we give an overview of many of these aspects, the respective approaches documented in literature, and provide concurrency analysis using the W-D model when possible.

It is hard to predict what the future holds for this highly active field of research (many have tried over the years). Below, we highlight potential directions for future research in parallel and distributed deep learning.

As research progresses, DNN architectures are becoming deeper and more interconnected, between consecutive and non-consecutive layers (``skip connections''). Apart from accuracy, considerable effort is devoted to reducing the memory footprint and number of operations \cite{mobilenets,regevo}, in order to successfully run inference on mobile devices. This also means that post-training DNN compression \cite{deepcompression} will likely be researched further, and training compressible networks will be desirable. Since mobile hardware is limited in memory capacity and has to be energy efficient, specialized DNN computational hardware is frequently proposed \cite{sze17survey}. We see this trend with the NVIDIA Tensor Cores \cite{volta}, the Tensor Processing Unit \cite{tpu}, other ASICs and FPGAs \cite{diannao,nurvitadhi17}, and even neuromorphic computing \cite{truenorth}. Handling DNN sparsity (e.g., after compression) is a focus for some ASICs \cite{cambriconx}, and advances in recurrent networks and attention learning \cite{xu15,chan16} indicate that training and inference hardware would also need to work efficiently with variable-length inputs.

Computing individual operators is highly optimized today (Section \ref{sec:layercomp}), and thus current research is oriented towards inter-layer and whole-DNN optimization. TensorFlow XLA \cite{xla}, Tensor Comprehensions \cite{vasilache18}, Latte \cite{latte} and TVM \cite{tvm} compile entire neural network graphs at once, performing a variety of transformations (e.g., fusion) to optimize execution time, achieving 4$\times$ speedup over manually tuned individual operators. We expect research to continue in this direction to the point where DNN evaluation is close to optimal in terms of operations and shared-memory optimizations.

Techniques applied in distributed deep learning are converging to the point where a standard programming interface (or framework) can be designed. In the future, ecosystems such as Ease.ml \cite{easeml} may make the definition of a training scheme (e.g., with respect to centralization and gradient consistency) easier, hiding most of the low-level infrastructure setup. Combining the increasing support for cloud systems and elastic training \cite{ppedl} (where nodes can be spun up and removed at will) with the latest developments in evolutionary algorithms (see Section \ref{sec:dist:opt}), we may see adaptive and financially-viable optimization methods rising to prominence.

Finally, deep learning is being used to solve increasingly complex problems such as routing algorithms \cite{graves2016hybrid} and hierarchical task combination \cite{frans17}. Research towards Artificial General Intelligence is now focusing on multi-purpose networks \cite{kaiser17,zeroshot}, which creates new, unexplored opportunities for model parallelism and different training algorithms. Searching for adequate multi-purpose networks may be beyond the ingenuity of a human team, and as meta-optimization (specifically, architecture search) and progressive training \cite{proggan} increase in usability and quality; parameter sweeps and manual DNN architecture engineering will become obsolete. Supporting this claim is the fact that the current state-of-the-art CNN in computer vision \cite{regevo} (CIFAR-10 and ImageNet datasets) is the result of an automated architecture search. Exploiting parallelism is necessary for such breakthroughs and others, going hand in hand with the advancement of deep learning as a field.

\begin{acks}
T.B.N. is supported by the ETH Zurich Postdoctoral Fellowship and Marie Curie Actions for People COFUND program.
\end{acks}

\renewcommand*{\bibfont}{\ssmall}

{\sf
	\bibliographystyle{ACM-Reference-Format}
	\bibliography{references}
}

\newpage

\appendix

\section{Analysis of Influential Convolutional Neural Networks}
\label{app:dnntrends}
\begin{figure}[h]
	\includegraphics[height=0.8in]{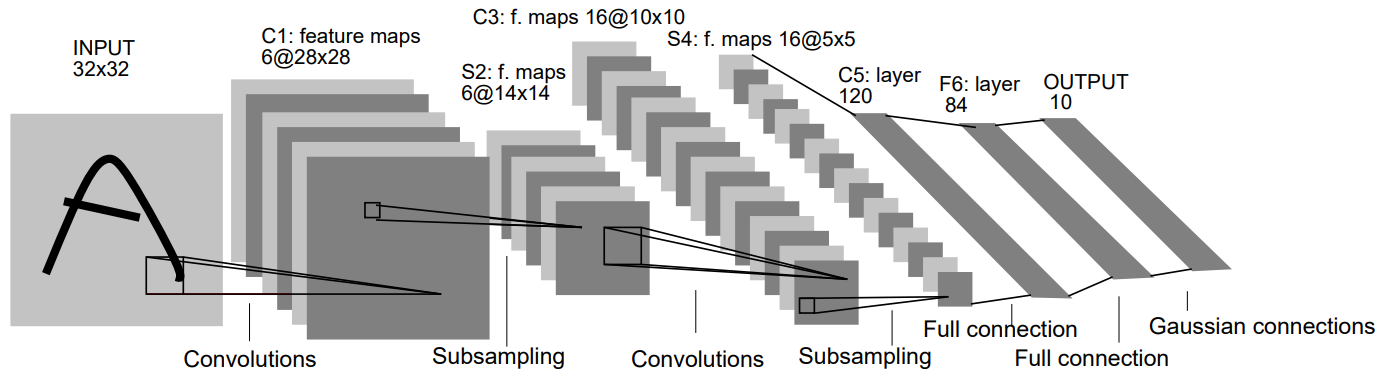}
	\caption{The LeNet-5 Convolutional Neural Network (adapted from \cite{lecun98})}
	\label{fig:net:lenet}
\end{figure}

\subsection{LeNet \cite{lecun98}}

The first successful convolutional neural network, which was designed to identify hand-written digits in the MNIST dataset \cite{mnistlecun,mnist}. As shown in Fig. \ref{fig:net:lenet}, LeNet-5 takes a single-channel 2D input, performs a series of 6 convolutions, then subsamples the filtered images by max-pooling. This convolution-pooling layer sequence occurs again, followed by 2 fully connected layers and a final fully connected softmax layer to produce the results.

For inference (forward evaluation) of an image, analyzing this network with respect to average parallelism ($\mathbf{W}/\mathbf{D}$) yields the following result (see Appendix \ref{app:layers} for details regarding each layer):
\[
\begingroup
\setlength\arraycolsep{2pt}
\begin{matrix*}[l]
\mathbf{W}&=&\mathbf{W}_{conv(32,5,1,6)}+\mathbf{W}_{pool(28,2,6)}+\mathbf{W}_{conv(14,5,6,16)}+\mathbf{W}_{pool(10,2,16)}~+\\
           &&\mathbf{W}_{fc(400,120)}+\mathbf{W}_{fc(120,84)}+\mathbf{W}_{fc(84,10)}\\
           &=& 117,600+14,112+470,400+4,800+48,000+10,080+840\\
           &=& 665,832
\\[6pt]
\mathbf{D}&=&5+2+9+2+9+7+7=41
\end{matrix*}
\endgroup
\]
This indicates that even the simplest DNN exhibits high levels of concurrency, linearly increasing with the minibatch size.

\begin{figure}[h!]
	\includegraphics[height=0.8in]{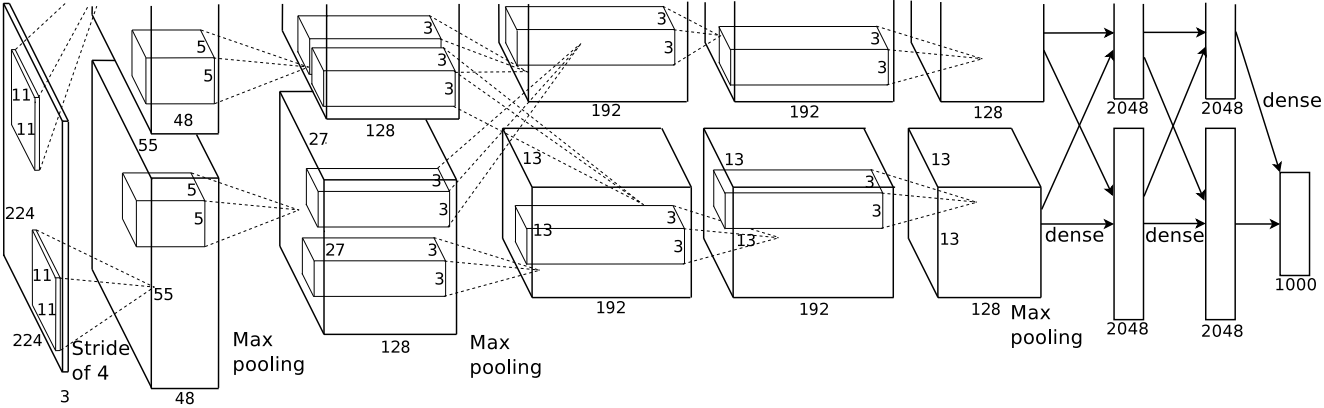}
	\caption{AlexNet (adapted from \cite{alexnet})}
	\label{fig:net:alexnet}
\end{figure}

\subsection{AlexNet \cite{alexnet}}

The AlexNet architecture was the winner of the ImageNet Large-Scale Visual Recognition (ILSVRC) 2012 \cite{imagenet} competition. Yielding nearly a twofold increase in accuracy over the preceding state-of-the-art (26.2\% top-5 error), this network played a major role in the current state of DNN and Machine Learning.

Similar to LeNet, AlexNet contains a series of convolution-pooling layers followed by fully connected layers. However, it also uses sequences of convolutions and Local Response Normalization layers. Two major factors in the success of AlexNet are data augmentation and network regularization (Dropout) during training. 

The network was implemented for GPUs (using a specialized cuda-convnet framework), training on ImageNet with a minibatch size of 128 images at a time.

\begin{figure}[h!]
	\includegraphics[height=0.8in]{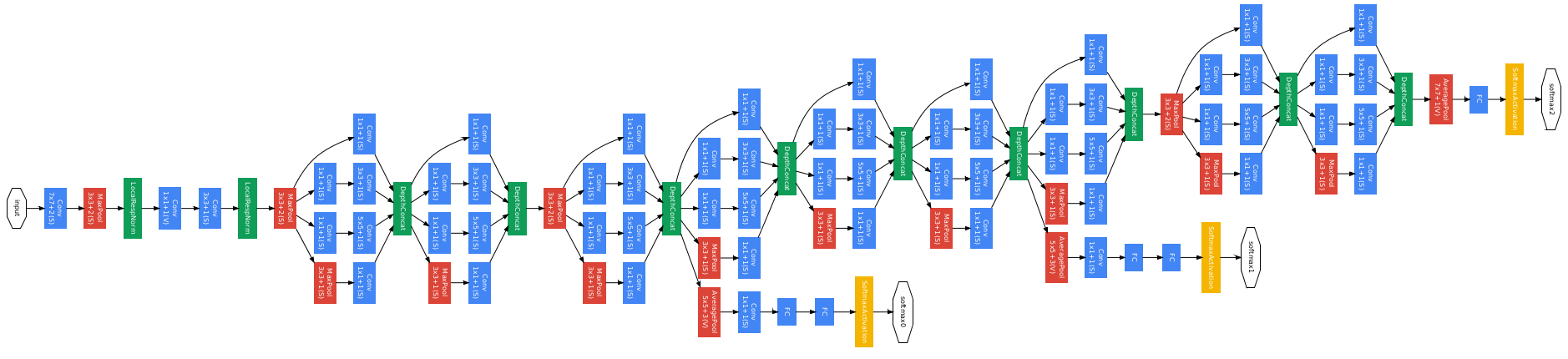}
	\caption{The GoogLeNet (Inception-v1) Neural Architecture (adapted from \cite{inception})}
	\label{fig:net:inception}
\end{figure}

\subsection{GoogLeNet \cite{inception}}

As opposed to the large number of parameters in AlexNet, the GoogLeNet architecture employs smaller series of convolutions organized in repeating \textit{modules}. Inspired by the Network-in-Network \cite{nin} architecture, these modules invoke $1\times 1\times C_{in}$ convolutions (sometimes referred to as $1\times 1$ convolutions). Such modules increase the expressiveness by increasing the depth of the DNN, and at the same time act as dimensionality reduction modules, essentially trading breadth (number of neurons per layer) for depth.

\begin{figure}[h!]
	\includegraphics[height=0.8in]{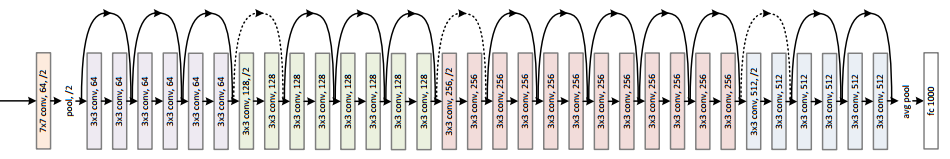}
	\caption{ResNet (adapted from \cite{resnet})}
	\label{fig:net:resnet}
\end{figure}

\subsection{ResNet \cite{resnet}}

The trend of DNNs becoming deeper and narrower continued with successful networks like VGG \cite{vgg} (published at the same time as GoogLeNet), comprising up to 30 layers of convolutions. However, as networks increased in depth, their successful training had become harder. 

The authors of ResNet address the depth issue by training a slightly different inter-layer interaction: instead of composing layers as described in Section \ref{sec:dnn:bprop}, every convolutional module would add its input to the output (as shown in Fig. \ref{fig:net:resnet}). Residuals are implemented as ``shortcut'' identity connections to the network. The system then trains layers with respect to their residuals instead of their original values, and, according to the authors, this solves the inherent degradation in accuracy as networks become deeper. 

With ResNet, it became possible to train networks with depths of 50 to 152 layers, further increasing the quality of the results by allowing higher-level features to be learned.

\begin{figure}[h!]
	\centering
	\begin{subfigure}[b]{0.2\linewidth}
		\centering
		\includegraphics[height=0.6in]{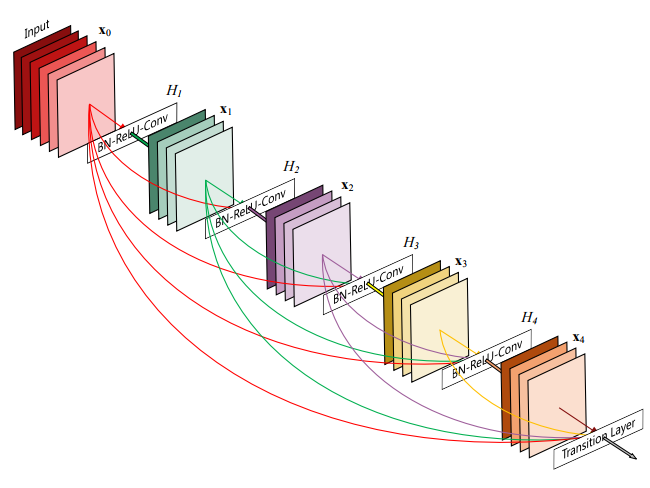}
		\caption{Dense Block}
	\end{subfigure}
	\qquad
	\begin{subfigure}[b]{0.7\linewidth}
		\centering
		\includegraphics[height=0.5in]{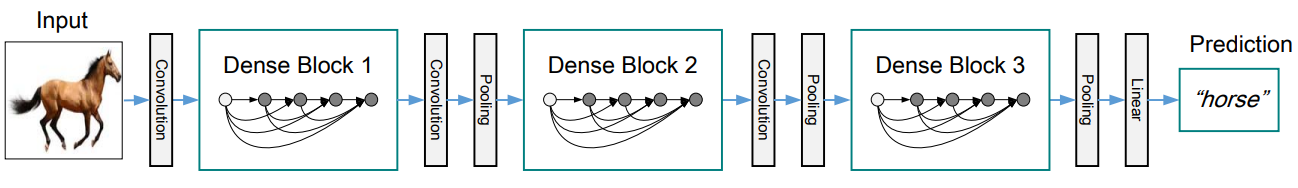}
		\caption{Full Network}		
	\end{subfigure}

	\caption{DenseNet (adapted from \cite{densenet})}
	\label{fig:net:densenet}
\end{figure}

\subsection{DenseNet \cite{densenet}}

Following the success of ResNets, DenseNets further increase the number of connections between layers. As opposed to the module identity shortcut, densely-connected blocks \textbf{concatenate} each layer's outputs to the inputs of the next layers, similar to \cite{deepstack}. According to the authors, this type of connection induces better gradient propagation, as features in subsequent layers are not required to be strictly high-level. Rather, since each layer also receives the inputs of the previous level, features can be constructed from both low-level information and the resulting high-level outputs.

The practical result is that with half the parameters and required operations, DenseNets achieve similar results as ResNets. When their depth is increased to 250 layers, DenseNets reduce the validation error by a factor of $\sim$2 in CIFAR-10 compared to ResNets. This comes at the cost of increasing the number of parameters by almost an order of magnitude. Specifically, ResNets achieve 6.41\% error with 1.7M parameters, whereas DenseNets achieve 3.62\% error with 15.3M parameters.

\subsection{Outlook}
In summary, recent DNN architectures for image recognition heavily base on convolutional operations, are very deep, and become narrower with time. While it is not known which direction the next breakthrough will take; efforts on DNN compression \cite{deepcompression} and more recent architectures \cite{dpn,regevo} seem to focus on decreasing the number of parameters and operations, while maintaining the same accuracy or increasing it. Nevertheless, it is still necessary to use vast computational resources to train these networks, let alone find new networks automatically.

\section{DNN Layer Computation Formulas}
\label{app:layers}

\subsection{Preamble}
The loss function is $\ell$, we are computing the tensor $y$ from the input tensor $x$, using a layer function $f$ with parameters $w$. Overall, the following three functions have to be computed:
\begin{enumerate}
	\item \textbf{Forward evaluation}: $y\equiv f(w,x)$
	\item \textbf{Gradient w.r.t. parameters}: $\nabla w\equiv \frac{\partial \ell}{\partial w}=\frac{\partial \ell}{\partial y}\frac{\partial y}{\partial w}$ (chain rule).
	\item \textbf{Gradient backpropagation}: $\nabla x\equiv \frac{\partial \ell}{\partial x}=\frac{\partial \ell}{\partial y}\frac{\partial y}{\partial x}$.
\end{enumerate}

In the backpropagation algorithm, we use the computed $\nabla x$ to compute the gradients of the preceding layers in the DNN.

\subsection{Activation}
\begin{itemize}
	\item $x\in\mathbb{R}^{W\times H\times C\times N}$; No parameters, thus $w$ is nonexistent;
	\item $f(x) = \sigma(x)$;
	\item $\nabla x_{i,j,k,l} = \frac{\partial \ell}{\partial y_{i,j,k,l}}\cdot \sigma'(x_{i,j,k,l})$.
	\item Examples of activation functions:
	\begin{enumerate}
		\item $ReLU(x) = \max\{0,x\}$; $ReLU'(x) = \begin{cases}1&x>0\\0&\text{otherwise}\end{cases}$.
		\item Sigmoidal function: $\sigma(x)=\frac{1}{1+e^{-x}}; \sigma'(x)=\sigma(x)\cdot(1-\sigma(x))$.
	\end{enumerate}
\end{itemize}

\subsection{Fully Connected Layers}
\begin{itemize}
	\item $w\in\mathbb{R}^{C_{out}\times C_{in}}$, $x\in\mathbb{R}^{C_{in}\times N}$, and $\frac{\partial\ell}{\partial y}\in\mathbb{R}^{C_{out}\times N}$;
	\item $f(w,x) = w\cdot x$;
	\item $\nabla w = x\cdot \left(\frac{\partial\ell}{\partial y}\right)^T$;
	\item $\nabla x = w^T\cdot \frac{\partial\ell}{\partial y}$.
\end{itemize}

\subsection{Convolution (Direct)}
\begin{itemize}
	\item $w\in\mathbb{R}^{C_{out}\times C_{in}\times K_y\times K_x}$, $x\in\mathbb{R}^{W\times H\times C_{in}\times N}$, $y\in\mathbb{R}^{W'\times H' \times C_{out}\times N}$, and $\frac{\partial\ell}{\partial y}\in\mathbb{R}^{W'\times H' \times C_{out}\times N}$;
	\item $W',H'$ are the output sizes of the convolution, defined as $W'=\left\lfloor\frac{W-K_x+2Pad_x}{Stride_x}\right\rfloor+1$ and $H'=\left\lfloor\frac{H-K_y+2Pad_y}{Stride_y}\right\rfloor+1$.
	\item Using input feature $c$ and output feature $c'$:
	\begin{enumerate}
		\item $y_{c'} = \sum_{c=0}^{C_{in}}x_c*w_{c',c}$;
		\item $\frac{\partial \ell}{\partial x_c}=\frac{\partial\ell}{\partial y_c'}*w^T_{c',c}$;
		\item $\frac{\partial \ell}{\partial w_{c',c}}=\frac{\partial\ell}{\partial y_c'}*x_c$.
	\end{enumerate}
\end{itemize}

\subsection{Pooling}
\begin{itemize}
	\item No parameters, $x,y,\frac{\partial\ell}{y}\in\mathbb{R}^{W\times H\times C\times N}$;
	\item $y_{i,j,k,l}=\max_{k_x\in [-K_x,K_x],k_y\in [-K_y,K_y]}x_{i+k_x,j+k_y,k,l}$;
	\item $\nabla x_{i,j,k,l}=\begin{cases}1 & y_{i,j,k,l} = x_{i,j,k,l}\\0 & \text{otherwise}\end{cases}$.
\end{itemize}

\subsection{Batch Normalization}
\begin{itemize}
	\item $w=\{\gamma_k,\beta_k\}_{k=0}^{C}\in\mathbb{R}^{2\times C}$, $x,y\in\mathbb{R}^{C\times N}$;
	\item Forward evaluation algorithm:
	\begin{enumerate}
		\item $E_k\leftarrow \frac{1}{N}\sum_{i=0}^Nx_{k,i}$;
		\item $V_k\leftarrow \frac{1}{N}\sum_{i=0}^N\left(x_{k,i}-E_k\right)^2$;
		\item $\hat{x}_{k,i}\leftarrow\frac{x_{k,i}-E_k}{\sqrt{V_k+\varepsilon}}$
		\item $y_{k,i}\leftarrow\gamma\cdot\hat{x}_{k,i}+\beta$
	\end{enumerate} 
	\item Gradients:
	\begin{enumerate}
		\item $\frac{\partial \ell}{\partial \gamma_k}=\sum_{i=0}^{N}\frac{\partial \ell}{\partial y_{k,i}}\hat{x}_{k,i}$
		\item $\frac{\partial \ell}{\partial \beta_k}=\sum_{i=0}^{N}\frac{\partial \ell}{\partial y_{k,i}}$
	\end{enumerate}
	\item Backpropagation (unsimplified):
	\begin{enumerate}
		\item $\nabla\sigma_k=\sum_{m=0}^{N}\left(\frac{\partial \ell}{\partial y_{k,m}}\gamma_k\cdot(x_{k,m}-E_k)\cdot (V_k+\varepsilon)^{-3/2}\right)$
		\item $\frac{\partial \ell}{\partial x_{k,i}}=\frac{\partial \ell}{\partial y_{k,i}}\cdot\frac{\gamma_k}{\sqrt{V_k+\varepsilon}}-\frac{x_{k,i}-E_k}{N}\nabla\sigma_k+\frac{1}{N}\sum_{m=0}^{N}\left(\frac{\partial \ell}{\partial y_{k,m}}\cdot\frac{-\gamma_k}{\sqrt{V_k+\varepsilon}}\right)+\frac{\sum_{m=0}^N(x_{k,m}-E_k)}{N^2}\nabla\sigma_k$
	\end{enumerate}
\end{itemize}

\newpage
\section{Convolution Computation Analysis}
\label{app:conv}

Assuming convolution of a 4D tensor with a 4D kernel:

\begin{itemize}
	\item Input tensor ($x$) shape: $N\times C_{in}\times H\times W$.
	\item Kernel tensor ($w$) shape: $C_{out}\times C_{in}\times K_y\times K_x$.
	\item Output tensor ($y$) shape: $N\times C_{out}\times H'\times W'$.
	\item In the general case $W'=\left\lfloor\frac{W-K_x+2P_x}{S_x}\right\rfloor+1$ and $H'=\left\lfloor\frac{H-K_y+2P_y}{S_y}\right\rfloor+1$ for padding $P_x,P_y$ and strides $S_x,S_y$. 
	\item However, assuming zero padding and a stride of 1 element we obtain $W'=W-K_x+1$ and $H'=H-K_y+1$.
\end{itemize}

\subsection{Direct Convolution}

Algorithm:

\begin{algorithm}[h!]
	\begin{algorithmic}[1]
		\For{$i = 0$ \textbf{to} $N$ \textbf{in parallel}}
			\For{$j = 0$ \textbf{to} $C_{out}$ \textbf{in parallel}}
				\For{$k = 0$ \textbf{to} $H'$ \textbf{in parallel}}
					\For{$l = 0$ \textbf{to} $W'$ \textbf{in parallel}}
						\For{$m = 0$ \textbf{to} $C_{in}$} \Comment Depth: $\log_2 C_{in}$
							\For{$k_y = 0$ \textbf{to} $K_y$} \Comment Depth: $\log_2 K_y$
								\For{$k_x = 0$ \textbf{to} $K_x$} \Comment Depth: $\log_2 K_x$
									\State $y_{i,j,k,l}~$+=$~x_{i,m,k+k_y,l+k_x}\cdot w_{j,m,k_y,k_x}$ \Comment Work: $1$
								\EndFor
							\EndFor
						\EndFor
					\EndFor
				\EndFor
			\EndFor
		\EndFor
	\end{algorithmic}
	\caption{Direct Convolution}
	\label{app:alg:convdirect}
\end{algorithm}

Overall cost: 
\[
\begingroup
\setlength\arraycolsep{2pt}
\begin{matrix*}[l]
	\mathbf{W}&=&N\cdot C_{out}\cdot H'\cdot W'\cdot C_{in}\cdot K_y\cdot K_x \\[6pt]
	\mathbf{D}&=&\left\lceil\log_2 C_{in}\right\rceil+\left\lceil\log_2 K_y\right\rceil+\left\lceil\log_2 K_x\right\rceil
\end{matrix*}
\endgroup
\]

\newpage
\subsection{im2col}

\begin{itemize}
	\item Input matrix $A$ is a result of a data-layout transformation, sized $\left(C_{in}\cdot K_y\cdot K_x\right)\times \left(N\cdot H'\cdot W'\right)$.
	\item Kernel matrix $F$ is the reshaped tensor $w$, with dimensions $C_{out}\times \left(C_{in}\cdot K_y\cdot K_x\right)$.
	\item Output matrix $B$ has a size of $C_{out}\times\left(N\cdot H'\cdot W'\right)$ and is reshaped to the output.
\end{itemize}

Algorithm:

\begin{algorithm}[h!]
	\begin{algorithmic}[1]	
		\For{$i = 0$ \textbf{to} $C_{in}\cdot K_y\cdot K_x$ \textbf{in parallel}}
			\For{$j = 0$ \textbf{to} $N\cdot H'\cdot W'$ \textbf{in parallel}}
				\State $A_{i,j}\leftarrow x_{\ldots}$ \Comment im2col. Work: N/A, Depth: N/A (layout only)
			\EndFor
		\EndFor
		
		\State \Comment Matrix Multiplication 
		\State $B\leftarrow F\cdot A$ \Comment Work: $C_{out}\cdot (C_{in}\cdot K_y\cdot K_x)\cdot (N\cdot H'\cdot W')$
		\State \Comment Depth: $\log_2 \left(C_{in}\cdot K_y\cdot K_x\right)$
		
		\For{$i = 0$ \textbf{to} $N$ \textbf{in parallel}} 
			\For{$j = 0$ \textbf{to} $C_{out}$ \textbf{in parallel}}
				\For{$k = 0$ \textbf{to} $H'$ \textbf{in parallel}}
					\For{$l = 0$ \textbf{to} $W'$ \textbf{in parallel}}				
						\State $y_{i,j,k,l}\leftarrow B_{\ldots}$ \Comment col2im. Work: N/A, Depth: N/A (layout only)
					\EndFor
				\EndFor
			\EndFor
		\EndFor		
	\end{algorithmic}
	\caption{im2col Convolution}
	\label{app:alg:convim2col}
\end{algorithm}

Overall cost: 
\[
\begingroup
\setlength\arraycolsep{2pt}
\begin{matrix*}[l]
\mathbf{W}&=&N\cdot C_{out}\cdot H'\cdot W'\cdot C_{in}\cdot K_y\cdot K_x \\[6pt]
\mathbf{D}&=&\left\lceil\log_2 C_{in}\right\rceil+\left\lceil\log_2 K_y\right\rceil+\left\lceil\log_2 K_x\right\rceil
\end{matrix*}
\endgroup
\]

\newpage
\subsection{FFT}

\begin{itemize}
	\item Formula: $y_{i,j,*,*}=\mathcal{F}^{-1}\left(\sum_{k=0}^{C_{in}}\mathcal{F}\left(x_{i,k,*,*}\right)\circ\mathcal{F}\left(w_{j,k,*,*}\right)\right)$, where $\circ$ denotes element-wise multiplication and $w$ is padded to $H\cdot W$.
	\item $\hat{x},\hat{w}$ are the transformed inputs and kernels reshaped as 3D tensors, obtained by batching 2-dimensional FFTs. 
	\item We refer to $\hat{x}_k$ (or $\hat{w}_k$) as the $k$th 2D slice in the tensor.
	\item In practical implementations, $\hat{x}$ and $\hat{w}$ are reshaped to $W\cdot H$ 2D matrices (sized $N\times C_{in}$ and $C_{in}\times C_{out}$ respectively) to transform the point-wise multiplication and sum from the above formula to a batched complex matrix-matrix multiplication.
\end{itemize}

Algorithm:

\begin{algorithm}[h!]
	\begin{algorithmic}[1]	
		\For{$i = 0$ \textbf{to} $C_{out}$ \textbf{in parallel}}
			\For{$j = 0$ \textbf{to} $C_{in}$ \textbf{in parallel}}	
				\State $\hat{w}_{i,j}\leftarrow \mathcal{F}(w_{i,j})$ \Comment 2D FFT. Work: $c\cdot HW \log_2 HW$, Depth: $\log_2 HW$
			\EndFor
		\EndFor
		\For{$i = 0$ \textbf{to} $N$ \textbf{in parallel}} 
			\For{$j = 0$ \textbf{to} $C_{in}$ \textbf{in parallel}}
				\State $\hat{x}_{i,j}\leftarrow \mathcal{F}(x_{i,j})$ \Comment 2D FFT. Work: $c\cdot HW \log_2 HW$, Depth: $\log_2 HW$
			\EndFor
		\EndFor
		
		\For{$i = 0$ \textbf{to} $N$ \textbf{in parallel}} 
			\For{$j = 0$ \textbf{to} $C_{out}$ \textbf{in parallel}}
				\For{$k = 0$ \textbf{to} $H'$ \textbf{in parallel}}
					\For{$l = 0$ \textbf{to} $W'$ \textbf{in parallel}} \Comment Batched MM
						\State $\hat{y}_{i,j}=\sum_{m=0}^{C_{in}}\hat{x}_{i,m}\cdot \hat{w}_{j,m}$ \Comment Work: $H\cdot W \cdot N \cdot C_{in} \cdot C_{out}$
					\EndFor \Comment Depth: $\log_2 C_{in}$
				\EndFor					
			\EndFor					
		\EndFor
		
		\For{$i = 0$ \textbf{to} $N$ \textbf{in parallel}} 
			\For{$j = 0$ \textbf{to} $C_{out}$ \textbf{in parallel}}
				\State $y_{i,j}\leftarrow \mathcal{F}^{-1}(\hat{y}_{i,j})$ \Comment 2D IFFT. Work: $c\cdot H'W' \log_2 H'W'$, Depth: $\log_2 H'W'$
			\EndFor
		\EndFor
	\end{algorithmic}
	\caption{FFT Convolution}
	\label{app:alg:convfft}
\end{algorithm}

Overall cost: 
\[
\begingroup
\setlength\arraycolsep{2pt}
\begin{matrix*}[l]
\mathbf{W}&=&c\cdot HW\log_2 (HW)\cdot (C_{out}\cdot C_{in} + N\cdot C_{in} + N\cdot C_{out}) + H\cdot W \cdot N \cdot C_{in} \cdot C_{out}; \\[6pt]
\mathbf{D}&=&2\left\lceil\log_2 HW\right\rceil+\left\lceil\log_2 C_{in}\right\rceil;
\end{matrix*}
\endgroup
\]
where $c$ is the hidden constant in 2D FFT.

\newpage
\subsection{Winograd}

\begin{itemize}
	\item Assuming $F(m\times m, r\times r)$ \cite{lavin16}, i.e., output tile size of $m\times m$ and convolution kernel size is $r\times r$. 
	\item $\alpha = m+r-1$ is the input tile size. Neighboring tiles overlap by $r-1$.
	\item Total number of tiles: $P=N\cdot \left\lceil H/m \right\rceil\cdot \left\lceil W/m \right\rceil$.
	\item $A\in \mathbb{R}^{\alpha\times m},B\in \mathbb{R}^{\alpha\times \alpha},G\in \mathbb{R}^{\alpha\times r}$ are Winograd minimal filtering matrices.
\end{itemize}

Algorithm:

\begin{algorithm}[h!]
	\begin{algorithmic}[1]	
		\For{$k = 0$ \textbf{to} $C_{out}$ \textbf{in parallel}}
			\For{$c = 0$ \textbf{to} $C_{in}$ \textbf{in parallel}}
				\State $u\leftarrow Gw_{k,c,*,*}G^T$ \Comment Winograd transform $r\times r \rightarrow \alpha\times \alpha$. Work: $\mathbf{W}_{WG}$, Depth: $\mathbf{D}_{WG}$
				\For{$s = 0$ \textbf{to} $\alpha$ \textbf{in parallel}}				
					\For{$n = 0$ \textbf{to} $\alpha$ \textbf{in parallel}}
						\State $U_{k,c}^{(s,n)}\leftarrow u_{s,n}$ \Comment Scatter. Work: N/A, Depth: N/A (layout only)
					\EndFor
				\EndFor
			\EndFor
		\EndFor
		\For{$b = 0$ \textbf{to} $P$ \textbf{in parallel}}
			\For{$c = 0$ \textbf{to} $C_{in}$ \textbf{in parallel}}
				\State $v\leftarrow B^Tx_{b,c,*,*}B$ \Comment Winograd transform $\alpha \times \alpha \rightarrow \alpha\times \alpha$. Work: $\mathbf{W}_{WD}$, Depth: $\mathbf{D}_{WD}$
				\For{$s = 0$ \textbf{to} $\alpha$ \textbf{in parallel}}				
					\For{$n = 0$ \textbf{to} $\alpha$ \textbf{in parallel}}
						\State $V_{c,b}^{(s,n)}\leftarrow v_{s,n}$ \Comment Scatter. Work: N/A, Depth: N/A (layout only)
					\EndFor
				\EndFor	
			\EndFor
		\EndFor	
		\For{$s = 0$ \textbf{to} $\alpha$ \textbf{in parallel}}				
			\For{$n = 0$ \textbf{to} $\alpha$ \textbf{in parallel}}
				\State $Z^{(s,n)}\leftarrow U^{(s,n)}\cdot V^{(s,n)}$ \Comment Matrix Multiplication. Work: $C_{out}\cdot C_{in} \cdot P$, Depth: $\log_2 C_{in}$
			\EndFor
		\EndFor	
		\For{$k = 0$ \textbf{to} $C_{out}$ \textbf{in parallel}}
			\For{$b = 0$ \textbf{to} $P$ \textbf{in parallel}}
				\For{$s = 0$ \textbf{to} $\alpha$ \textbf{in parallel}}				
					\For{$n = 0$ \textbf{to} $\alpha$ \textbf{in parallel}}
						\State $z_{s,n}\leftarrow Z_{k,b}^{(s,n)}$ \Comment Gather. Work: N/A, Depth: N/A (layout only)
					\EndFor
				\EndFor	
				\State $y_{b,k,*,*}\leftarrow A^TzA$ \Comment Winograd transform $\alpha \times \alpha \rightarrow m\times m$. Work: $\mathbf{W}_{WY}$, Depth: $\mathbf{D}_{WY}$
			\EndFor
		\EndFor	
	\end{algorithmic}
	\caption{Winograd Convolution}
	\label{app:alg:convwinograd}
\end{algorithm}

A Winograd transform from $a\times a\rightarrow b\times b$ consists of two matrix multiplications, thus:
\[
\begingroup
\setlength\arraycolsep{2pt}
\begin{matrix*}[l]
\mathbf{W}_{WT}(a,b) &=& b\cdot a\cdot a + b\cdot a\cdot b=a^2b+b^2a\\
\mathbf{D}_{WT}(a,b) &=& 2\left\lceil\log_2 a\right\rceil;
\end{matrix*}
\endgroup
\]
and therefore $\mathbf{W}_{WG}=r^2\alpha+\alpha^2r,\mathbf{W}_{WD}=2\alpha^3,\mathbf{W}_{WY}=\alpha^2m+m^2\alpha$, and $\mathbf{D}_{WG}=2\left\lceil\log_2 r\right\rceil,\mathbf{D}_{WD}=2\left\lceil\log_2 \alpha\right\rceil,\mathbf{D}_{WY}=2\left\lceil\log_2 \alpha\right\rceil$.

Overall cost: 
\[
\begingroup
\setlength\arraycolsep{2pt}
\begin{matrix*}[l]
\mathbf{W}&=&\mathbf{W}_{WG}+\mathbf{W}_{WD}+C_{out}\cdot C_{in}\cdot P + \mathbf{W}_{WY}\\
		  &=& \alpha (r^2 + \alpha r + 2\alpha^2+ \alpha m + m^2) + C_{out}\cdot C_{in}\cdot P\\[6pt]
\mathbf{D}&=&2\left\lceil\log_2 r\right\rceil+4\left\lceil\log_2 \alpha\right\rceil+\left\lceil\log_2 C_{in}\right\rceil
\end{matrix*}
\endgroup
\]

\end{document}